\documentclass[12pt]{article}
\usepackage{amsmath, amsthm, amssymb}
\usepackage{graphicx}
\usepackage{natbib}
\usepackage{bbm, dsfont}
\usepackage{optidef}
\usepackage{tikz}
\usepackage{tcolorbox}
\usepackage[table,xcdraw]{xcolor}
\usepackage{bm}  % in preamble
\usepackage{enumitem} 
\usetikzlibrary{calc} 
\usepackage{setspace}

\usepackage{booktabs}   % in preamble
\usepackage{array}  
\allowdisplaybreaks[4]

\usetikzlibrary{positioning} % for more styling options if desired
\usepackage{url} 
\usepackage{adjustbox}
\usepackage{enumitem}
\usepackage[linesnumbered,ruled]{algorithm2e}
\usepackage{subcaption}% not crucial - just used below for the URL 
\usepackage{hyperref}
\definecolor{darkbyzantium}{rgb}{0.36, 0.22, 0.33}
\definecolor{darkpastelgreen}{rgb}{0.01, 0.75, 0.24}
\newcommand{\blind}{0}

\newtheorem{proposition}{Proposition}
\newtheorem{lemma}{Lemma}
\newtheorem{theorem}{Theorem}
\newtheorem{corollary}{Corollary}
\newtheorem{remark}{Remark}
\newenvironment{narrowquote}
  {\begin{list}{}{\leftmargin=1em \rightmargin=1em}\item\relax}
  {\end{list}}

\newcommand{\bx}{{\bf x}}

\newcommand{\bepsilon}{\boldsymbol \epsilon}

\newcommand{\bff}{{\bf f}}
\newcommand{\bh}{{\bf h}}
\newcommand{\bPh}{\boldsymbol \Psi}
\newcommand{\bY}{{\bf y}}
\SetAlCapFnt{\footnotesize}          % caption text
\SetAlCapNameFnt{\footnotesize}  

\usepackage{setspace}   % for \setstretch
\makeatletter
\let\jasa@origthebibliography\thebibliography
\let\endjasa@origthebibliography\endthebibliography

\renewenvironment{thebibliography}[1]%
  {%
    \jasa@origthebibliography{#1}%
    \setlength{\itemsep}{0pt}% space between \bibitem blocks
    \setlength{\parsep}{0pt}%  space between paragraphs within an item
    \setlength{\topsep}{0pt}%  space above and below the list
    \setlength{\partopsep}{0pt}%
    \setlength{\parskip}{0pt}%  extra space between paragraphs
    \setstretch{1}%           single-line spacing (no extra leading)
  }%
  {%
    \endjasa@origthebibliography%
  }
\makeatother

\begin{document}

\def\spacingset#1{\renewcommand{\baselinestretch}%
{#1}\small\normalsize} \spacingset{1}

%%%%%%%%%%%%%%%%%%%%%%%%%%%%%%%%%%%%%%%%%%%%%%%%%%%%%%%%%%%%%%%%%%%%%%%%%%%%%%

\if0\blind
{
  \title{\bf Extracting Interpretable Models from Tree Ensembles: Computational and Statistical Perspectives}
  \date{}
  \author{
    Brian Liu \\
   MIT
    \and
    Rahul Mazumder \\
   MIT
    \and
    Peter Radchenko \\
    University of Sydney
}
  \maketitle
} \fi

\if1\blind
{
  \bigskip
  \bigskip
  \bigskip
  \begin{center}
    {\LARGE\bf Title}
\end{center}
  \medskip
} \fi

\bigskip
\begin{abstract} 

Tree ensembles are non-parametric methods widely recognized for their accuracy and ability to capture complex interactions. While these models excel at prediction, they are difficult to interpret and may fail to uncover useful relationships in the data. We propose an estimator to extract compact sets of decision rules from tree ensembles. The extracted models are accurate and can be manually examined to reveal relationships between the predictors and the response. A key novelty of our estimator is the flexibility to jointly control the number of rules extracted and the interaction depth of each rule, which improves accuracy. We develop a tailored exact algorithm to efficiently solve optimization problems underlying our estimator and an approximate algorithm for computing regularization paths — sequences of solutions that correspond to varying model sizes. We also establish novel non-asymptotic prediction error bounds for our proposed approach, comparing it to an oracle that chooses the best data-dependent linear combination of the rules in the ensemble subject to the same complexity constraint as our estimator. The bounds illustrate that the large-sample predictive performance of our estimator is on par with that of the oracle. Through experiments, we demonstrate that our estimator outperforms existing algorithms for rule extraction.

% We present an optimization-based framework for extracting compact rule sets from tree ensembles. The resulting rule sets can perform comparably to the original ensemble in terms of predictive accuracy and consist of sequences of conditional statements that practitioners can examine to gain insights into the relationships between the predictors and the response. Moreover, the framework offers flexibility to jointly control both the number of rules extracted and the interaction depth of each rule, which improves compactness and interpretability. We develop a tailored optimal algorithm to efficiently solve problems within the framework, as well as an approximate algorithm for computing regularization paths --- sequences of solutions corresponding to varying model sizes --- which allows practitioners to explore the trade-off between model complexity and accuracy. 
% We establish non-asymptotic prediction error bounds for our proposed approach, comparing it to an oracle that chooses the best data-dependent linear combination of the rules in the ensemble subject to the same complexity constraint as our estimator. We show that the large-sample predictive performance of our estimator is on par with that of the oracle. Through real-world experiments, we demonstrate that our proposed framework substantially outperforms existing algorithms for rule extraction and model compression.
\end{abstract}

\noindent%
\vfill

\newpage
\spacingset{1.75} % DON'T change the spacing! CHANGE BACK TO 1.75 if JCGS!!!!

\section{Introduction}

Tree ensemble algorithms, such as random forests \citep{breiman2001random} and gradient boosting  \citep{friedman2002stochastic}, are among the most well-known methods in statistical modeling and have achieved success across a broad range of disciplines, from remote sensing \citep{belgiu2016random} to genomics \citep{chen2012random}. Since their introduction, tree ensembles have earned a reputation for excellent predictive performance, in fact, Leo Breiman notably claims in \textit{Statistical Modeling: The Two Cultures} \citep{breiman2001statistical} that ``Random Forests are A+ predictors.” By combining a large number of decision trees, either independently trained on bootstrapped samples of the data or sequentially on the residuals of the prior model, tree ensembles attain high predictive accuracy, albeit for an increase in model complexity and often at the expense of interpretability. Compared to a single tree, ensembles are substantially more challenging to interpret and, indeed, for random forests, Breiman~\cite[p.~208]{breiman2001statistical} claims that ``on interpretability, they rate an F." 
\begin{figure}[h]
  \centering
  \begin{adjustbox}{width=.75\linewidth,center}  % ← scale everything to 90 %
  \begin{minipage}{0.375\textwidth}
    \centering
    \includegraphics[width=\linewidth]{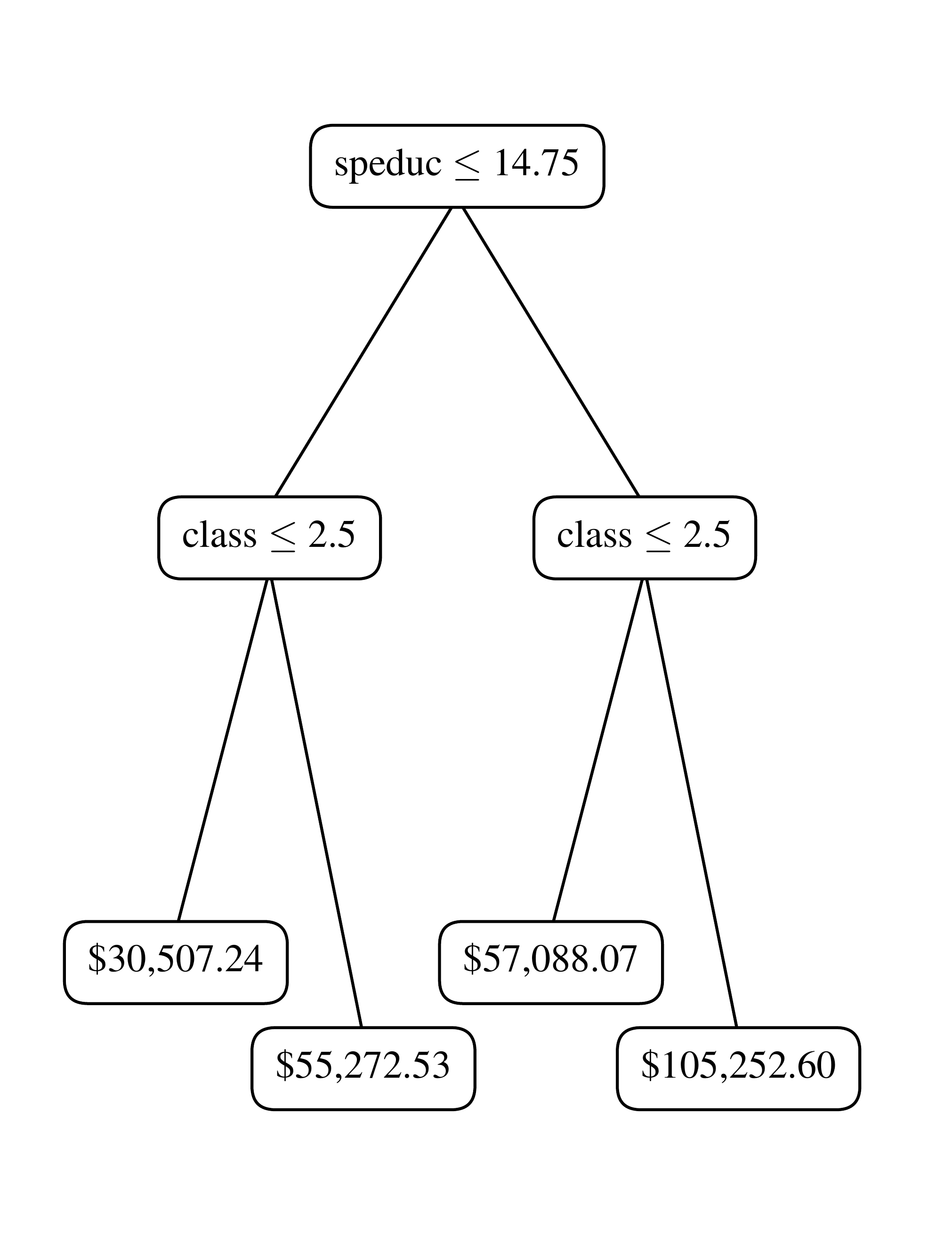}
  \end{minipage}%
  \hfill
  \begin{minipage}{0.6\textwidth}
    \begin{tcolorbox}[colback=white,colframe=black,
                      title={Model Explanation},boxsep=4pt]
      \begin{itemize}[left=0pt,itemsep=4pt,topsep=2pt,parsep=0pt,partopsep=0pt]
        \item If \texttt{speduc} $\leq$ 14.75 and \texttt{class} $\leq$ 2.5, then predicted income = \$30{,}507.24.
        \item If \texttt{speduc} $\leq$ 14.75 and \texttt{class} $>$ 2.5, then predicted income = \$55{,}272.53.
        \item If \texttt{speduc} $>$ 14.75 and \texttt{class} $\leq$ 2.5, then predicted income = \$57{,}088.07.
        \item If \texttt{speduc} $>$ 14.75 and \texttt{class} $>$ 2.5, then predicted income = \$105{,}252.60.
      \end{itemize}
    \end{tcolorbox}
  \end{minipage}
  \end{adjustbox}
  \caption{Decision tree fit to predict household income using survey responses. Feature \texttt{speduc} is the spouse’s education (years); \texttt{class} is the self-identified income class.}
  \label{tree_intro.fig}
\end{figure}

Consider, for example, the single decision tree shown in the left panel of Figure~\ref{tree_intro.fig}, constructed to predict household income based on survey responses from the 2022 General Social Survey (GSS) \citep{GSS2024}---a dataset with 4,000 respondents and 15,000 covariates. This tree is highly interpretable, and its predictions can be fully explained in four sentences corresponding to the four rules listed in the right panel of the figure. Each rule consists of a sequence of conditions obtained by traversing the tree from the root node to a leaf. Tree ensembles can also be described by similar sets of rules, but these are rarely human-interpretable for two reasons. First, the number of rules is prohibitively large; for example, a boosting ensemble of 500 depth 6 trees requires 32,000 rules (sentences) to explain.  Second, especially in deeper ensembles, each individual rule may be too complex to interpret. As Breiman \citep[p.~208]{breiman2001statistical} observes for random forests: \begin{narrowquote}
``their mechanism for producing a prediction is difficult to understand.
Trying to delve into the tangled web that generated a plurality vote from 100 trees is a Herculean task.''
 \end{narrowquote}
Yet, ensembles are significantly more accurate than single trees. On the GSS survey example, the decision tree shown in Figure \ref{tree_intro.fig} achieves an out-of-sample $R^2$ of \textbf{0.28} while the 500 depth 6 tree boosting ensemble mentioned above attains an out-of-sample $R^2$ of \textbf{0.55}.

\vspace{2mm}
\noindent\textbf{Motivation:}
\begin{narrowquote}
    ``Framing the question as the choice between accuracy and interpretability is an incorrect interpretation of what the goal of statistical analysis is. The point of a model is to get useful information about the relation between the response and predictor variables.''~\citep[p.~210]{breiman2001statistical}
\end{narrowquote}

\begin{figure}[h]
    \centering
    \begin{adjustbox}{width=.8\linewidth,center} 
    % Left plot
    \begin{minipage}{0.28\textwidth}
        \centering
        \includegraphics[width=\linewidth, height=0.25\textheight]{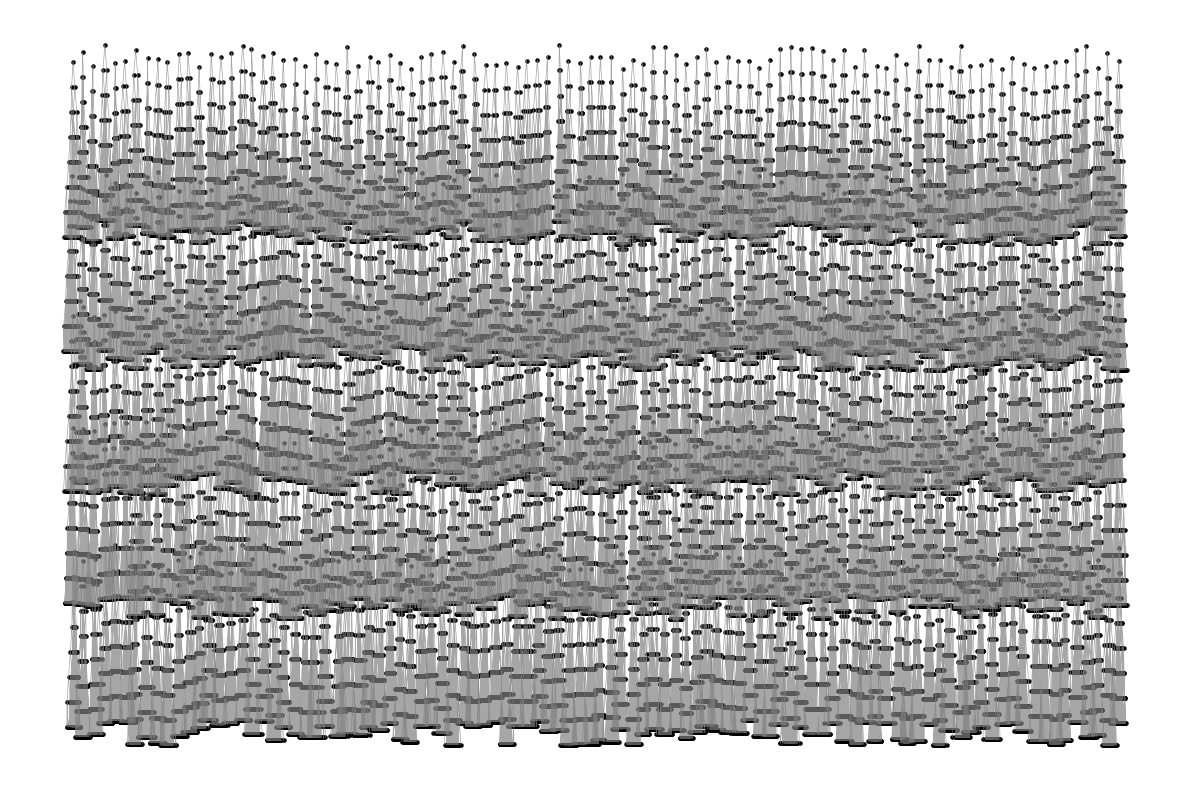}
        \caption*{(a) Boosting ensemble of 500 depth 6 trees: OOS $R^2 = 0.55$.}
    \end{minipage}%
    \hfill
    % Right plot
    \begin{minipage}{0.7\textwidth}
        \centering
        \includegraphics[width=\linewidth]{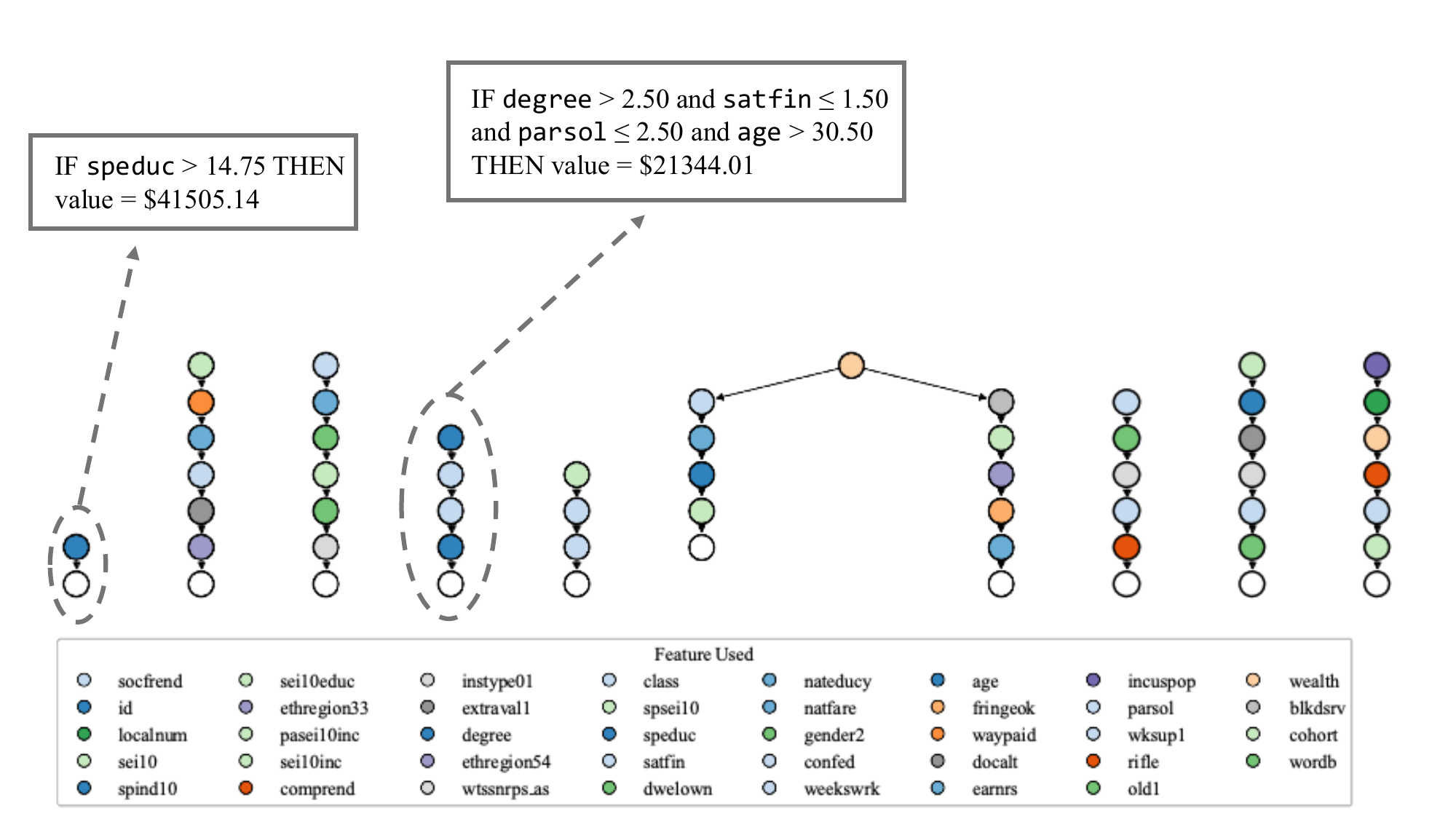}
        \caption*{(b) Extracted model of 10 rules: OOS $R^2 = 0.50$.}
    \end{minipage}
    \end{adjustbox}
    \caption{Application of our proposed estimator to the GSS example.}
    \label{framework_example.fig}
\end{figure}

To bridge the trade-off between model accuracy and interpretability, we propose a novel estimator to extract rule sets from tree ensembles. These rule sets retain predictive accuracy comparable to the full models, while being compact enough for practitioners to manually examine and understand the relationship between the predictors and the response. We consider compactness along two dimensions: the number of rules extracted (i.e., the number of sentences a practitioner must read to explain the model) and the interaction depth of each rule (i.e., the complexity of each sentence). In Figure~\ref{framework_example.fig}, we demonstrate an application of our proposed estimator to the GSS example discussed above. Panel (a) visualizes the original boosting ensemble of 500 depth 6 trees (32,000 rules) that achieves an out-of-sample $R^2$ of 0.55. In panel (b), we apply our estimator to extract  \textbf{10} rules from the ensemble, which achieve an out-of-sample $R^2$ of \textbf{0.50}. Our estimator prunes the interaction depth of many rules and extracts a rule set compact enough to be examined by hand. Our proposed approach builds on prior work by \cite{friedman2008predictive} and \cite{meinshausen2010node}, who use the LASSO and non-negative garrote to extract rules from random forests. Compared to existing methods, a key novelty of our estimator is its ability to jointly control both the number and depths of the extracted rules, a flexibility that leads to substantial improvements in predictive performance. In addition, to the best of our knowledge, we are the first to establish prediction error bounds for pruning rules from tree ensembles. We summarize our main contributions.

\begin{itemize}[leftmargin=10pt,    % no indent
                labelindent=0pt,   % bullet sits at margin
                itemsep=0pt,       % no space between \items
                parsep=0pt,        % no paragraph spacing inside items
                topsep=0pt,        % no space above/below the list
                partopsep=0pt,
                label=$\bullet$] 
    \item We propose a novel estimator that uses an optimization-based framework to extract interpretable models from tree ensembles, 
     offering the flexibility to jointly control both the number of rules extracted and their interaction depths (\S\ref{framework.section}).
    
    \item Our estimator is given by a discrete optimization problem that is difficult to solve due to its non-convexity. We develop a novel, specialized exact algorithm that solves this problem to global optimality and scales to sizes significantly beyond the capabilities of commercial solvers such as Gurobi \citep{gurobi} and Mosek \citep{mosek}.  With this exact algorithm, our estimator has theoretical guarantees (\S\ref{optimal_method.section}).

    % We develop a specialized exact algorithm that solves  problems in this optimization framework to global optimality and scales to sizes significantly beyond the capabilities of commercial solvers such as Gurobi \citep{gurobi} and Mosek \citep{mosek}. With this exact algorithm, our estimator has theoretical guarantees (\S\ref{optimal_method.section}).
    
    \item We develop an approximate algorithm tailored for efficiently extracting sequences of models of varying sizes (i.e., regularization paths), allowing practitioners to assess the trade-off between model complexity and predictive accuracy. Although our estimator, when used with this algorithm, does not offer theoretical guarantees, the extracted rule sets are practically useful and exhibit strong empirical performance (\S\ref{approximate_algorithm.section}) .

    \item We establish non-asymptotic oracle prediction error bounds for our proposed approach. To the best of our knowledge, there are no other existing prediction error bounds for ensemble pruning algorithms (\S\ref{theory.section}).
    \item We show through real world experiments that our estimator consistently outperforms state-of-the-art pruning algorithms at extracting interpretable rule sets \color{black}{as well as algorithms that construct interpretable tree-based algorithms directly from the data (\S\ref{experiments.section}).}
    \item \color{black} We propose a scorecard representation of the models extracted by our estimator that parallels scoring systems used in high-stakes applications, such as criminal justice and healthcare, and demonstrate the interpretability of this representation through a real-world case study (\S\ref{case_study.section}). \color{black}
    
\end{itemize}

\vspace{2mm}
% \noindent{\textbf{Roadmap:}} The remainder of our paper is organized as follows. We first discuss preliminaries on tree ensemble and overview existing rule extraction algorithms (\S\ref{preliminaries.section}), along with the advantages of our proposed approach (\S\ref{our_proposed_framework.section}). We then introduce our estimator to extract interpretable rule sets from tree ensembles, that jointly controls for the number of rules extracted and interaction depth (\S\ref{framework.section}). We present our exact algorithm to solve problems formulated within our estimator to global optimality  (\S\ref{optimal_method.section}) as well as an approximate algorithm (\S\ref{approximate_algorithm.section}) to efficiently compute regularization paths. We then present a theoretical analysis of our proposed approach (\S\ref{theory.section}). Finally, we conclude with experimental evaluation of the computation time and predictive performance of our proposed approach (\S\ref{experiments.section}).

\subsection{Preliminaries and Related Work} \label{preliminaries.section}

We first overview  trees, rules, and ensembles, and then discuss existing rule-extraction (pruning) methods while highlighting the advantages of our proposed approach.

\noindent{\textbf{Trees, Rules, and Ensembles:}} We focus on regression trees for predicting a continuous response and use the terms regression tree and decision tree interchangeably throughout this paper. Regression trees consist of internal nodes and leaf nodes, where internal nodes represent splits that divide the data based on a selected feature and threshold. Splits are constructed greedily to grow the tree until a prespecified stopping criterion, such as the maximum depth of the tree, is reached. In the resulting leaf nodes, all observations in the node are assigned the same predicted value, the average response. Trees can be decomposed into sets of decision rules (see \S\ref{trees_to_rules.appx}), where rules are defined by the sequences of splits obtained by traversing the tree from the root to leaves. The prediction of the tree is the sum of its rules, an important property that we leverage in our proposed estimator. Tree ensembles improve accuracy by combining trees, and these powerful models often often outperform deep learning algorithms on tabular data \citep{shwartz2022tabular}.

% . Popular approaches include bagging (e.g., random forests, adaptive bagging), which averages independently trained trees, and boosting, which fits each tree on the residuals of the prior model. These ensembles are powerful and often outperform deep learning algorithms on tabular data \citep{shwartz2022tabular}. However, this gain in accuracy comes at the cost of increased model size and complexity.

% Tree ensembles can grow to be extremely large
% % , for example, a random forest of 1000 decision trees fit on a dataset with 100000 observations can contain over 100 million leaf nodes.
% and, as such, these ensembles can be unwieldy, slow at inference, and difficult to interpret due to their complexity.

\vspace{2mm}
\noindent{\textbf{Pruning:}} 
Tree ensembles can be pruned to improve model compactness and interpretability. One of the earliest computational approaches, ISLE, uses the LASSO to extract entire decision trees from an ensemble \citep{friedman2003importance}. Subsequently, RuleFit \citep{friedman2008predictive} decomposes a tree ensemble into a large collection of decision rules and uses the LASSO to select a sparse subset. A closely related algorithm, Node Harvest \citep{meinshausen2010node}, applies the non-negative garrote. Both RuleFit and Node Harvest become computationally expensive when the number of leaf nodes is large, which limits their application to smaller, shallower ensembles, typically of depth $\leq 3$. This limitation is restrictive, as deeper ensembles and rules may be needed to capture higher-order interactions in the data. More recently, FIRE \citep{liu2023fire} uses techniques from non-convex optimization to improve the computational efficiency and predictive performance of rule extraction. Importantly, none of these methods are able to prune the depths of the extracted rules (see \S\ref{related_works.appx} for  additional discussion).

Interaction depth plays an important role in model complexity; in fact the depth of an ensemble, or its "interaction order is perhaps a more natural way to think of model complexity" \citep[Chapter~17]{efron2021computer}
 compared to other parameters such as the number of estimators. Extracting rules from deep ensembles without pruning their depth may not yield an interpretable model, as deep rules with long sequences of conditional statements can be difficult to understand. Conversely, a shallow rule set extracted from a shallow ensemble may lack the flexibility needed for strong predictive performance. Since existing rule extraction algorithms do not support depth pruning, they are unable to select rules of varying complexities. Recently, ForestPrune \citep{liu2023forestprune}, introduces an algorithm for pruning depth layers from tree ensembles. However, it is limited to pruning entire trees and does not support rule extraction. As a result, ForestPrune produces models that are less flexible than rule-based approaches.

\begin{figure}[h]
    \centering
    \includegraphics[width=0.85\linewidth]{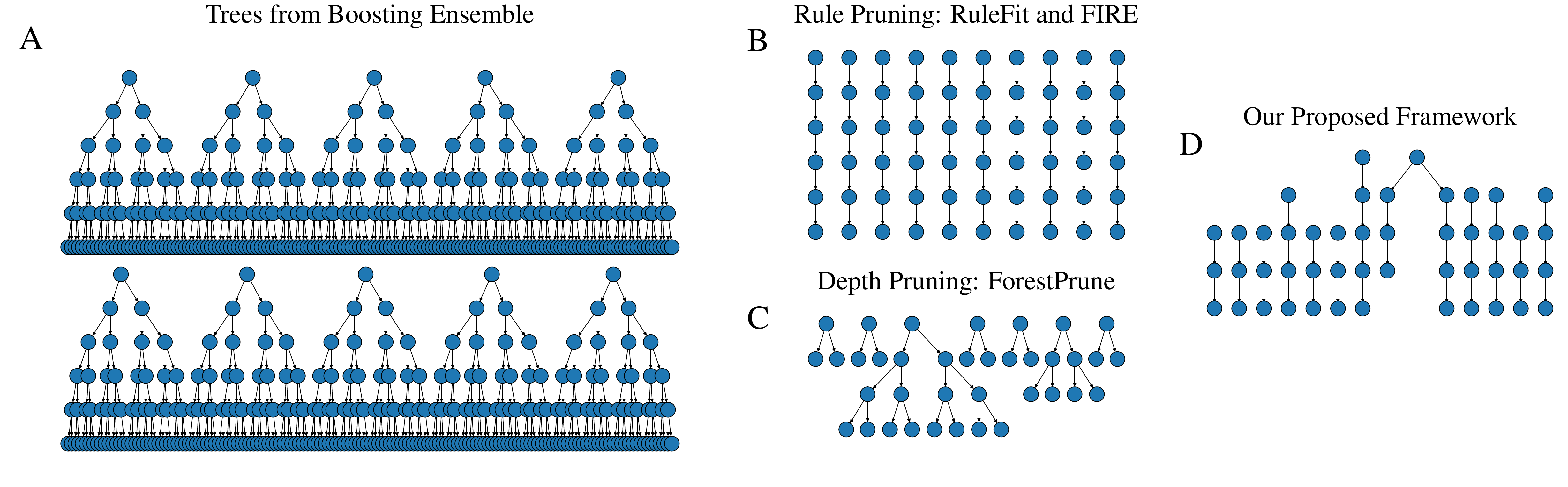}
    \caption{Our proposed framework, which jointly prunes depth and rules, compared with existing pruning approaches.}
    \label{prelim_example.fig}
\end{figure}

\vspace{2mm}
\noindent{\textbf{Our Proposed Approach:}} We propose a novel estimator that uses an optimization-based framework to simultaneously prune both rules and their interaction depths. Compared to existing approaches, our estimator is significantly more flexible and can extract rules with varying complexities. This allows us to extract  shallow rules for interpretability while retaining select deeper rules to capture higher-order interactions. We illustrate an application of our framework in comparison with existing methods in Figure~\ref{prelim_example.fig}. Specifically, we fit a boosting ensemble of 250 depth 5 decision trees on the \textsc{Wind} dataset from \cite{haslett1989space}; a sample of 10 trees from the ensemble is displayed in panel A. In panel B, we illustrate the application of a rule pruning method such as RuleFit or FIRE. These methods extract rules without pruning depth, so all extracted rules retain the same depth as the original ensemble (depth 5). In panel C, we illustrate the application of ForestPrune. This method prunes entire trees but cannot extract individual rules, so the resulting model consists of a collection of pruned trees. Finally, in panel D, we illustrate the application of our framework, which jointly prunes both rules and interaction depth. This added flexibility improves the predictive performance of our estimator, compared to all existing methods, as demonstrated in \S\ref{interpretable_rules_experiment.section}. We also note that, in contrast to our method, none of the existing algorithms for pruning tree ensembles discussed above offer theoretical guarantees.

\section{Proposed Estimator} \label{framework.section}

In this section, we present our estimator to jointly prune interaction depth and decision rules from a tree ensemble. Our estimator relies on the fact that we can represent the prediction of a tree ensemble as a linear combination of the extracted rules from each tree.

\vspace{2mm}
\noindent{\textbf{Notation:}} Given data matrix $X \in \mathbbm{R}^{n\times p}$ and mean-centered response $\mathbf{y} \in \mathbbm{R}^n$, we fit a tree ensemble $\mathcal{T}$ of~$T$ decision trees, $\{\Gamma_1(\mathbf{x}) \ldots \Gamma_T(\mathbf{x})\}$, which are functions of the feature vector $\bx\in\mathbb{R}^p$. Let~$m$ denote the total number of nodes in $\mathcal{T}$. Each decision tree $\Gamma_t(\mathbf{x})$, for all $t \in \mathcal{T}$ contains a set of nodes, where $\mathcal{N}_t$ is the set of node indices for tree $t$. This set includes \emph{both} the leaf and the internal nodes. Each node $i \in \mathcal{N}_t$ has a pre-specified attribute $a_i^t$: this attribute can represent the depth of the node in the tree, the number of observations in the node, or some other prespecified characteristic. Also, for each node $i \in \mathcal{N}_t$ let $\mathcal{C}_i^t$ represent the set of successor (descendant) nodes in tree $t \in \mathcal{T}$ and let $\mathcal{P}_i^t$ represent the set of predecessor (ancestor) nodes. Finally, we use~${\mathbf{M}_i^t} $ to represent the prediction vector for node~$i$ of tree~$t$, for all $i \in \mathcal{N}_t$ and $t \in \mathcal{T}$. More specifically, we define vector ${\mathbf{M}_i^t} \in \mathbb{R}^n$ element-wise as follows: \(
(\mathbf{M}_i^{\,t})_k = \mu_i^t\,\mathbbm{1}_{\{\text{node } i \text{ of tree } t \text{ contains obs.\ }k\}}
\), where $\mu_i^t$ is the average response of all of the training observations in node $i$ of tree $t$.

\subsection{ Formulation} \label{formulation.section}
% \begin{figure}[h]
%     \centering
%    \begin{tikzpicture}[
%   every node/.style = {circle, draw, minimum size=7mm, inner sep=2pt},
%   % Adjust these styles to increase or decrease spacing
%   level 1/.style = {level distance=15mm, sibling distance=35mm},
%   level 2/.style = {level distance=15mm, sibling distance=20mm}
% ]
% \node (z1) {$z_1^t, w_1^t$}
%   child {
%     node (z2)  {$z_2^t, w_2^t$}
%       child { node (z4)  {$z_4^t, w_4^t$}}
%       child { node (z5) {$z_5^t, w_5^t$} }
%   }
%   child {
%     node (z3) {$z_3^t, w_3^t$}
%       child { node (z6) {$z_6^t, w_6^t$} }
%       child { node (z7) {$z_7^t, w_7^t$} }
%   };
% \end{tikzpicture}
%     \caption{Visualization of decision variables in our optimization formulation for a single decision tree $t \in \mathcal{T}$ of depth 2. Each binary decision variable $z_i^t$ indicates whether node $i$ is a leaf node in the decision tree and continuous variable $w_i^t$ represents the weight. For each node $i$ in tree $t$, $\mathcal{C}_i^t$ represents the set of child nodes. For example for node 3, $\mathcal{C}_3^t = \{6,7 \}$. If node 3 is set to a leaf node in tree $t$, $z_3^t = 1$, nodes 6 and 7 cannot be leaf nodes, i.e., we must have $z_6^t = 0$ and $z_7^t = 0$.
%     }\label{tree_based_vis.pdf}
% \end{figure}
% \vspace{2mm}

\begin{figure}[h]
    \centering
    \begin{adjustbox}{width=.7\linewidth,center}  
    \begin{subfigure}{0.48\textwidth}
        \centering
        \begin{tikzpicture}[
          every node/.style = {circle, draw, minimum size=7mm, inner sep=2pt},
          % Adjust these styles to increase or decrease spacing
          level 1/.style = {level distance=15mm, sibling distance=35mm},
          level 2/.style = {level distance=15mm, sibling distance=20mm}
        ]
        \node (z1) {$z_1^t, w_1^t$}
          child {
            node (z2)  {$z_2^t, w_2^t$}
              child { node (z4)  {$z_4^t, w_4^t$}}
              child { node (z5) {$z_5^t, w_5^t$} }
          }
          child {
            node (z3) {$z_3^t, w_3^t$}
              child { node (z6) {$z_6^t, w_6^t$} }
              child { node (z7) {$z_7^t, w_7^t$} }
          };
        \end{tikzpicture}
        \caption{Variables in formulation.}
    \end{subfigure}
    \hfill
    \begin{subfigure}{0.48\textwidth}
        \centering
        \begin{tikzpicture}[
          every node/.style = {circle, draw, minimum size=7mm, inner sep=2pt},
          % Adjust these styles to increase or decrease spacing
          level 1/.style = {level distance=15mm, sibling distance=35mm},
          level 2/.style = {level distance=15mm, sibling distance=20mm}
        ]
        \node(z1) {$z_1^t = 0$}
          child {
            node (z2)  {$z_2^t = 0$}
              child { node[draw=darkpastelgreen, text=darkpastelgreen, thick]  (z4)  {$z_4^t = 1$}}
              child { node (z5) {$z_5^t= 0 $} }
          }
          child {
            node[draw=darkpastelgreen, text=darkpastelgreen, thick] (z3) {$z_3^t= 1$}
              child { node (z6) {$z_6^t = 0$} }
              child { node (z7) {$z_7^t= 0$} }
          };
        \end{tikzpicture}
        \caption{Pruning 2 decision rules from the tree.}
    \end{subfigure}
    \end{adjustbox}
    \caption{Set
    $\mathcal{C}_i^t$ represents the  descendants of node $i$, e.g.,  $\mathcal{C}_3^t = \{6,7 \}$. If node 3 is selected to be a terminal node of a extracted rule then nodes 6 and 7 cannot be selected. }
    \label{tree_based_vis.fig}
\end{figure}

We present the optimization-based formulation of our proposed estimator. For each node $i \in \mathcal{N}_t$, for all $t \in \mathcal{T}$, we introduce a binary decision variable $z_i^t \in \{0,1\}$ and a continuous decision variable $w_i^t \in \mathbb{R}$. A visualization of these decision variables for a single tree $t \in \mathcal{T}$ is shown in the left panel of Figure~\ref{tree_based_vis.fig}. The binary variable $z_i^t$ indicates whether we \emph{select} node $i \in \mathcal{N}_t$  as a terminal node for a rule in the pruned ensemble. If $z_i^t = 1$, we extract the corresponding rule by traversing tree $t$ from the root to node $i$. For example, in the right panel of Figure~\ref{tree_based_vis.fig}, setting $z_4^t = 1$ extracts the depth 2  rule obtained by traversing the tree from nodes 1 (root) to 2 to 4. Likewise, setting $z_3^t = 1$ extracts the depth 1  rule obtained by traversing the tree from nodes 1 (root) to~3. Continuous variable $w_i^t$ specifies the weight assigned to each extracted rule, and the prediction of our pruned model is then given by the weighted sum of the rule predictions: $\sum_{t \in \mathcal{T}} \sum_{i \in \mathcal{N}_t} {\mathbf{M}_i^t}\,w_i^t$. We use the following problem to prune rules and interaction depth from tree ensembles.
\begin{align} \label{tree_based_problem1.prob}
\min_{\{z_i^t,\,w_i^t\}}\quad \frac{1}{2}\Bigl\| 
 \mathbf{y} - \sum_{t \in \mathcal{T}} \sum_{i \in \mathcal{N}_t} {\mathbf{M}_i^t}\,w_i^t\Bigr\|_2^2 + \frac{1}{2\gamma}\,\sum_{t \in \mathcal{T}} \sum_{i\in \mathcal{N}_t} (w_i^t)^2 \\
\text{s.t.} \quad \sum_{j \in \mathcal{C}_i^t} z_j^t \ \leq \ |\mathcal{C}_i^t| (1 - z_i^t),
\quad \forall \ i \in \mathcal{N}_t, \ t \in \mathcal{T}, \tag{Constraint \ref{tree_based_problem1.prob}a} \\
\sum_{t \in \mathcal{T}} \sum_{i \in \mathcal{N}_t} a_i^t\,z_i^t 
\leq K, \tag{Constraint \ref{tree_based_problem1.prob}b} \\    (1 - z_i^t)\;w_i^t = 0, \quad
\forall \ i \in \mathcal{N}_t,\ t \in \mathcal{T}, \tag{Constraint \ref{tree_based_problem1.prob}c} \\ 
z_i^t \in \{0,1\}, \quad \forall 
\ i \in \mathcal{N}_t, \ t \in \mathcal{T}. \tag{Constraint \ref{tree_based_problem1.prob}d}
\end{align}
The first term in the objective of Problem \eqref{tree_based_problem1.prob} is the quadratic loss function, which captures data fidelity, and the second term is a ridge regularization penalty. \color{black} We include this penalty mainly for computational reasons; however, regularization may also improve the predictive performance of our estimator in low signal-to-noise ratio regimes \citep{mazumder2023subset}, as has been observed for other tree-based models in similar settings \citep{,pmlr-v162-agarwal22b,Agarwal2025Integrating}. \color{black} Parameter $K$  restricts the sum of attributes of the extracted rules and parameter~$\gamma$ controls the ridge penalty; \color{black} we explore the effect of $\gamma$ on our estimator in \S~\ref{tuning_gamma.appx} of the appendix. \color{black} We discuss the constraints of Problem \eqref{tree_based_problem1.prob} below.

\vspace{2mm}
\noindent{\textbf{Constraint (\ref{tree_based_problem1.prob}a), Valid Rule Prunings:}} This constraint ensures that the rule prunings are valid for each tree. If node $i$ in tree $t$ is selected to be the terminal node of an extracted rule, none of its successor  nodes, $j \in \mathcal{C}_i^t$, can be selected to be terminal nodes.

\vspace{2mm}
\noindent{\textbf{Constraint (\ref{tree_based_problem1.prob}b), Attribute-Weighted Budget:}} This constraint controls the sum of attributes of the rules extracted into the pruned model. For example, if we set attribute $a_i^t = 1$ for all $i \in \mathcal{N}_t$ and $t \in \mathcal{T}$, this constraint directly restricts the number of rules selected into the pruned model to be less than or equal to $K$. There are many interesting choices for attribute $a_i^t$, which is prespecified. For example, setting $a_i^t$ to the depth of each node can result in shallower rule sets; we discuss various choices of $a_i^t$ in \S\ref{node_attributes.section}.

\vspace{2mm}
\noindent{\textbf{Constraint (\ref{tree_based_problem1.prob}c), Coupling:}} Finally, constraint (\ref{tree_based_problem1.prob}c) couples binary variable $z_i^t$ with variable $w_i^t$ and constraint  (\ref{tree_based_problem1.prob}d) restricts variables $z_i^t$ to be binary. This ensures that only extracted rules are assigned a non-zero weight.

\vspace{2mm}

\color{black}
Solving Problem~\eqref{tree_based_problem1.prob} yields optimal solutions $(z_i^t)^*$ and $(w_i^t)^*$ for all $i \in \mathcal{N}_t$ and $t \in \mathcal{T}$. The corresponding prediction $\sum_{t \in \mathcal{T}} \sum_{i \in \mathcal{N}_t} {\mathbf{M}_i^t}\, (w_i^t)^*$ gives an estimate of the underlying regression function, under suitable compression constraints. We formally establish this in our theoretical analysis of our estimator in \S\ref{theory.section}.
\color{black}
Problem \eqref{tree_based_problem1.prob} is a mixed-integer program (MIP) with $m$ binary decision variables and $m$ continuous decision variables, where $m$ is the total number of nodes in $\mathcal{T}$. We note that $m$ is typically large. For instance, a boosting ensemble of 1000 depth 3 trees has $m \approx 10^4$; off-the-shelf optimization software such as Gurobi struggle to scale to these problem sizes. Consequently, we present a tailored exact algorithm (\S\ref{optimal_method.section}) to solve Problem \eqref{tree_based_problem1.prob} to optimality, that scales to problem sizes beyond the capabilities of commercial solvers. 

% For an ensemble of 1000 depth 7 decision trees $m$ can be up to the order of magnitude of $\approx 10^6$. 

\subsection{Practical Considerations} 

We  give further details on our estimator, discussing both regularization paths and how different choices of node attributes influence the extracted model.

\begin{figure}[h]
    \centering
    \includegraphics[width=0.7\linewidth]{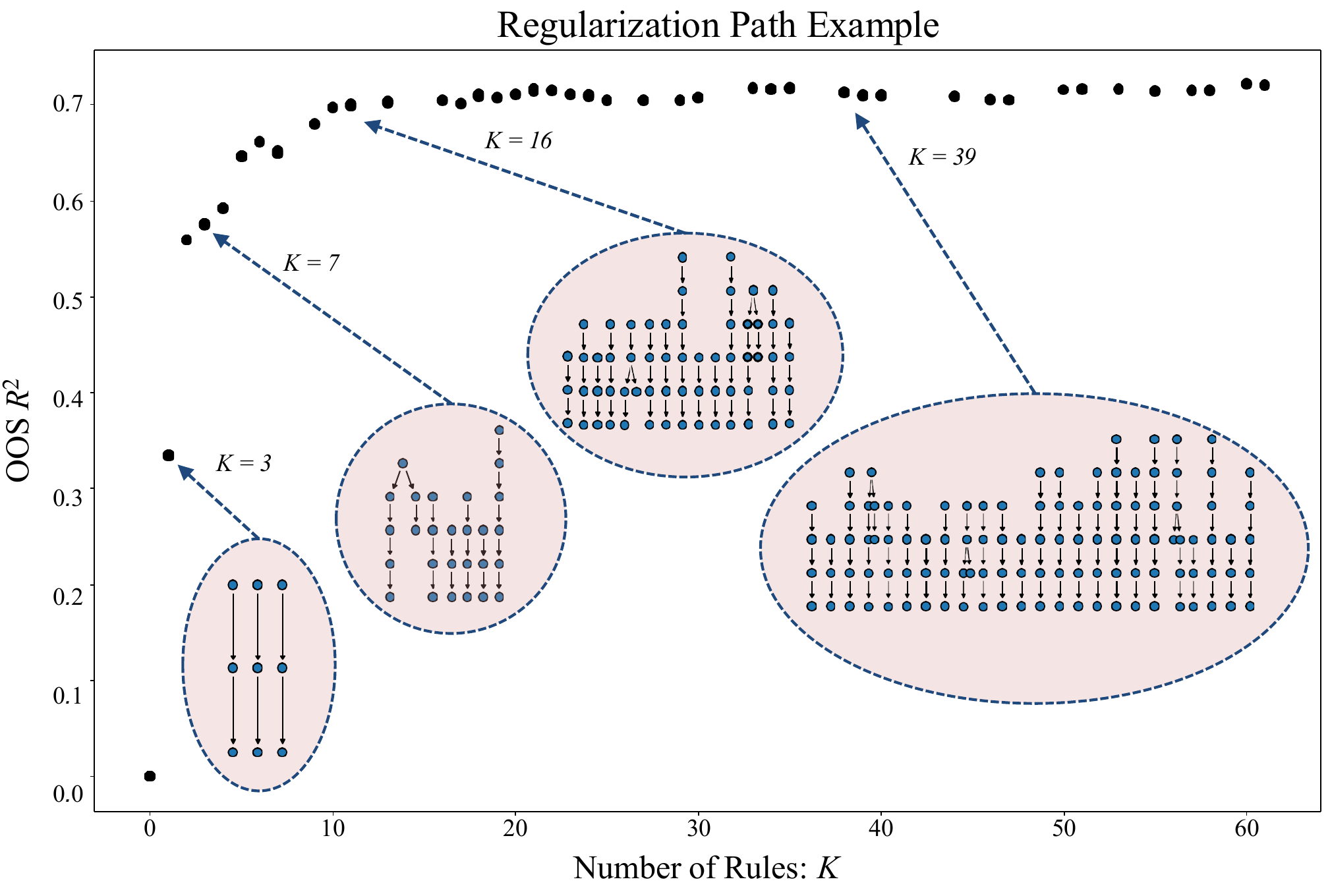}
    \caption{Example regularization path computed by our framework on the \textsc{Wind} example from \cite{haslett1989space}. The path consists of a sequence of solutions that correspond to varying model sizes and predictive performances. }
    \label{rp_example.fig}
\end{figure}

\subsubsection{Regularization Paths} \label{regularization_paths.section}

Solving Problem  ~\eqref{tree_based_problem1.prob}   within our optimization framework prunes an ensemble under a fixed budget $K$, which controls the number of rules extracted. The predictive accuracy of the extracted model can vary significantly with model size, and, in practice, appropriate values of $K$ may not be known a priori. As such, we recommend finding solutions to Problem ~\eqref{tree_based_problem1.prob} across a range of $K$ values to asses the tradeoff between model accuracy and size.

We refer to this process as computing the \textbf{regularization path} in our framework, that is, generating a sequence of solutions with varying model sizes, analogous to the regularization path in sparse regression. In Figure \ref{rp_example.fig}, we show an example of such a path, computed on an example where we prune a boosting ensemble of 500 depth 5 decision trees fit on the \textsc{Wind} dataset from \cite{haslett1989space}. In this example, we set every node attribute $a_i^t = 1, \ \forall \ i \in \mathcal{N}_t,\ t \in \mathcal{T}$, so that $K$ directly controls the number of rules in the extracted model. The horizontal axis in Figure \ref{rp_example.fig} shows $K$, the number of rules extracted, and the vertical axis shows the out-of-sample $R^2$ of the extracted model; each point in the scatter plot shows a solution obtained along the regularization path. The displays in the highlighted circles show snapshots of the extracted models with $K \in \{3,7,16,39\}$ rules.

From this regularization path, we observe that the out-of-sample predictive accuracies of the extracted rules increase sharply with $K$ up to $K = 16$. Beyond this point, further increases in $K$ lead to greater model complexity with minimal gains in accuracy. Computing regularization paths allows practitioners to explore the tradeoff between model size and predictive accuracy and select an appropriate model. A key component of our framework is an efficient procedure for computing regularization paths, which we present in \S\ref{approximate_algorithm.section}.

\subsubsection{Node Attributes} \label{node_attributes.section}

We discuss choices for node attributes $a_i^t, \ \forall 
\ i \in \mathcal{N}_t, \ t \in \mathcal{T}$ and their corresponding effects.

\vspace{2mm}
\noindent\textbf{Rule-Weighting:} In the rule-weighting scheme, we define \( a_i^t = 1 \) for all \( i \in \mathcal{N}_t, \ t \in \mathcal{T} \). With this choice, the attribute-weighted budget constraint~(\ref{tree_based_problem1.prob}b) directly controls the number of rules to extract. This assignment provides the most straightforward interpretation of constraint~(\ref{tree_based_problem1.prob}b) and affords direct control over model size, so we use it as the default setting.

\vspace{2mm}
\noindent\textbf{Depth-Weighting:} In the depth-weighting scheme, we define \( a_i^t = |\mathcal{P}_i^t| \) for all $i \in \mathcal{N}_t, \ t \in \mathcal{T} $, where $ |\mathcal{P}_i^t| $ denotes the number of ancestor nodes of node $i$ in tree $t $. This quantity corresponds to the interaction depth of node $i$ —that is, the number of  conditions traversed from the root to node $i$. Constraint (\ref{tree_based_problem1.prob}b) budgets the sum of interaction depths in the extracted model, which encourages the extraction of shallower rules. This may improve model interpretability and compactness compared to rule-weighting.

\vspace{2mm}
\noindent{\textbf{Feature-Weighting:}} In the feature-weighting scheme, we set $a_i^t$ equal to the number of distinct features that appear in $\mathcal{P}_i^t$  for all \( i \in \mathcal{N}_t, \ t \in \mathcal{T} \), i.e., the number of distinct features that appear in the ancestor nodes of node $i$ in tree $t$. This is the number of feature interactions required to reach node $i$ from the root node. As a result, constraint (\ref{tree_based_problem1.prob}b)  better controls the feature interaction depth and overall feature sparsity of the extracted model.

\begin{figure}[h]
    \centering
    \includegraphics[width=0.75\linewidth]{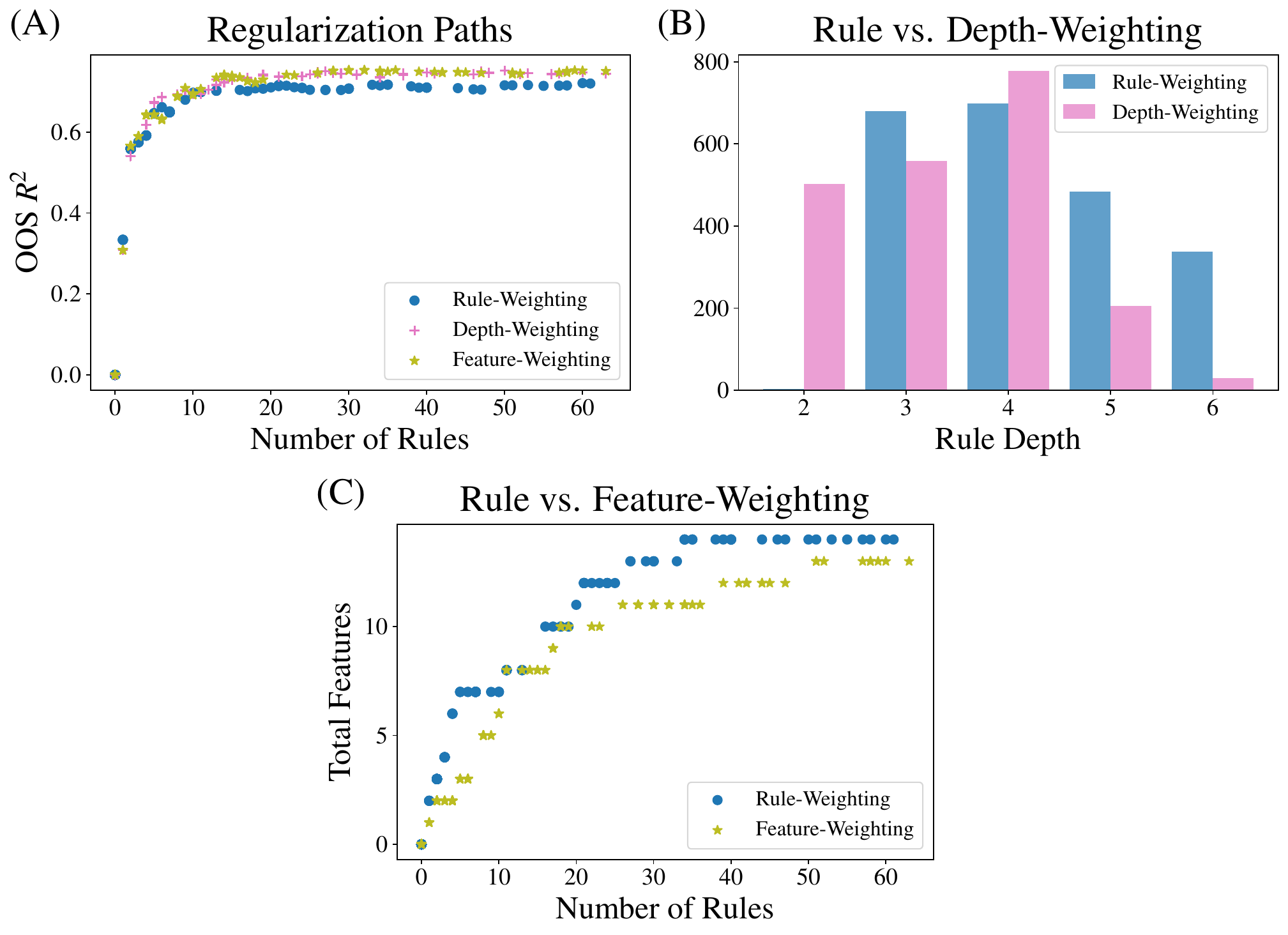}
    \caption{Effects of node attributes. Compared to rule-weighting, depth-weighting extracts shallower rules and feature-weighting promotes feature sparsity.}
    \label{node_attributes.fig}
\end{figure}

\vspace{2mm}
\noindent{\textbf{Comparison of Attribute Weighting Schemes:}} In Figure~\ref{node_attributes.fig}, we compare the effects of rule-weighting, depth-weighting, and feature-weighting on the models extracted by our estimator. We prune the same boosted ensemble from above, of 500 depth 5 trees fit on the \textsc{Wind} dataset \citep{haslett1989space}, and compute regularization paths under each of the three weighting schemes. Panel A of Figure~\ref{node_attributes.fig} displays these paths, where the horizontal axis denotes model size (in number of rules) and the vertical axis shows the out-of-sample $R^2$. While all three schemes yield similar predictive performances, the extracted models exhibit markedly different structural properties. In panel B, we compare the distribution of rule-depths between the models extracted by depth-weighting and rule-weighting, across the entire regularization path. From this plot, it is apparent that depth-weighting extracts shallower rules. In panel C, we compare the total number of features used in the models extracted by feature-weighting and rule-weighting. From this plot, we see that feature-weighting promotes greater feature sparsity across the regularization path. \color{black} We further examine the effects of different attribute-weighting schemes in \S\ref{attribute_weights.appx} of the appendix.
\color{black}

% \vspace{2mm}
% \noindent{\textbf{Equal Tree Depth Constraint}:} One constraint that we may impose if for all rules selected from a decision tree, $\Gamma_t$, to be restricted to the same depth, for all $t \in \mathcal{T}$. This constraint can improve the computation time of approximate algorithms \citep{liu2023forestprune} and encourage shallower rule sets, which may improve interpretability.

% Assume without loss of generality that every decision tree in the ensemble has a maximum depth of $d_{max}$. Let auxiliary binary variable $d_k^t \in \{0,1\}$ for $k \in [d_{max}]$ and $t \in \mathcal{T}$ represent whether all of the rules selected from tree $t$ are restricted to depth $k$. Let $D_k^t$ represent the set of nodes from $\mathcal{N}_t$ that have depth $k$. We add the following constraints to Problem \eqref{tree_based_problem1.prob}.
% \begin{align}
%     \sum_{k=1}^{d_{\max}} d_k^t \leq 1
% \quad
% \forall\,t \in \mathcal{T}, \tag{Equal Tree Depth Constraint a} \\ \sum_{j \in D_k} z_j^t \ \leq \ d_k^t  \lvert D_k \rvert \quad
% \forall \ k\in [d_{\max}], \  t \in \mathcal{T}, \tag{Equal Tree Depth Constraint b} \\
% d_k^t \in \{0,1\} \quad \forall \ k \in [d_{\max}], \ t \in \mathcal{T} \tag{Equal Tree Depth Constraint c}.
% \end{align}
% These constraints ensure that all decision rules selected from decision tree $t \in \mathcal{T}$ are of the same depth.

\section{Optimization Algorithms}
\label{optimization_algorithm_section} In this section, we present our optimal algorithm (\S\ref{optimal_method.section}) to solve instances of Problem \eqref{tree_based_problem1.prob} as well as our approximate algorithm (\S\ref{approximate_algorithm.section}) to efficiently compute regularization paths. Empirically, we observe that solutions obtained by our optimal algorithm perform better in terms of out-of-sample performance compared to those obtained by our approximate algorithm, and, in addition, have theoretical guarantees. Our approximate algorithm, however, scales to larger problem sizes and can efficiently compute solution paths across varying levels of sparsity. While lacking in terms of theoretical guarantees, we observe that the solutions obtained by our approximate algorithm are practically useful.

\subsection{Optimal Algorithm} \label{optimal_method.section}

We present here a tailored exact algorithm\footnote{We use the terms exact algorithm and optimal algorithm interchangeably.} to solve instances of Problem \eqref{tree_based_problem1.prob} to global optimality. Our algorithm can scale to problem sizes beyond the capabilities of off-the-shelf optimization solvers, such as MOSEK and Gurobi,  and can solve instances of Problem \eqref{tree_based_problem1.prob} with thousands of observations and decision variables in seconds. To achieve this speedup, we reformulate Problem \eqref{tree_based_problem1.prob} as a convex integer program (CIP) that depends only on binary decision variables $\mathbf{z}$, halving the number of decision variables. We develop a tailored outer approximation algorithm which exploits problem sparsity and the convexity of the objective function to solve the reformulation to optimality.

\vspace{2mm}
\noindent{\textbf{Problem Reformulation:}} Let $\mathbf{z} \in \{0,1\}^m$ denote the vector containing decision variables $z_i^t$ for all $i\in \mathcal{N}_t$ and $t \in \mathcal{T}$; recall that tree ensemble $\mathcal{T}$ contains $m = \sum_{t \in \mathcal{T}} |\mathcal{N}_t|$ nodes. Let $\mathbb{I}_n$ denote the $n$-dimensional identity matrix and consider the integer program
\begin{align} \label{CIP_reformulation.prob}
\min_{\mathbf{z}}\quad q(\mathbf{z}) = \frac{1}{2} \mathbf{y}^\top \biggl(\mathbb{I}_{n} + \gamma \sum_{t \in \mathcal{T}} \sum_{i\in \mathcal{N}_t} z_{i}^t {\mathbf{M}_i^t}\bigl({\mathbf{M}_i^t}\bigr )^\top\biggr)^{-1}\mathbf{y} \\
\text{s.t.} \quad \sum_{j \in \mathcal{C}_i^t} z_j^t \ \leq \ |\mathcal{C}_i^t| (1-z_i^t),
\quad \forall \ i \in \mathcal{N}_t, \ t \in \mathcal{T}, \tag{Constraint \ref{CIP_reformulation.prob}a} \\
\sum_{t \in \mathcal{T}} \sum_{i \in \mathcal{N}_t} a_i^t\,z_i^t 
\leq K, \tag{Constraint \ref{CIP_reformulation.prob}b} \\  
z_i^t \in \{0,1\}, \quad \forall 
\ i \in \mathcal{N}_t, \ t \in \mathcal{T}. \tag{Constraint \ref{CIP_reformulation.prob}c}
\end{align}
The propositions below state some properties of Problem \eqref{CIP_reformulation.prob}.

\begin{proposition} \label{prop1_opt.prop} Problem \eqref{CIP_reformulation.prob} is an equivalent reformulation of Problem \eqref{tree_based_problem1.prob}.
\end{proposition}

\begin{proposition} \label{prop2_opt.prop} Objective $\mathbf{z} \mapsto q(\mathbf{z})$ where $q(\mathbf{z})$ is defined in \eqref{CIP_reformulation.prob} is convex on $\mathbf{z} \in [0,1]^m$.
\end{proposition}

We prove these propositions in \S\ref{optimal_alg_prop.appx} of the appendix. Problem \eqref{CIP_reformulation.prob} is a convex integer program with $m$ binary decision variables and linear constraints (\ref{CIP_reformulation.prob}a) and (\ref{CIP_reformulation.prob}b). We say that $\mathbf{z} \in \mathbb{C}_3$ if decision vector $\mathbf{z}$ satisfies the 3 constraints in  $\mathbb{C}_3 =\{ (\ref{CIP_reformulation.prob}\text{a}),(\ref{CIP_reformulation.prob}\text{b}), (\ref{CIP_reformulation.prob}\text{c})  \}$ 

\vspace{2mm}
\noindent{\textbf{Outer Approximation Algorithm:}} We present a specialized outer approximation (OA) algorithm to solve Problem \eqref{CIP_reformulation.prob} to optimality \citep{bertsimas2020sparse,duran1986outer}. We treat Problem~\eqref{CIP_reformulation.prob} as an \emph{implicit} reformulation; our outer approximation algorithm only needs to evaluate the objective $q(\mathbf{z})$ and its subgradient $\nabla q(\mathbf{z})$. Because Constraint~(\ref{CIP_reformulation.prob}b) typically yields a sparse~$\mathbf{z}$, we can perform these evaluations efficiently without explicitly constructing Problem~\eqref{CIP_reformulation.prob}---see Appendix~\S\ref{obj_subgrad_optimal.appx} for details.

% Problem \eqref{CIP_reformulation.prob} is an implicit reformulation; our outer approximation algorithm only requires us to evaluate objective $q(\mathbf{z})$ and subgradient $\nabla q(\mathbf{z})$. Since $\mathbf{z}$ is typically sparse, due to Constraint (\ref{CIP_reformulation.prob}b), we can do so efficiently without constructing  Problem \eqref{CIP_reformulation.prob}--we present our procedure in \S\ref{obj_subgrad_optimal.appx} of the appendix.

% and we present a procedure to efficiently do so, without constructing Problem \eqref{CIP_reformulation.prob}, in \S\ref{obj_subgrad_optimal.appx} of the appendix.

 Objective function $q(\mathbf{z})$ in Problem \eqref{CIP_reformulation.prob} is convex, so for all $\mathbf{z}' \in [0,1]^m$ we have: $q(\mathbf{z}' ) \geq q(\mathbf{z}) + \nabla q(\mathbf{z})^{\top}(\mathbf{z}' - \mathbf{z})$. Given a sequence of feasible solutions $\mathbf{z}^{(0)}, \mathbf{z}^{(1)}, \ldots \mathbf{z}^{(h)}$ for Problem \eqref{CIP_reformulation.prob}, we have that: $Q_{LB,h}(\mathbf{z}) := \max_{k=0,..,h} \ q(\mathbf{z}^{(k)}) + \nabla q(\mathbf{z}^{(k)})^{\top}(\mathbf{z}-\mathbf{z}^{(k)}) \ \leq q(\mathbf{z})$, where function $\mathbf{z} \mapsto  Q_{LB,h}(\mathbf{z})$ is a piecewise linear lower bound on function $\mathbf{z} \mapsto  q(\mathbf{z})$. In each iteration $h$ of our outer approximation algorithm, we minimize $ Q_{LB,h}(\mathbf{z})$ with respect to $\mathbf{z} \in \mathbb{C}_3$ to obtain $\mathbf{z}^{(h+1)}$ by solving the problem:
\begin{align} \label{ILP1.prob}
\min_{\mathbf{z},\nu} \quad \nu  \quad \text{s.t.} \quad \mathbf{z} \in \mathbb{C}_3, \quad  
\nu &\geq q(\mathbf{z}^{(k)}) + \nabla q(\mathbf{z}^{(k)})^{\top}(\mathbf{z}-\mathbf{z}^{(k)}) \quad \forall \ k \in \{0, 1, \ldots, h\},
\end{align}
where $\nu$ is a lower bound for the optimal objective value $q(\mathbf{z}^*)$. We present our full outer approximation algorithm in Algorithm \ref{OA_main.alg}.

\begin{algorithm}[h]
    \caption{ Outer Approximation Algorithm for Problem \eqref{CIP_reformulation.prob}} \label{OA_main.alg} 
    \footnotesize
    $\mathbf{z}^{(0)} \leftarrow$ \text{warm start}, $\nu^{(0)} \leftarrow q(\mathbf{z}^{(0)}), \ h \leftarrow 0$

    \While{$q(\mathbf{z}^{(h)}) > \nu^{(h)}$ }{
    add cutting plane $\nu^{(h)} \geq q(\mathbf{z}^{(h)}) + \nabla q(\mathbf{z}^{(h)})^{\top}(\mathbf{z}-\mathbf{z}^{(h)})$ to Problem \eqref{ILP1.prob}
    
    solve Problem \eqref{ILP1.prob} for  $\mathbf{z}^{(h+1)}$
    
    h = h+1
    }
    \textbf{return} $\mathbf{z}^*$
  \end{algorithm}

We terminate our algorithm when our objective value $q(\mathbf{z}^{(h)})$ is equal to the lower bound for the optimal objective value, $\nu^{(h)}$. Since the number of feasible solution in $\mathbb{C}_3$ is finite, Algorithm \ref{OA_main.alg} terminates in a finite number of iterations \citep{duran1986outer} and returns the optimal solution $\mathbf{z}^*$ to Problem~\eqref{CIP_reformulation.prob}. Let $S^*$ represent the set of nonzero indices of $\mathbf{z}^*$, i.e., $S^* = \{ i : z^*_i \neq 0 \}$, and let $\mathcal{M}$ be the matrix formed by taking the vectors 
${\mathbf{M}_i^t}$ as columns for all $i \in \mathcal{N}_t$ and $t \in \mathcal{T}$. Let $\mathcal{M}_{S^*}$ be the sub-matrix of $\mathcal{M}$  formed by selecting the columns corresponding to $S^*$. We have that $\mathbf{w}^* = \bigl(\frac{\mathbbm{I}_{|S^*|}}{\gamma} + (\mathcal{M}^t_{S^*})^\top \mathcal{M}_{S^*} \bigr)^{-1}\mathcal{M}_{S^*}y$, and that $\mathbf{w}^*$,  $\mathbf{z}^*$  is the optimal solution to Problem \eqref{tree_based_problem1.prob}. In each iteration of Algorithm \ref{OA_main.alg} we must solve  Problem \eqref{ILP1.prob}, an integer linear program with $m$ binary decision variables. This can be done efficiently in seconds using off-the-shelf solvers, for values up to $m \approx 10^5$. 

Our tailored exact algorithm solves Problem \eqref{tree_based_problem1.prob} to optimality over three orders of magnitude faster than commercial solvers. For example, consider a boosting ensemble of 100 depth 3 decision trees, fit on the \textsc{socmob} social mobility dataset from \cite{biblarz1993effects}. This dataset contains around 1000 observations and the resulting ensemble contains $m \approx 1500$ nodes. We want to use our estimator to extract a compact rule set of 20 rules.  Algorithm \ref{OA_main.alg} can solve Problem \eqref{tree_based_problem1.prob} to optimality in under \textbf{2 seconds}, while Gurobi fails to find the optimal solution in \textbf{3 hours}. In \S\ref{optimal_timing.section}, we evaluate the computation time of our exact algorithm compared to commercial solvers across a range of problem sizes.

\subsection{Approximate Algorithm} \label{approximate_algorithm.section}
We introduce an approximate algorithm that efficiently finds high-quality solutions to Problem \eqref{tree_based_problem1.prob} and is tailored for computing regularization paths. Motivated by the success of approximate algorithms for penalized formulations of high-dimensional sparse regression \citep{hazimeh2023grouped}, we consider this penalized reformulation of 
 Problem~\eqref{tree_based_problem1.prob}:\begin{align} \label{tree_based_penalized.prob}
\min_{\{z_i^t,\,w_i^t\}}\quad \frac{1}{2}\Bigl\| 
  \mathbf{y} - \sum_{t \in \mathcal{T}} \sum_{i \in \mathcal{N}_t} {\mathbf{M}_i^t}\,w_i^t\Bigr\|_2^2 + \frac{1}{2\gamma}\,\sum_{t \in \mathcal{T}} \sum_{i\in \mathcal{N}_t} (w_i^t)^2 + \lambda \sum_{t \in \mathcal{T}} \sum_{i \in \mathcal{N}_t} a_i^t\,z_i^t  \\
\text{s.t.} \quad \sum_{j \in \mathcal{C}_i^t} z_j^t \ \leq \ |\mathcal{C}_i^t| (1-z_i^t),  \quad (1 - z_i^t)\;w_i^t = 0, \quad  z_i^t \in \{0,1\}, \quad \forall 
\ i \in \mathcal{N}_t, \ t \in \mathcal{T}, \notag
\end{align} where parameter $\lambda$ controls the size of the extracted model. Our approximate algorithm  finds solutions to Problem \eqref{tree_based_penalized.prob} across a range of $\lambda$ values to compute regularization paths. As an aside, in \S\ref{exploring_convex.appx} of the appendix we explore the convex relaxation of Problem~\eqref{tree_based_penalized.prob} as a new estimator of independent interest.

% \subsubsection{Block Coordinate Descent Algorithm} 

Note that the first term in the objective function of Problem \eqref{tree_based_penalized.prob} is smooth with respect to $w$, the second and third terms are separable with respect to trees $t \in \mathcal{T}$. The constraints are  also all separable with respect to trees $t \in \mathcal{T}$. 
% Motivated by the success of block coordinate descent (BCD) algorithms for high-dimensional regression \citep{hazimeh2023grouped},
We apply the following block coordinate descent (BCD) algorithm to find high-quality solutions to Problem \eqref{tree_based_penalized.prob}. For a fixed block $t \in \mathcal{T}$, let $\delta$ represent the remaining blocks, i.e., $\delta = \mathcal{T} \setminus t$, and let residual vector $\mathbf{r} = \mathbf{y} - \sum_{t \in \delta} \sum_{i \in \mathcal{N}_t} {\mathbf{M}_i^t}\,w_i^t $. Each update of our BCD algorithm solves the MIP: \begin{align} \label{blockupdate.prob}
\min_{\mathbf{z}^t,\,\mathbf{w}^t}\quad \frac{1}{2}\Bigl\| 
  \mathbf{r} -  \sum_{i \in \mathcal{N}_t} {\mathbf{M}_i^t}\,w_i^t\Bigr\|_2^2 + \frac{1}{2\gamma}\, \sum_{i\in \mathcal{N}_t} (w_i^t)^2 + \lambda  \sum_{i \in \mathcal{N}_t} a_i^t\,z_i^t  \\
\text{s.t.} \quad \sum_{j \in \mathcal{C}_i^t} z_j^t \ \leq \ |\mathcal{C}_i^t| (1-z_i^t),
\quad
   (1 - z_i^t)\;w_i^t = 0, \quad 
z_i^t \in \{0,1\}, \quad \forall 
\ i \in \mathcal{N}_t, \notag
\end{align}
where the decision vectors $\mathbf{z}^t \in \{0,1\}^{|\mathcal{N}_t|}$ and $\mathbf{w}^t \in \mathbb{R}^{|\mathcal{N}_t|}$ represent the stacked decision variables $z_i^t$ and $w_i^t$ for all $i \in \mathcal{N}_t$, i.e., the decision variables associated with tree $t \in \mathcal{T}$.

\begin{algorithm}[h] 
    \caption{ Cyclic Block Coordinate Descent for Problem \eqref{tree_based_penalized.prob}} \label{cbcd.alg} 
    \footnotesize
$\mathbf{z},\mathbf{w} \leftarrow$ warm start

\Repeat{until objective no longer improves}{
    \For{$t \in \mathcal{T}$}{
    solve MIP \eqref{blockupdate.prob} for $z_i^t, \ w_i^t \quad \forall i \in \mathcal{N}_t$
    }}
    \textbf{return} $\mathbf{z}, \mathbf{w}$
  \end{algorithm}

We summarize our proposed cyclic BCD procedure in Algorithm \ref{cbcd.alg}. Algorithm \ref{cbcd.alg} returns a sequence of non-increasing objective values for Problem \eqref{tree_based_penalized.prob}, and we terminate our algorithm when the objective no longer improves. Our CBCD algorithm depends on our ability to solve Problem \eqref{blockupdate.prob}, a MIP with $|\mathcal{N}_t|$ binary and $|\mathcal{N}_t|$ continuous decision variables, efficiently. In the next section, we present a procedure to do so and introduce computational enhancements that allow Algorithm \eqref{cbcd.alg} to efficiently compute regularization paths.

% We present a procedure for doing so in \S\ref{efficient_block_update.section}, and in \S\ref{speedups.section}, we present computational speedups that enable the efficient application of Algorithm~\eqref{cbcd.alg} across a range of $\lambda$ values to compute regularization paths.

\subsubsection{Computational Enhancements} \label{computation_enhancements.section}

In \S\ref{approx_timing.section}, we evaluate the impact of these enhancements on the computation time of our approximate algorithm.

\vspace{2mm}
\noindent{\textbf{Efficient Block Update:}} 
We present an outer approximation procedure similar to the one discussed in \S\ref{optimal_method.section} to solve Problem \eqref{blockupdate.prob} efficiently. Consider the following IP:
\begin{align} \label{blockupdate_CIP.prob}
  \min_{\mathbf{z}^t}\quad q_{\mathbf{r}}(\mathbf{z}^t) = \frac{1}{2} \mathbf{r}^\top \biggl(\mathbb{I}_{n} + \gamma \sum_{i\in \mathcal{N}_t} z_{i}^t {\mathbf{M}_i^t}\bigl({\mathbf{M}_i^t}\bigr )^\top\biggr)^{-1} \mathbf{r} + \lambda  \sum_{i \in \mathcal{N}_t} a_i^t\,z_i^t\\
\text{s.t.} \quad \sum_{j \in \mathcal{C}_i^t} z_j^t \ \leq \ |\mathcal{C}_i^t| (1-z_i^t),
\quad z_i^t \in \{0,1\}, \quad \forall 
\ i \in \mathcal{N}_t. \notag
\end{align}
The following proposition establishes some properties of Problem \eqref{blockupdate_CIP.prob}.
\begin{proposition} \label{approx_alg_CBCD.prop}
    Problem \eqref{blockupdate_CIP.prob} is an equivalent reformulation of MIP Problem \eqref{blockupdate.prob} and objective function $\mathbf{z}^t \mapsto q_{\mathbf{r}}(\mathbf{z}^t)$ is convex on $\mathbf{z}^t \in [0,1]^{|\mathcal{N}_t|}$.
\end{proposition}
We prove this proposition and present an efficient procedure to evaluate $q_{\mathbf{r}}(\mathbf{z}^t)$ and  $\nabla q_{\mathbf{r}}(\mathbf{z}^t)$ in $\S\ref{approx_alg_prop.appx}$ of the appendix. 
% As an aside, let $\mathcal{M}^t$ denote the $n \times |\mathcal{N}_t|$ matrix formed by taking vectors ${\mathbf{M}_i^t}$ as columns, for all $i \in \mathcal{N}_t$. Let $S$ represent the set of indices of $\mathbf{z}^t$ such that $\mathbf{z}^t \neq 0$.
% , i.e., $S = \{ i : z^t_i \neq 0 \}$,  and let $\mathcal{M}^t_S$ be the sub-matrix of $\mathcal{M}^t$ formed by selecting the columns corresponding to indices in $S$. 
% Our procedure to  evaluate $q_{\mathbf{r}}(\mathbf{z}^t)$ and  $\nabla q_{\mathbf{r}}(\mathbf{z}^t)$ requires $\bigl(\frac{\mathbbm{I}_{|S|}}{\gamma} + (\mathcal{M}^t_S)^\top \mathcal{M}^t_S \bigr)^{-1}$, which can be obtained efficiently since $\mathbf{z}^t$ is typically sparse. 
In each iteration $h$ of our outer approximation algorithm we solve this ILP, using off-the-shelf solvers, to obtain $(\mathbf{z}^t)^{(h+1)}$:
\begin{align}\label{ILP2.prob}
    \min_{\mathbf{z}^t, \ \nu} \quad \nu \quad \text{s.t.} \quad & \sum_{j \in \mathcal{C}_i^t} z_j^t \ \leq \ |\mathcal{C}_i^t| (1-z_i^t), \quad z_i^t \in \{0,1\}, \quad \forall \ i \in \mathcal{N}_t,\\
    & \nu \geq q_{\mathbf{r}}\bigl((\mathbf{z}^t)^{(k)}\bigr) + \nabla q_{\mathbf{r}}\bigl((\mathbf{z}^t)^{(k)}\bigr)^{\top}\bigl(\mathbf{z}-(\mathbf{z}^t)^{(k)}\bigr) \quad \forall \ k \in \{0, \ldots, h\}. \notag
\end{align}
\begin{algorithm}[h]
    \caption{ Outer Approximation Algorithm for Problem \eqref{blockupdate_CIP.prob}} \label{OA_sub.alg} 
    \footnotesize
    $(\mathbf{z}^t)^{(0)} \leftarrow$ \text{warm start},  $\nu^{(0)} \leftarrow q_{\mathbf{r}}\bigl((\mathbf{z}^t)^{(0)}\bigr), \ h \leftarrow 0$

    \While{$q_{\mathbf{r}}\bigl((\mathbf{z}^t)^{(h)}\bigr) > \nu^{(h)}$ }{
    add cutting plane $\nu^{(h)} \geq q_{\mathbf{r}}\bigl((\mathbf{z}^t)^{(h)}\bigr) + \nabla q_{\mathbf{r}}\bigl((\mathbf{z}^t)^{(h)}\bigr)^{\top}\bigl(\mathbf{z}-(\mathbf{z}^t)^{(h)}\bigr)$ to Problem \eqref{ILP2.prob}
    
    solve Problem \eqref{ILP2.prob} for  $(\mathbf{z}^t)^{(h+1)}$
    
    h = h+1
    }
    \textbf{return} $(\mathbf{z}^t)^*$
  \end{algorithm}
  
We present our outer approximation procedure in Algorithm \ref{OA_sub.alg}, which converges in a finite number of iterations \citep{fletcher1994solving} and returns the optimal solution $(\mathbf{z}^t)^*$ to Problem \eqref{blockupdate_CIP.prob}. Let $S^*$ correspond to the set of indices where $(\mathbf{z}^t)^* \neq0$, let $\mathcal{M}^t$ denote the $n \times |\mathcal{N}_t|$ matrix formed by taking vectors ${\mathbf{M}_i^t}$ as columns for all $i \in \mathcal{N}_t$, and let $\mathcal{M}^t_{S^*}$ be the sub-matrix of $\mathcal{M}^t$ formed by selecting the columns corresponding to indices in $S^*$.  We have that $(\mathbf{w}^t)^* = \bigl(\frac{\mathbbm{I}_{|S^*|}}{\gamma} + (\mathcal{M}^t_{S^*})^\top \mathcal{M}^t_{S^*} \bigr)^{-1}\mathcal{M}^t_{S^*}r$, and that $(\mathbf{w}^t)^*$,  $(\mathbf{z}^t)^*$  is the optimal solution to Problem \eqref{blockupdate.prob}. Below, we present several  enhancements to speed up our CBCD Algorithm~\ref{cbcd.alg}, along with a procedure to efficiently compute regularization paths.

\begin{figure}[h]
    \centering
    \includegraphics[width=.85\linewidth]{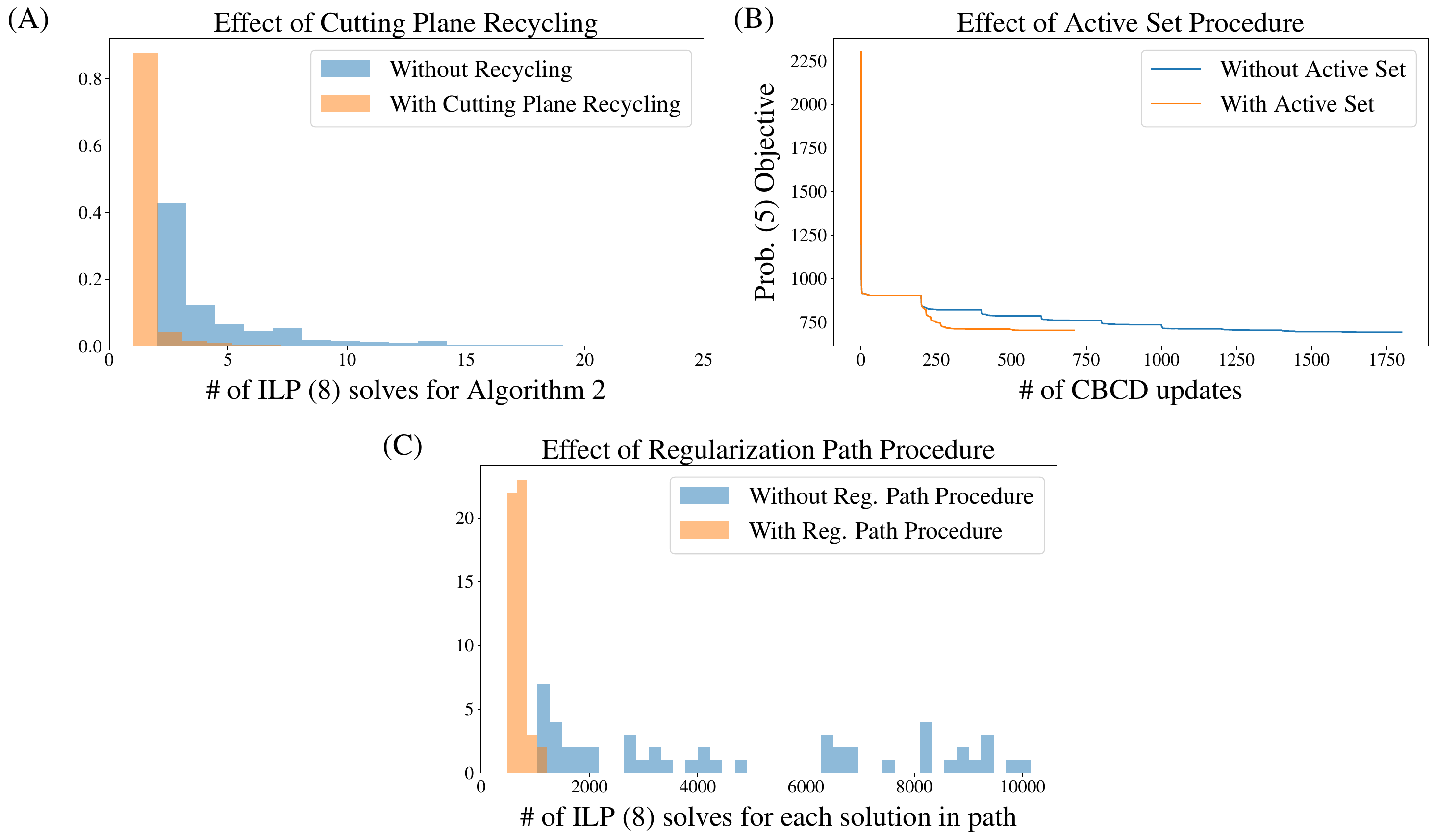}
    \caption{\textbf{Panel A:} Cutting plane recycling reduces the number of ILP solves required for Algorithm \ref{OA_sub.alg}. \textbf{Panel B:} Active set optimization reduces the number of CBCD updates required in Algorithm \ref{cbcd.alg}. \textbf{Panel C:} Our efficient regularization path procedure reduces the number of ILP solves required for each solution in the path. These visualizations are from an application of our framework on the \textsc{wind} dataset from \cite{haslett1989space}.}
    \label{CBCD_speedups.fig}
\end{figure}

\vspace{2mm}
\noindent{\textbf{Cutting Plane Recycling:}} In Algorithm \ref{cbcd.alg} we update each block, tree $t \in \mathcal{T}$, multiple times. Only the residual vector $\mathbf{r}$ changes between updates; the constraints in Problem \eqref{blockupdate_CIP.prob} remain the same. As such, we can recycle cutting planes across subsequent block updates.

Consider two subsequent block updates of tree $t \in \mathcal{T}$. Let $\mathbf{z}^t$ be a feasible solution for Problem \eqref{blockupdate_CIP.prob} obtained while applying Algorithm \ref{OA_sub.alg} during the first block update. This feasible solution corresponds to cutting plane $\nu \geq q_{\mathbf{r}}(\mathbf{z}^t)+ \nabla q_{\mathbf{r}}(\mathbf{z}^t)^{\top}(\mathbf{z}-\mathbf{z}^t)$. Solution $\mathbf{z}^t$ is also feasible for Problem \eqref{blockupdate_CIP.prob} during the subsequent second update, with residual vector $\mathbf{r}'$. Importantly, we can obtain $q_{\mathbf{r}'}(\mathbf{z}^t)$ and $\nabla q_{\mathbf{r}'}(\mathbf{z}^t)$ efficiently  by caching the LU decomposition of the regularized Hessian matrix required to compute $q_{\mathbf{r}}(\mathbf{z}^t)$ and $\nabla q_{\mathbf{r}'}(\mathbf{z}^t)$ during the first update.  As a result, we obtain cutting plane: $\nu^{(h)} \geq q_{\mathbf{r}'}(\mathbf{z}^t)+ \nabla q_{\mathbf{r}'}(\mathbf{z}^t)^{\top}(\mathbf{z}-\mathbf{z}^t)$ for little computational cost during the second update.  We discuss the full technical details of this procedure in \S\ref{cutting_plane_recycle.appx} of the appendix. Our cutting plane recycling procedure significantly reduces the number of OA iterations, solves of ILP \eqref{ILP2.prob}, required to solve Problem \eqref{blockupdate_CIP.prob} to optimality each time a block is updated in Algorithm~\eqref{cbcd.alg}. We show an example in panel A of Figure \ref{CBCD_speedups.fig}; the histogram shows the distribution of the number of ILP solves needed in Algorithm \eqref{cbcd.alg}, comparing runs with and without our cutting plane recycling procedure.

\vspace{2mm}
\noindent{\textbf{Active Set Optimization:}} We also use this active set procedure to reduce the number of CBCD update performed in Algorithm \ref{cbcd.alg}. We first conduct block updates across a few sweeps of trees $t \in \mathcal{T}$ in order to identify an active set $\mathcal{A}$ of blocks such that $\mathbf{z}^t \neq \mathbf{0}, \ \forall \  t \in \mathcal{A}$. On this active set, we cyclically update blocks  $t \in \mathcal{A}$  until the objective no longer improves. We then update all blocks $t \in \mathcal{T}$  to update our active set, terminating our algorithm when the active set no longer changes. We show a visualization of the effect of this procedure in panel B of Figure \ref{CBCD_speedups.fig}. The plot shows the objective function of Problem \eqref{tree_based_penalized.prob} during our CBCD algorithm, comparing runs with and without our active set procedure.

\vspace{2mm}
\noindent{\textbf{Efficient Regularization Paths:}} Finally, we present a procedure to efficiently compute regularization paths, i.e., sequences of solutions for Problem \eqref{tree_based_penalized.prob} with varying sparsities, computed across a range of $\lambda$ values. Given a descending sequence of $\lambda$ values, we apply Algorithm \ref{cbcd.alg} repeatedly. We use the solution obtained by each application of Algorithm \ref{cbcd.alg} to warm start the subsequent solve. We also apply the cutting plane recycling method discussed in the section above each time we update tree $t \in \mathcal{T}$. These procedures reduce the number of CBCD updates and ILP updates required to compute regularization paths. We show a visualization of the effect of this procedure in panel C in Figure \ref{CBCD_speedups.fig}; the histograms show the distribution of the number of ILP solves required to compute each solution along the regularization path, comparing runs with and without our  method.

\subsubsection{Discussion} With our computational enhancements (\S\ref{computation_enhancements.section}), our CBCD algorithm scales to  large problem sizes.
% significantly beyond the capabilities of our optimal algorithm \footnote{We note that our specialized optimal algorithm scales to problem sizes significantly beyond the capabilities of commercial optimization solvers such as Gurobi}. 
For example, consider a boosting ensemble of 500 depth 7 trees, fit on the \textsc{wind} dataset from \cite{haslett1989space}, which has 6500 observations---the resulting ensemble has over 95000 nodes. Our CBCD algorithm can find high-quality solutions in less than \textbf{30 seconds}, and, moreover, can compute the regularization path of solutions of varying sparsities in less than 4 minutes. In \S\ref{approx_timing.section}, we evaluate the computation time of our approximate algorithm for computing regularization paths. \color{black} Importantly, we observe that our approximate  algorithm obtains high-quality solutions along the regularization path which, although worse than those obtained by our exact algorithm, remain practically useful in real-world settings. We explore this in detail in \S\ref{opt_v_approx.appx} of the appendix.
\color{black}

% While the relax-and-round Algorithm~\eqref{LRround.alg} is computationally efficient, we do not recommend using it as a standalone method to find solutions to Problem~\eqref{tree_based_problem1.prob}. Instead, we recommend using it either to warm-start the CBCD algorithm or to quickly evaluate hyperparameter choices during model tuning.

% Importantly, the efficient block update procedure presented in \S\ref{efficient_block_update.section} and the computational speedups presented in \S\ref{speedups.section} make CBCD a practical algorithm, despite the NP-hard nature of each block update. We show the impacts of these enhancements through an ablation study in \S\ref{approx_timing.section}. As such, we typically use CBCD as our main approximate algorithm. We note, however, that our relax-and-round procedures can be used to efficiently find solutions to warm-start our CBCD algorithm and to evaluate different hyperparameters for model tuning.

\section{Statistical Theory}
\label{theory.section}

In this section, we derive non-asymptotic prediction error bounds for our estimator, illustrating that in large samples, our approach performs similarly to the oracle that chooses the best linear combination of the rules in the ensemble. The proofs of all the results in this section are provided in Appendix~\ref{proofs.theory.section}. We note that we are not aware of other existing prediction error bounds for ensemble pruning algorithms.

\noindent\textbf{Introduction}. We assume the model $y_i=f^*(\bx_i)+\epsilon_i$ for our training data~$\{(y_i,\bx_i)\}_{i=1}^n$, where~$n$ is the sample size, $p$ is the number of features/predictors, $f^*:\mathbb{R}^p\rightarrow\mathbb{R}$ is the true regression function, $\bx_i$ is the feature vector for the $i$-th observation, and~$\epsilon_i$ are independent $N(0,\sigma^2)$ with $\sigma>0$. For concreteness, we focus on the case of deterministic design.

For our theoretical analysis, we assume that we have a binary tree ensemble constructed using the training data. We apply our proposed approach on the \textit{same} training data and prune the ensemble by solving Problem~\eqref{tree_based_problem1.prob}. We do not make assumptions about the specific method that was used to construct the tree ensemble. For example, the ensemble could come from a gradient boosting or a random forest procedure.

Recall that the prediction vector of a solution to Problem~\eqref{tree_based_problem1.prob} is given by a linear combination $\sum_{t \in \mathcal{T}} \sum_{j \in \mathcal{N}_t} {\mathbf{M_j^t}}\,w_j^t$. The elements of~${\mathbf{M_j^t}}$ corresponding to the observations in node~$j$ of tree~$t$ are equal to the average response at the node, and the remaining elements are zero. Coefficients~$w_j^t$ uniquely determine the corresponding regression function ${f:\mathbb{R}^p\rightarrow\mathbb{R}}$, obtained by replacing each~${\mathbf{M_j^t}}$ with the underlying rescaled indicator function\footnote{The indicator is used to specify the node, and the multiplier equals the average observed response for the node.}, or rules, which we denote by $r_j^t$. Thus, each candidate regression function is given by $f(\bx)=\sum_{t \in \mathcal{T}} \sum_{j \in \mathcal{N}_t} r_j^t(\bx)\,w_j^t$.

\noindent\textbf{Notation}. Problem~\eqref{tree_based_problem1.prob} imposes constraints on~$f$, specifying the class of candidate regression functions. Depending on the choice of the attributes $a_j^t$ used in the constraints, we denote the corresponding functional classes as $\mathcal{C}_1(K)$, $\mathcal{C}_2(K)$, or $\mathcal{C}_3(K)$. We focus on the  three attribute choices described in Section~\ref{node_attributes.section}, i.e., 
(1) $a_j^t\equiv1$; (2) $a_j^t\,=$ depth of node~$j$ in tree~$t$; and (3) $a_j^t\,=$ number of distinct features used in rule $r_j^t$. %More formally,
%\begin{equation*}
%    \mathcal{C}_k(K)=\Big\{f:\mathbb{R}^p\rightarrow\mathbb{R},\;\text{s.t.}\;f(\bx)=\sum_{t \in \mathcal{T}} \sum_{i \in \mathcal{N}_t} r_i^t(\bx)\,w_i^t \;\text{for}\; w_i^t\in\mathbb{R} \;\text{for}\;t \in \mathcal{T},\;i \in \mathcal{N}_t  \Big\}
%\end{equation*}
%for $k\in\{1,2,3\}$.
We observe the following relationship between the functional classes: $\mathcal{C}_1(K)\supseteq\mathcal{C}_3(K)\supseteq\mathcal{C}_2(K)$. We write~$\widehat{f}_1$, $\widehat{f}_2$ and~$\widehat{f}_3$ for the estimated regression functions corresponding to each of the three classes and assume that these estimates are obtained by solving Problem~\eqref{tree_based_problem1.prob} to global optimality. 

We write~$\lesssim$ and~$\gtrsim$ to indicate that inequalities~$\lesssim$ and~$\gtrsim$, respectively, hold up to positive \textit{universal} multiplicative factors. We let~$D$ denote the maximum depth of the trees in the ensemble. In practice, $D$ is often specified during the construction of the ensemble and used as the depth of every tree. We note that we allow both $p$ and $D$ to depend on~$n$. 

Given a candidate regression function $f(\bx)=\sum_{t \in \mathcal{T}} \sum_{j \in \mathcal{N}_t} r_j^t(\bx)\,w_j^t$, we write $\|w(f)\|$ for the Euclidean norm of the coefficient vector, i.e., $\|w(f)\|^2=\sum_{t \in \mathcal{T}} \sum_{j \in \mathcal{N}_t} (w_j^t)^2$. Similarly, we define $\|w(f)\|_{\infty}=\max_{t \in \mathcal{T}, j \in \mathcal{N}_t}|w_j^t|$.  We also write~$\|f\|_n$ for the empirical $L_2$ norm of~$f$, i.e., $\|f\|_n^2=\sum_{i=1}^n f^2(\bx_i)/n$. 

\noindent\textbf{Results}. The following theorem establishes non-asymptotic bounds on the prediction error, $\|\widehat{f}_k-f^*\|_n^2$, for $k=1,2,3$, comparing the predictive performance of our approach with that of an oracle, which chooses the best possible (data-dependent) coefficients~$w_j^t$ in the linear combination $\sum_{t \in \mathcal{T}} \sum_{j \in \mathcal{N}_t}r_j^t(\bx)\,w_j^t$ under the imposed constraint on the attributes. We treat the attribute budget~$K$ as a user-supplied value rather than a parameter that we tune.  We emphasize that we do not impose any assumptions on the method that was used to construct our tree ensemble.

\begin{theorem}
\label{thm.rates}
\color{black} Let $r_1 = {KD\log(enp)}/{n}$ and $r_2 = r_3 = {K\log(enp)}/{n}$. \color{black} Then, \color{black} for each $k\in\{1,2,3\}$ \color{black} and $\delta_0\in(0,1)$, inequality
\begin{equation*}
\|\widehat{f}_k-f^*\|_n^2\lesssim \inf_{f_k\in \mathcal{C}_k(K)} \Big\{\|f_k-f^*\|_n^2 +\tfrac1{n\gamma}\|w(f_k)\|^2\Big\}+ \sigma^2r_k +\sigma^2\frac{\log(1/\delta_0)}{n}
\end{equation*}
holds with probability at least $1-\delta_0$.
\end{theorem}
\begin{remark}
   It is interesting to note that there is no obvious ``true'' linear combination of the rules~$r_j^t(\bx)$, which would serve as a natural target for our estimators. This phenomenon occurs because the rules are constructed using the training data via ensemble methods such as gradient boosting or random forests. We overcome this issue by comparing our estimators with the \textbf{best} possible linear combination of the rules $r_j^t(\bx)$.
\end{remark}
\begin{remark}
    We note that the functions in classes $\mathcal{C}_k(K)$ are random, because the rules~$r_i^t(\bx)$ are constructed from the training data. Thus, the $\inf$ on the right-hand sides of the bounds in Theorem~\ref{thm.rates} is computed pointwise, i.e., for every training sample. In other words, the oracle coefficients in the linear combination of the rules depend on the tree ensemble.
\end{remark}

%in the case where a good approximation to~$f^*$ can be achieved via an oracle linear combination of the rules.

\color{black} For a simple illustration of the prediction error rates in Theorem~\ref{thm.rates}, consider the special case $\tfrac1{\gamma}=0$ where no ridge penalty is imposed. \color{black} Let $\delta_1 = (enp)^{-KD}$, and $\delta_2 = \delta_3 = (enp)^{-K}$. Theorem~\ref{thm.rates} implies that $\|\widehat{f}_k-f^*\|_n^2\lesssim \sigma^2r_k$ with probability at least $1-\delta_k$, which is high as long as either~$n$ or~$p$ is large. In particular, $n\vee p\rightarrow\infty$ implies $\delta_k\rightarrow0$ for each $k\in\{1,2,3\}$.
We note that $r_1\ge r_2 = r_3$ and $r_1/r_2=D$. Consequently, both the depth-weighting scheme and the feature-weighting scheme may significantly improve over the rule-weighting scheme when the trees in the ensemble are deep and the true regression function can be well approximated using the allocated attribute budget. This observation is consistent with the results in Figure~\ref{node_attributes.fig}.

The error rates in Theorem~\ref{thm.rates} can also apply to approximate solutions obtained after an early termination of the MIP solver. Upon termination, the solver provides the upper and lower bounds on the value of the objective function. We denote these bounds by $UB$ and $LB$, respectively, and write $\tau=(UB-LB)/UB$ for the corresponding optimality gap. Let~$\widetilde{f}_k$, with $k\in\{1,2,3\}$, denote the approximate solutions produced by the solver for Problem~\eqref{tree_based_problem1.prob} with our three attribute choices, respectively. Recall the earlier definitions of~$r_k$ and~$\delta_k$. The following result demonstrates that the bounds in Theorem~\ref{thm.rates} hold for the approximate solution~$\widetilde{f}_k$ when $\tau$ is bounded away from one and $\tau\lesssim r_k$.

\begin{corollary}
\label{cor.approx}
Suppose that $\tau\le 1-a$ for some positive universal constant~$a$. Then, for each $k\in\{1,2,3\}$, inequality
\begin{equation*}
\|\widetilde{f}_k-f^*\|_n^2\lesssim \inf_{\color{black} f_k\in \mathcal{C}_k(K)} \Big\{\|f_k-f^*\|_n^2 \color{black}+\tfrac1{n\gamma}\|w(f_k)\|^2\Big\}+ \color{black} \sigma^2r_k + \sigma^2\tau
\end{equation*}
holds with probability at least $1-\delta_k$.
    
\end{corollary}

\section{Experiments and Case Study} \label{experiments.section}
We present in this section experiments to evaluate our proposed algorithm.

\subsection{Computation Time} \label{optimal_timing.section} We first evaluate the computation time of our proposed exact algorithm (Algorithm \ref{OA_main.alg}) against two state-of-the art commercial solvers, Gurobi and MOSEK, by solving various size instances of Problem \eqref{tree_based_problem1.prob}. For each instance, we extract a set of $K = 20$ decision rules, a typical use case for our framework. The results of this timing experiment are presented in Table \ref{optimal_method_time.table}. We show the computation time of each method averaged over 5 runs, with standard errors shown in parentheses. The left two columns of the table report ensemble size, specified by the number of trees and depth, and the size of Problem \eqref{tree_based_problem1.prob}, specified by the number of observations and binary decision variables. Cells highlighted in red indicate instances where the method fails to find the optimal solution within the 3-hour time limit. 

\begin{table}[h]\centering \scalebox{0.7}{
\begin{tabular}{|c|c|c|c|c|}
\hline
\textbf{\begin{tabular}[c]{@{}c@{}}Ensemble Size\\ \# Trees / Depth\end{tabular}} & \textbf{\begin{tabular}[c]{@{}c@{}}Problem Size\\ Obs. / Binary Vars.\end{tabular}} & \textbf{\begin{tabular}[c]{@{}c@{}}Proposed Exact \\ Algorithm\end{tabular}} & \textbf{Gurobi}                         & \textbf{MOSEK}                          \\ \hline
\textbf{100 / 3}                                                                  & \textbf{1000 / 1500}                                                                & 1.69s (0.04)                                                                 & \cellcolor[HTML]{FFCCC9}\textgreater 3h & \cellcolor[HTML]{FFCCC9}\textgreater 3h \\ \hline
\textbf{100 / 5}                                                                  & \textbf{1000 / 5000}                                                                & 7.64s (0.14)                                                                 & \cellcolor[HTML]{FFCCC9}\textgreater 3h & \cellcolor[HTML]{FFCCC9}\textgreater 3h \\ \hline
\textbf{500 / 4}                                                                  & \textbf{1000 / 10000}                                                               & 14.2s (0.18)                                                                 & \cellcolor[HTML]{FFCCC9}\textgreater 3h & \cellcolor[HTML]{FFCCC9}\textgreater 3h \\ \hline
\textbf{100 / 3}                                                                  & \textbf{3000 / 1500}                                                                & 8.8s (0.15)                                                                  & \cellcolor[HTML]{FFCCC9}\textgreater 3h & \cellcolor[HTML]{FFCCC9}\textgreater 3h \\ \hline
\textbf{100 / 5}                                                                  & \textbf{3000 / 5000}                                                                & 18.3s (0.18)                                                                 & \cellcolor[HTML]{FFCCC9}\textgreater 3h & \cellcolor[HTML]{FFCCC9}\textgreater 3h \\ \hline
\textbf{500 / 4}                                                                  & \textbf{3000 / 10000}                                                               & 38.1s (0.42)                                                                 & \cellcolor[HTML]{FFCCC9}\textgreater 3h & \cellcolor[HTML]{FFCCC9}\textgreater 3h \\ \hline
\textbf{100 / 3}                                                                  & \textbf{5000 / 1500}                                                                & 164.8s (1.23)                                                                & \cellcolor[HTML]{FFCCC9}\textgreater 3h & \cellcolor[HTML]{FFCCC9}\textgreater 3h \\ \hline
\textbf{100 / 5}                                                                  & \textbf{5000 / 5000}                                                                & 247.49s (2.45)                                                               & \cellcolor[HTML]{FFCCC9}\textgreater 3h & \cellcolor[HTML]{FFCCC9}\textgreater 3h \\ \hline
\textbf{500 / 4}                                                                  & \textbf{5000 / 10000}                                                               & 10m 21s (10.41)                                                              & \cellcolor[HTML]{FFCCC9}\textgreater 3h & \cellcolor[HTML]{FFCCC9}\textgreater 3h \\ \hline
\textbf{100 / 3}                                                                  & \textbf{10000 / 1500}                                                               & 6m 39s (12.50)                                                               & \cellcolor[HTML]{FFCCC9}\textgreater 3h & \cellcolor[HTML]{FFCCC9}\textgreater 3h \\ \hline
\textbf{100 / 5}                                                                  & \textbf{10000 / 5000}                                                               & 41m 21s (6.19)                                                               & \cellcolor[HTML]{FFCCC9}\textgreater 3h & \cellcolor[HTML]{FFCCC9}\textgreater 3h \\ \hline
\textbf{500 / 4}                                                                  & \textbf{10000 / 10000}                                                              & 58m 30s (10.65)                                                              & \cellcolor[HTML]{FFCCC9}\textgreater 3h & \cellcolor[HTML]{FFCCC9}\textgreater 3h \\ \hline
\end{tabular}}
\caption{Computation time for a single solve of Problem \eqref{tree_based_problem1.prob} using our exact algorithm.}
\label{optimal_method_time.table}
\end{table}

From Table \ref{optimal_method_time.table}, we observe that our optimal algorithm significantly outperforms commercial solvers in solving Problem \eqref{tree_based_problem1.prob}. Even for the smallest instance of Problem \eqref{tree_based_problem1.prob} considered, both Gurobi and MOSEK fail to find an optimal solution within a 3-hour time limit. In contrast, our tailored algorithm typically identifies the optimal solution within seconds to minutes. This timing experiment demonstrates that solving Problem~\eqref{tree_based_problem1.prob} to optimality using our exact algorithm is practical for a wide range of problem sizes. In the next section, we assess the computational efficiency of our approximate algorithm when computing regularization paths. \color{black} We also compare the computation time of our optimal algorithm directly against that of our approximate algorithm in \S\ref{OA_vs_approx.appx} of the appendix. \color{black}

% many problem sizes, provided the desired number of rules $K$ is specified in advance. In practice, however, the optimal number of rules may be unknown, and computing regularization paths by repeatedly solving Problem~\eqref{tree_based_problem1.prob} to optimality across multiple values of $K$ can be computationally prohibitive for larger problems.  In the next section, we use our approximate algorithm to compute regularization paths and show that this approach enables practitioners to efficiently assess the trade-off between model size and accuracy.

% This timing experiment demonstrates that solving Problem \eqref{tree_based_problem1.prob} to optimality using our exact algorithm is practical, provided the problem size is moderate and the desired number of rules $K$ is prespecified. In practice, however, tree ensembles can grow extremely large; for instance, a boosted ensemble of 500 depth 7 trees may contain over 100,000 nodes, corresponding to decision variables in Problem \eqref{tree_based_problem1.prob}. Moreover, the best number of rules to extract is often unknown in advance. Practitioners typically generate rule sets of varying sizes by computing regularization paths in order to evaluate the tradeoff between model size and accuracy or to select the model with the best validation performance. This requires solving Problem \eqref{tree_based_problem1.prob} repeatedly across multiple values of $K$, which can be computationally expensive. 

% In the next section, we use our CBCD algorithm to compute regularization paths in order to extract rule sets of varying sizes 

\subsubsection{Computation Time: Regularization Paths}\label{approx_timing.section}

\begin{table}[h] \centering
\scalebox{.7}{
\begin{tabular}{|c|c|c|c|}
\hline
\textbf{\begin{tabular}[c]{@{}c@{}}Ensemble Size\\ \# Trees / Depth\end{tabular}} & \textbf{\begin{tabular}[c]{@{}c@{}}Problem Size\\ Obs. / Binary Vars.\end{tabular}} & \textbf{\begin{tabular}[c]{@{}c@{}}Approximate\\ Algorithm\end{tabular}} & \textbf{\begin{tabular}[c]{@{}c@{}}w/o Computational \\ Enhancements\end{tabular}} \\ \hline
\textbf{500 / 4}                                                                  & \textbf{5000 / 10000}                                                               & 45s (0.79)                                                                    & \cellcolor[HTML]{FFCCC9}\textgreater 3h                                                \\ \hline
\textbf{300 / 5}                                                                  & \textbf{5000 / 15000}                                                               & 1m 7s (1.54)                                                                  & \cellcolor[HTML]{FFCCC9}\textgreater 3h                                                \\ \hline
\textbf{600 / 5}                                                                  & \textbf{5000 / 30000}                                                               & 1m 32s (1.40)                                                                 & \cellcolor[HTML]{FFCCC9}\textgreater 3h                                                \\ \hline
\textbf{500 / 6}                                                                  & \textbf{5000 / 50000}                                                               & 2m 51s (4.12)                                                                 & \cellcolor[HTML]{FFCCC9}\textgreater 3h                                                \\ \hline
\textbf{500 / 7}                                                                  & \textbf{5000 / 100000}                                                              & 3m 30s (6.91)                                                                 & \cellcolor[HTML]{FFCCC9}\textgreater 3h                                                \\ \hline
\textbf{500 / 4}                                                                  & \textbf{10000 / 10000}                                                              & 52.4s (0.77)                                                                  & \cellcolor[HTML]{FFCCC9}\textgreater 3h                                                \\ \hline
\textbf{300 / 5}                                                                  & \textbf{10000 / 15000}                                                              & 1m 17s (1.37)                                                                 & \cellcolor[HTML]{FFCCC9}\textgreater 3h                                                \\ \hline
\textbf{600 / 5}                                                                  & \textbf{10000 / 30000}                                                              & 2m 5s (4.53)                                                                  & \cellcolor[HTML]{FFCCC9}\textgreater 3h                                                \\ \hline
\textbf{500 / 6}                                                                  & \textbf{10000 / 50000}                                                              & 2m 57s (6.12)                                                                 & \cellcolor[HTML]{FFCCC9}\textgreater 3h                                                \\ \hline
\textbf{500 / 7}                                                                  & \textbf{10000 / 100000}                                                             & 3m 59s (7.41)                                                                 & \cellcolor[HTML]{FFCCC9}\textgreater 3h                                                \\ \hline
\textbf{500 / 4}                                                                  & \textbf{15000 / 10000}                                                              & 1m 33s (2.33)                                                                 & \cellcolor[HTML]{FFCCC9}\textgreater 3h                                                \\ \hline
\textbf{300 / 5}                                                                  & \textbf{15000 / 15000}                                                              & 1m 57s (3.12)                                                                 & \cellcolor[HTML]{FFCCC9}\textgreater{}3h                                               \\ \hline
\textbf{600 / 5}                                                                  & \textbf{15000 / 30000}                                                              & 2m 51s (3.51)                                                                 & \cellcolor[HTML]{FFCCC9}\textgreater 3h                                                \\ \hline
\textbf{500 / 6}                                                                  & \textbf{15000 / 50000}                                                              & 3m 56s (8.49)                                                                 & \cellcolor[HTML]{FFCCC9}\textgreater 3h                                                \\ \hline
\textbf{500 / 7}                                                                  & \textbf{15000 / 100000}                                                             & 4m 59s (10.48)                                                                & \cellcolor[HTML]{FFCCC9}\textgreater 3h                                                \\ \hline
\end{tabular}}
\caption{ Computation time required to generate the regularization path for 50 values of $\lambda$. The rightmost column reports runtimes without the enhancements described in \S\ref{computation_enhancements.section}.}
\label{CBCD_times.tbl}
\end{table}

We use this procedure to evaluate the computational efficiency of our approximate algorithm for computing regularization paths. For various size instances of Problem~\eqref{tree_based_penalized.prob}, we apply Algorithm~\ref{cbcd.alg} across 50 values of $\lambda$, evenly spaced on a logarithmic scale from $10^0$ to $10^3$. We incorporate the computational enhancements discussed in \S\ref{computation_enhancements.section} and record the total time required to compute the full regularization path. This yields a sequence of solutions to Problem~\eqref{tree_based_problem1.prob} with varying sparsity levels, corresponding to rule sets of different sizes. For comparison, we also run Algorithm~\ref{cbcd.alg} without the enhancements from \S\ref{computation_enhancements.section}, using Gurobi to solve Problem~\eqref{blockupdate.prob} at each block update. 

The results of this timing experiment are reported in Table~\ref{CBCD_times.tbl}, where we show the computation time of each method averaged across 5 runs (the standard errors are shown in parentheses). The left two columns  report the ensemble size and the dimensions of each instance of Problem~\eqref{tree_based_penalized.prob}, specified by the number of observations and binary decision variables (i.e., nodes in the tree ensemble). The smallest instance considered, 5,000 observations and 10,000 binary decision variables, matches the largest problem size evaluated in our timing experiments for the exact algorithm. In this experiment, we scale to significantly larger ensembles, evaluating regularization paths for problems with up to 100,000 binary decision variables. Table~\ref{CBCD_times.tbl} shows that our approximate algorithm (Algorithm~\ref{cbcd.alg}), when combined with the computational enhancements from \S\ref{computation_enhancements.section}, is able to compute regularization paths across all problem sizes within minutes. In contrast, without these enhancements, the algorithm fails to compute any regularization path within three hours. These results demonstrate that our enhanced approximate algorithm can efficiently compute regularization paths for large-scale instances of Problem~\eqref{tree_based_penalized.prob}. 

% This enables practitioners to extract rule sets of varying sizes from large tree ensembles and effectively evaluate the trade-off between model size and predictive performance.

% In contrast, the optimal algorithm (Algorithm \ref{OA_main.alg}) fails to compute regularization paths--by solving instances of Problem \eqref{tree_based_problem1.prob} with the corresponding sparsities--within a 3-hour time limit for any of the problem sizes considered. In addition, our abated baseline also fails to compute regularization paths for any problem size within 3 hours. These results demonstrate that our approximate CBCD algorithm, when combined with the speedup techniques from \S\ref{efficient_block_update.section} and \S\ref{speedups.section}, can efficiently compute regularization paths even for large-scale instances of Problem \eqref{tree_based_penalized.prob}. This allows practitioners to use our approximate algorithm to efficiently extract rule sets of varying sizes from large tree ensembles in order to evaluate the trade-off between model compactness and predictive performance.

\subsection{Predictive Performance} \label{interpretable_rules_experiment.section}

In this section, we evaluate how well our estimator performs at extracting interpretable sets of decision rules from tree ensembles. We follow the experimental procedure discussed below on 25 regression datasets from the OpenML repository \citep{bischl2017openml}, and the full list of datasets along with metadata can be found in \S\ref{datasets.appx} of the appendix.

\begin{figure}[h]
    \centering
\includegraphics[width=0.85\linewidth]{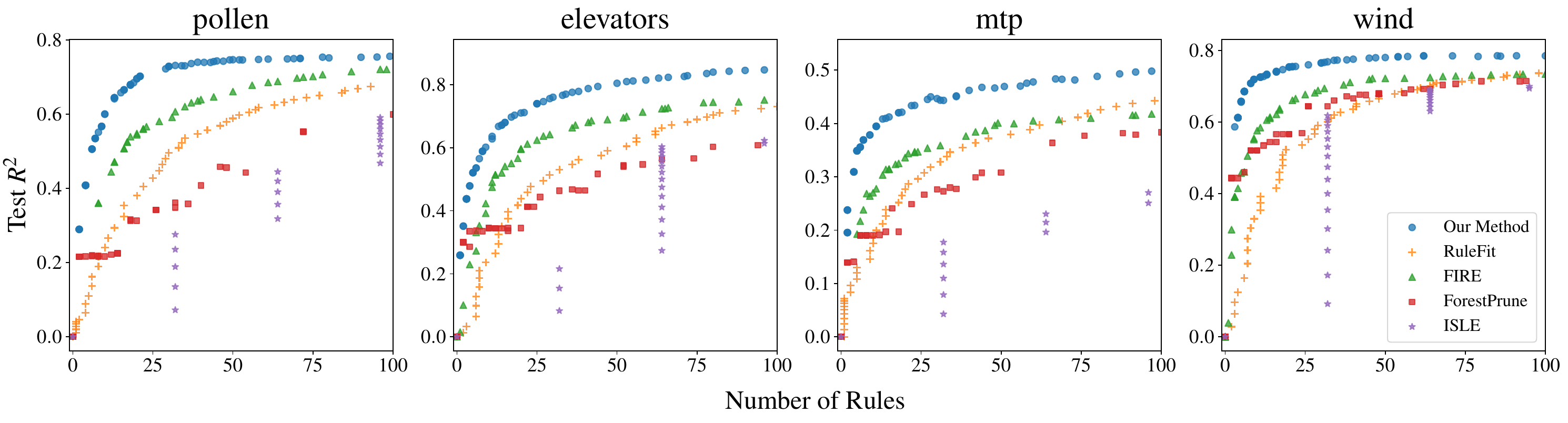}
    \caption{Visualization of regularization paths for all methods in our experiment on 4 datasets from OpenML \citep{bischl2017openml}. The horizontal axes show the number of rules extracted and the vertical axis shows out-of-sample $R^2$.}
    \label{experiment_reg_paths.fig}
\end{figure}

\vspace{2mm}
\noindent{\textbf{Experimental Procedure:}} 
For each dataset, we perform 5-fold cross-validation. On the training folds, we fit gradient boosting tree ensembles with maximum depths of $\{3, 5, 7\}$ and for each ensemble, we apply our estimator and use Algorithm \ref{cbcd.alg} to compute regularization paths. Along each path, we extract sets of $K = \{10, 15, 20, 25\}$ decision rules and evaluate the out-of-sample $R^2$ of the resulting compact models. We restrict our extracted rule sets to have no more than 25 rules to preserve interpretability, so that the rules can be examined by hand. We compare the performance of our framework against these competing algorithms.
% \begin{itemize}[leftmargin=*,noitemsep, topsep=0pt] 
     \textbf{RuleFit} \citep{friedman2008predictive}, which uses the LASSO to select decision rules without pruning depth.
     \textbf{FIRE} \citep{liu2023fire}, which uses a sparsity-inducing MCP-penalty to select rules without pruning rule depth.
     \textbf{ForestPrune} \citep{liu2023forestprune}, which selects trees and prunes depths without explicitly selecting rules.
     \textbf{ISLE} \citep{friedman2003importance}, which uses the LASSO to select trees.
% \end{itemize}
For each competing method, we compute regularization paths and extract models with at most $K = \{10, 15, 20,25\}$ decision rules, or leaf nodes for ISLE and ForestPrune. We compare the out-of-sample $R^2$ of these models against the compact models obtained by our framework, and we show a visualization of this experiment in Figure \ref{experiment_reg_paths.fig}.

\begin{figure}[h]
  \centering
  \begin{adjustbox}{width=\linewidth,center}
  \begin{tabular}{cc}
    % --- Figure ---
    \begin{minipage}{0.7\linewidth}
      \centering
      \includegraphics[width=\linewidth]{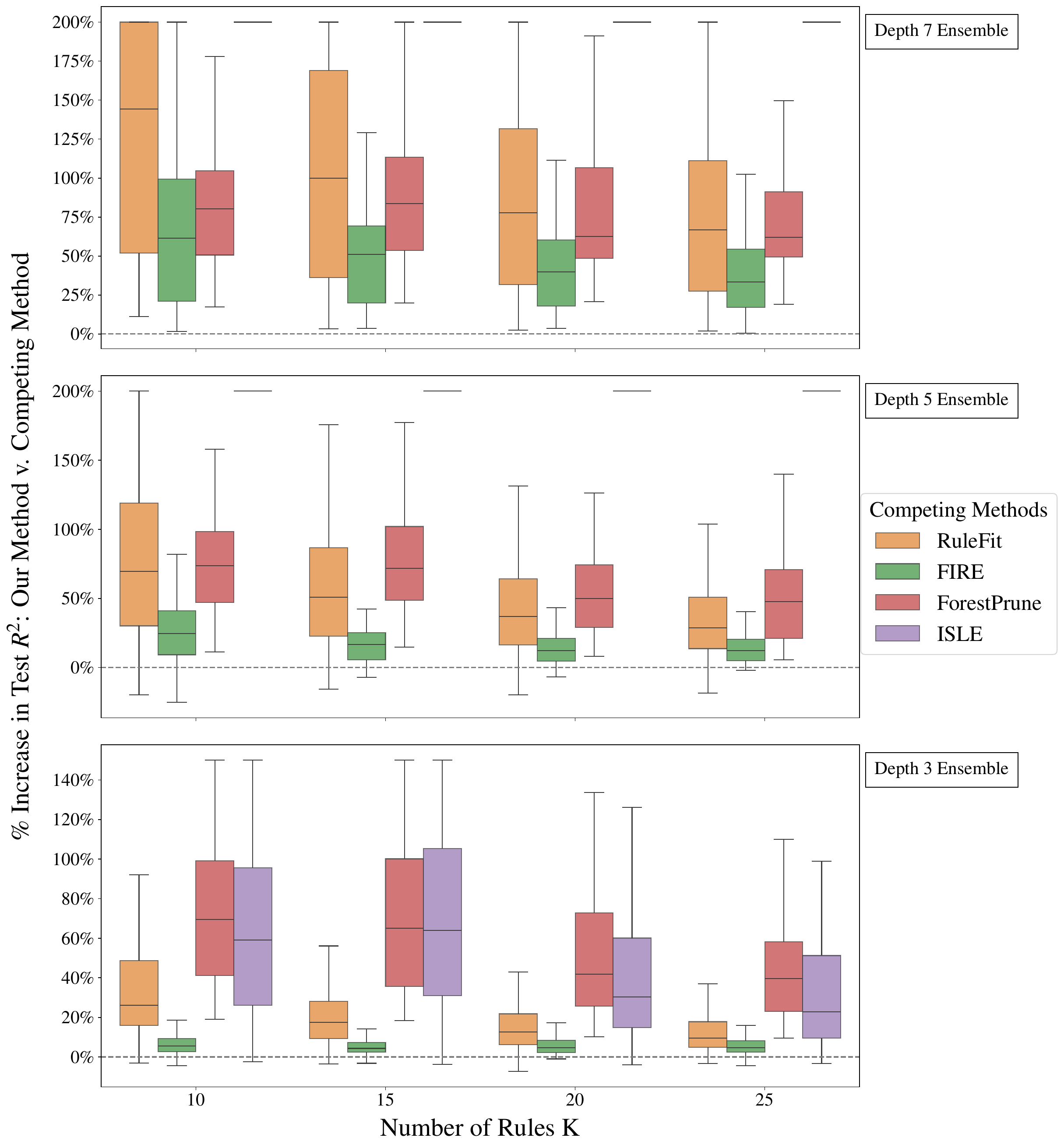}
    \end{minipage}
    &
    % --- Side caption ---
    \begin{minipage}{0.28\linewidth}
      \captionof{figure}{%
      Experimental results. The horizontal axis shows the budget on the number of rules, 
        and the vertical axes show the percent increase in test $R^2$ 
        between our algorithm and competing methods. 
       Values above 0 (indicated by the dashed line) 
        show that we outperform the competing method.%
      }
\label{rule_experiment_results.pdf}
    \end{minipage}
  \end{tabular}
  \end{adjustbox}
\end{figure}

\vspace{2mm}
\noindent{\textbf{Experimental Results:}}
Across all datasets, folds, ensemble depths, and budgets of rules $K$, we compare the out-of-sample $R^2$ of our framework against our competing methods. We report the percent increase in out-of-sample $R^2$ between our framework and our competing algorithms, defined by: $(\text{Our Method} \ R^2 - \text{Competing Method} \ R^2)/(\text{Competing Method} \ R^2)$. Percent increases above 0 indicate that our method outperforms the competing algorithm. In Figure \ref{rule_experiment_results.pdf}, we show the distribution of these results across all datasets and folds in our experiment. Each panel in the figure shows results for a different ensemble depth, $\{3,5,7\}$. The horizontal axes in each panel shows $K$, the budget on the number of rules, and the vertical axes show the percent increase between our algorithm and our competing methods. The box-plots are color-coded by competing method and show the distribution of percent increases across all datasets and folds.

% \begin{figure}[h]
%     \centering
%     \includegraphics[width=.7\linewidth]{figures/rule_experiments_results.pdf}
%     \caption{Experimental results for extracting interpretable rule sets. The horizontal axis shows the budget on the number of rules and the vertical axes show the percent increase in test $R^2$ between our algorithm and the competing methods. Percent increases above 0 (indicated by the dashed line) show that our algorithm outperforms the competing method.}
%     \label{rule_experiment_results.pdf}
% \end{figure}

% We also show the results of our experiment by datasets in \S\ref{experiment_results.appx} of the appendix.

We observe that our estimator consistently outperforms competing methods in constructing interpretable rule sets; the interquartile ranges of all box plots in Figure \ref{rule_experiment_results.pdf} are strictly positive. Compared to the best-performing competing method, FIRE, our approach achieves approximately a \textbf{50\%}, \textbf{20\%}, and \textbf{8\%} increase in out-of-sample $R^2$ when extracting rule sets from ensembles of depths 7, 5, and 3, respectively. Our method also shows increasing performance gains over RuleFit and FIRE as ensemble depth increases, owing to its ability to prune rule depth. In contrast, these competing methods are limited to extracting sets of high-depth rules, which tend to perform poorly. In addition, ForestPrune performs poorly due to its limited flexibility: it can only prune tree depth and cannot extract decision rules from trees, which restricts its effectiveness in producing sparse rule sets. Finally we note that ISLE fails to extract sparse rule sets when ensemble depth is greater than 3. A single  tree of depth 5 already contains more than 25 rules, so we omit ISLE from the depth 5 and depth 7 displays.

These results demonstrate that our method significantly outperforms competing algorithms in extracting sparse, interpretable rule sets. 
\color{black} In \S\ref{comparison_original.appx} of the appendix, we compare the performance of the extracted rule sets against the predictive performance of the original ensembles. Our results indicate that even the most compact extracted models ($K = 10$) retain a high degree of the predictive accuracy of the original ensembles, with a median decrease in out-of-sample $R^2$ of around 10\%. Finally, in \S\ref{model_compress.appx} of the appendix, we examine how effectively our estimator can compress large  ensembles to achieve memory and inference savings. We find that we can compress ensembles by up to three orders of magnitude with only about a  $5\%$ decrease in out-of-sample $R^2$. While the resulting models may not be interpretable, this highlights the ability of our method to serve as a powerful compression tool to reduce memory usage and inference cost.

\subsubsection{Comparisons Against Directly Constructing Interpretable Models}

\begin{figure}[h]
    \centering
    \includegraphics[width=0.85\linewidth]{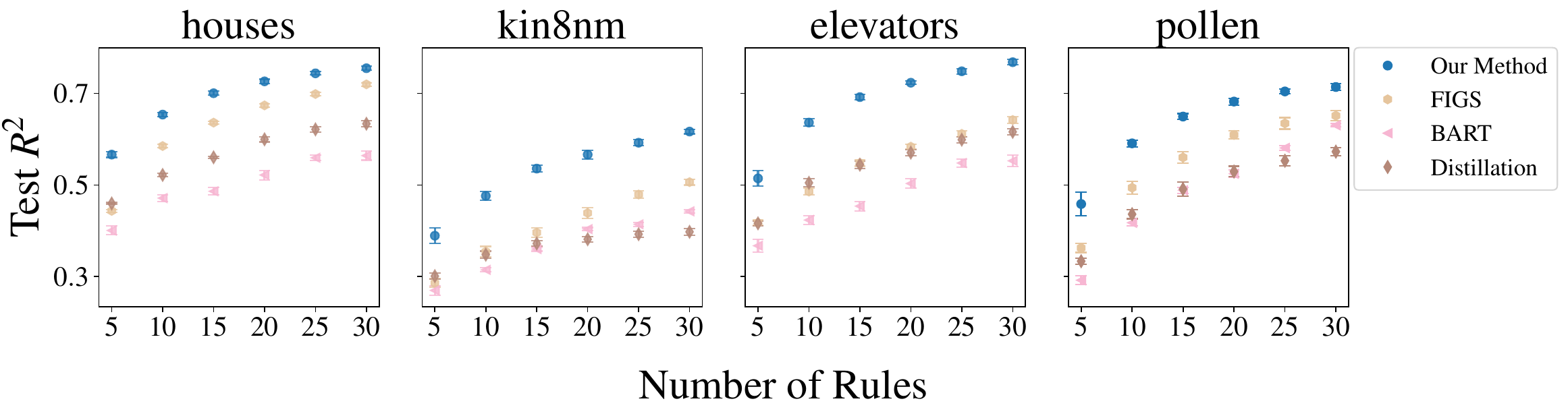}
    \caption{Comparisons of our estimator against tree-based interpretable models.}
    \label{revision_scatter.fig}
\end{figure}

We also compare our estimator for rule extraction against algorithms that directly construct interpretable tree-based models from the data. These include: FIGS \citep{tan2025fast}, which iteratively constructs sums of trees, Bayesian Additive Regression Trees \citep{chipman2010bart}, which can produce compact tree ensembles, and distillation trees, where student CART trees are trained to approximate the predictions of the black-box ensemble \citep{bucilua2006model}. We repeat the experimental procedure discussed above and use each competing method to construct interpretable models of $K = \{5, 10, 15, 20, 25, 30\}$ rules. We also apply our estimator to extract rule sets of the same size from a  gradient boosting ensemble. In Figure \ref{revision_scatter.fig}, we visualize the results of our experiment on 4 datasets from OpenML. The horizontal axes show model size, the vertical axis shows test $R^2$ averaged across a 5-fold cross-validation, and the error bars show standard error. We see from these plots that our estimator  consistently achieves higher test $R^2$ scores across all model sizes.

\begin{figure}[h]
    \centering
    \includegraphics[width=0.75\linewidth]{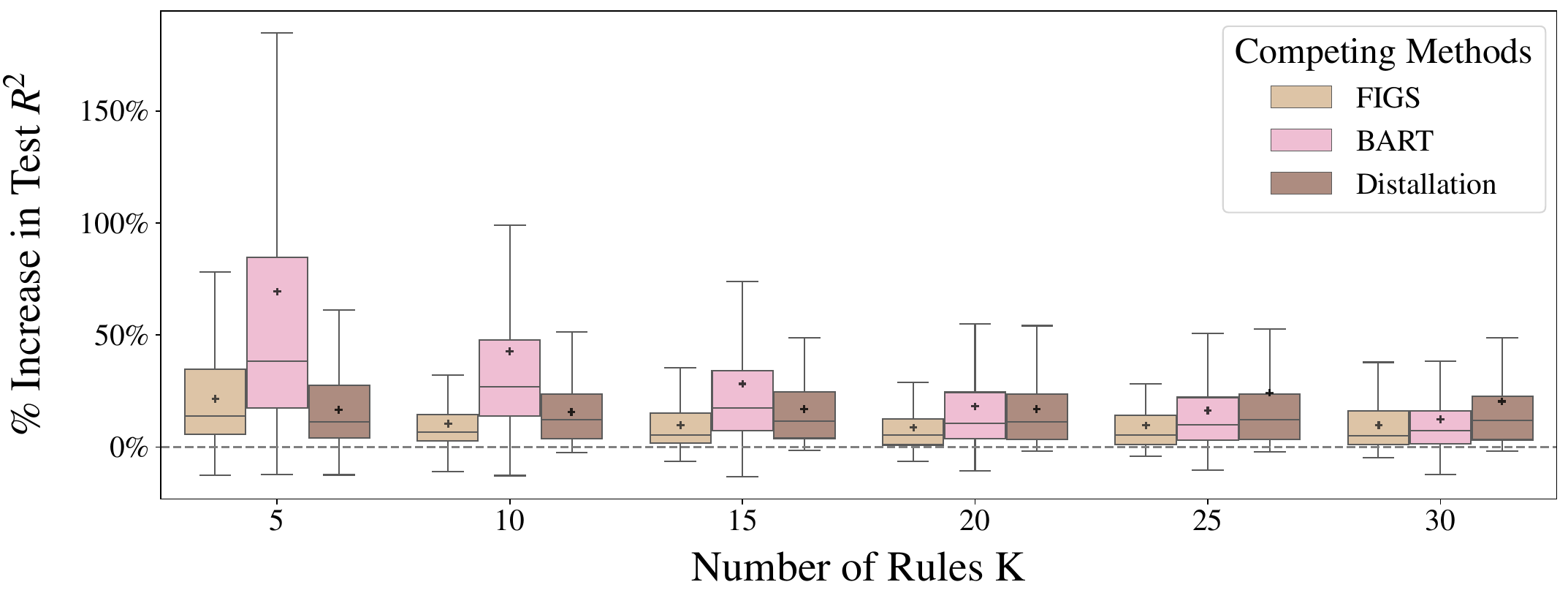}
    \caption{Distribution of  results. Our estimator  consistently outperforms algorithms that directly constructs interpretable models.}
    \label{revision_box.fig}
\end{figure}

In Figure \ref{revision_box.fig}, we show the distribution of our results across all datasets and folds in our experiment. The horizontal axis of the plot shows model sizes and the vertical axis shows the percent increase in test $R^2$ between our method and the competing algorithm. As before, percent increases above 0 show that our method outperforms the competing algorithm. The cross in each box-plot indicates the mean of the distribution.

We note that across all model sizes considered, our proposed estimator for extracting rule sets consistently outperforms the competing algorithms. This likely stems from the greater flexibility of rule sets compared to the methods discussed above, which are constrained to form compact tree ensembles. While such structural restrictions can make those models easier to visualize, they may also reduce model accuracy. In \S\ref{scorecard.section}, we introduce an alternative non-graphical way to present our extracted rule sets, offering a complementary and easily interpretable means of conveying the model.

\color{black}

\color{black}
\subsection{Case Study: Intensive Care Unit Length of Stay Prediction} \label{case_study.section}

We conclude with a case study that illustrates how to interpret the models extracted by our estimator in a real-world application; additional details of the study are provided in \S\ref{additional_details_case.appx}. Using the publicly available MIMIC-III critical care database \citep{kallfelz2021mimic}, we  consider a cohort of adult patients aged 18-65 who were admitted to the intensive care unit (ICU). For each patient, we compile clinical features from the first 24 hours of admission, including vital signs, laboratory values, administered therapies, and admission characteristics, with the goal of predicting overall ICU length of stay (LOS). The resulting dataset consists of 21,023 ICU admissions with 217 covariates, which we randomly partition into a training set (70\%) and a test set (30\%).

Predicting patient-level ICU length of stay is challenging; baseline interpretable models such as LASSO regression or a single decision tree achieve only modest out-of-sample $R^2$ scores of 0.19 and 0.21, respectively. A gradient boosting ensemble of 1000 depth 3 trees performs significantly better, with an out-of-sample $R^2$ score of 0.32, consistent with prior studies. However, this accuracy comes at the cost of interpretability, as the ensemble is a black-box model. When we apply our estimator, we extract an interpretable set of \textbf{10} decision rules that achieves an out-of-sample $R^2$ score of \textbf{0.31}. 

\begin{figure}[h]
    \centering
    \includegraphics[width=0.7\linewidth]{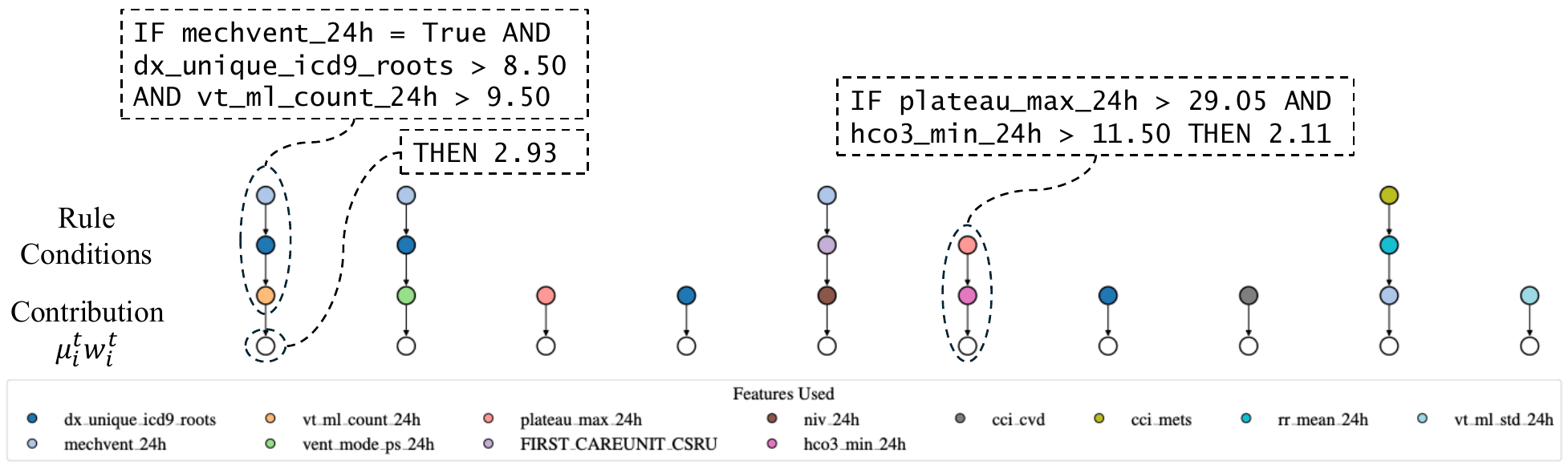}
    \caption{Interpretable rule set extracted by our estimator to predict ICU length of stay.}
    \label{mimic_case.fig}
\end{figure}

In Figure~\ref{mimic_case.fig}, we present the extracted rule set, and we can examine this visualization to identify the features used and the interaction depths of the rules. To interpret the model, we further decompose how the rule set generates predictions.

\subsubsection{Interpreting the Extracted Model} \label{scorecard.section}

Recall from \S\ref{formulation.section} that we can write the prediction vector of the rule set as $\sum_{(i,t) \in \mathcal{S}} \mathbf{M}_i^t \, w_i^t$, where $\mathcal{S} = \{ (i,t) : w_i^t \neq 0 \}$ denotes the set of extracted rules with nonzero weights. Each extracted rule  $(i, \ t) \in \mathcal{S}$ contains a sequence of if-then conditions. When these conditions are satisfied by an observation, the corresponding contribution to the prediction\footnote{Recall that $(\mathbf{M}_i^{\,t})_k = \mu_i^t\,\mathbbm{1}_{\{\text{node } i \text{ of tree } t \text{ contains obs.\ }k\}}$} is given by $\mu_i^t \, w_i^t$. To show how the extracted model generates predictions, we can represent it in \textbf{scorecard form}, as shown in Table \ref{scorecard_form.table}.

\begin{table}[ht]
\centering
\scalebox{0.75}{
\renewcommand{\arraystretch}{1.25}  % row spacing
\setlength{\tabcolsep}{8pt}        % column spacing
\begin{tabular}{c >{\raggedright\arraybackslash}p{10.5cm} r}
\toprule
 & \textbf{Rule Conditions} & \textbf{Contribution} \\
\midrule
$\square$ &  \texttt{mechvent\_24h} = True AND \texttt{dx\_unique\_icd9\_roots} $>$ 8.50 AND \texttt{vt\_ml\_count\_24h} $>$ 9.50 & \textcolor{green!40!black}{+2.93 days} \\
\addlinespace[0.7em]
$\square$ &  \texttt{mechvent\_24h} = True AND \texttt{first\_careunit\_csru} = False AND \texttt{niv\_24h} = False & \textcolor{green!40!black}{+2.35 days} \\
\addlinespace[0.7em]
$\square$ &  \texttt{mechvent\_24h} = True AND \texttt{dx\_unique\_icd9\_roots} $>$ 16.50 AND \texttt{vent\_mode\_ps\_24h} = False & \textcolor{green!40!black}{+3.05 days} \\
\addlinespace[0.7em]
$\square$ &  \texttt{plateau\_max\_24h} $\leq$ 23.35 & \textcolor{red!70!black}{-1.82 days} \\
\addlinespace[0.7em]
$\square$ &  \texttt{cci\_mets} = False AND \texttt{rr\_mean\_24h} $>$ 20.85 AND \texttt{mechvent\_24h} = True & \textcolor{green!40!black}{+1.65 days} \\
\addlinespace[0.7em]
$\square$ &  \texttt{cci\_cvd} = True & \textcolor{green!40!black}{+1.65 days} \\
\addlinespace[0.7em]
$\square$ &  \texttt{dx\_unique\_icd9\_roots} $>$ 21.50 & \textcolor{green!40!black}{+1.86 days} \\
\addlinespace[0.7em]
$\square$ &  \texttt{plateau\_max\_24h} $>$ 29.05 AND \texttt{hco3\_min\_24h} $>$ 11.50 & \textcolor{green!40!black}{+2.11 days} \\
\addlinespace[0.7em]
$\square$ &  \texttt{dx\_unique\_icd9\_roots} $\leq$ 7.50 & \textcolor{red!70!black}{-0.85 days} \\
\addlinespace[0.7em]
$\square$ &  \texttt{vt\_ml\_std\_24h} $>$ 170.29 & \textcolor{red!70!black}{-0.97 days} \\
\bottomrule
\end{tabular}}
\caption{Scorecard form of extracted rule set.}
\label{scorecard_form.table}
\end{table}

Each item in this scorecard corresponds to a rule: the left column shows rule conditions and the right column shows the contribution, $\mu_i^t \, w_i^t$, when the conditions are satisfied. Importantly, we note that weights $w_i^t$ for all $(i, \ t) \in \mathcal{S}$ are not shown to the practitioner, only the value of the contributions are displayed. To generate predictions for a new observation, a practitioner proceeds through the scorecard sequentially; whenever the conditions of a rule are satisfied, the corresponding contribution is added to the prediction. The final prediction is obtained by summing the contributions from all satisfied rules and adding the mean of the training response (recall from \S\ref{framework.section} that $y$ is mean-centered).

\vspace{2mm}
\noindent{\textbf{Remarks on Scorecard Form:}} We order the rules in descending importance,  which can be assessed using several metrics. For instance, rules can be ranked by the magnitude of their contribution, where more important rules have a larger impact on the model’s prediction. Alternatively, rules can be ranked by the proportion of observations in the training data that satisfy the rule conditions, which reflects how broadly applicable a rule is. Finally, a combined measure of impact and applicability can be used: \cite{friedman2008predictive} compute rule importance via the expression: $|\text{contribution}| \,\sqrt{\text{proportion}\,(1-\text{proportion})}$. The rules in Table \ref{scorecard_form.table} are ordered according to this metric.

We also note that when presented in scorecard form, our extracted model closely parallels established scoring systems deployed in high-stakes settings, such as the Public Safety Assessment model \citep{LJAF2013PSA}, which is used to guide judicial decision-making, and the APACHE II scoring system \cite{Knaus1985APACHEII}, which is used to assess disease severity. We examine these similarities in detail in \S\ref{existing_interpretable.appx} of the appendix.

\vspace{2mm}
\noindent{\textbf{Uncovering Real-World Insights:}} Importantly, practitioners can examine the scorecard rule by rule to understand how predictions are generated and to uncover insights from the data. We illustrate this process by explaining the four most important rules in Table \ref{scorecard_form.table} and we examine the remainder of the rules in \S\ref{rule_discuss.appx} of the appendix.
\begin{itemize}[leftmargin=0pt ,itemsep=1pt, topsep=1pt]
    \item \textbf{Rule 1:} If a patient is placed on mechanical ventilation within the first 24 hours, has high diagnostic complexity (measured using ICD-9 root codes at admission), and requires frequent ventilator adjustments (measured using tidal volume measurements), then the predicted length of stay increases by 2.93 days. 

    \item \textbf{Rule 2:} If a patient is placed on mechanical ventilation within the first 24 hours and was not admitted to the cardiac surgery recovery unit (CSRU) and did not receive non-invasive ventilation, then the predicted length of stay increases by 2.35 days. 
    
    \item \textbf{Rule 3:} If a patient is placed on mechanical ventilation within the first 24 hours, has very high diagnostic complexity, and does not receive pressure support ventilation, then the predicted length of stay increases by 3.05 days.
    \item \textbf{Rule 4:}  If a patient’s maximum plateau pressure is below 23.35 cmH$_2$O, then the predicted length of stay decreases by 1.82 days.
\end{itemize}

Rule 1 to 3 highlight that patients who require  ventilation and present with complex diagnoses are predicted to have longer ICU stays. Rule 4 suggests that less aggressive ventilator settings in the first 24 hours are associated with shorter predicted ICU stays. To conclude, this case study demonstrates that our estimator can prune a complex tree ensemble into a compact rule set with comparable predictive accuracy. Presented in scorecard form,  these rule sets allows practitioners to readily explain predictions and uncover useful relationships in the data. 
\color{black}

% \begin{itemize}
%     \item Rule 4: If a patient does not have metastatic cancer and has a high respiratory rate and is placed on mechanical ventilation within the first 24 hours then the predicted length of stay increases by 1.65 days.
%     \item Rule 5: If a patient has cardiovascular disease then the predicted length of stay increases by 1.65 days.
%     \item Rule 6: If a patient has an extremely complicated diagnosis (measured using ICD-9 root codes at admission) the predicted length of stay increases by 1.86  days.
%     \item Rule 7: If a patient has an aggressive ventilator setting then the predicted length of stay increases by 2.11 days.
%     \item Rule 8: If a patient has a simpler diagnosis (measured using ICD-9 codes at admission) the predicted length of stay decreases by 2.11 days.
% \end{itemize}

\noindent\textbf{Acknowledgments.}
The authors acknowledge support from the Office of Naval Research (ONR N000142212665), the MIT Sloan Health Systems Initiative, and the MIT Health and Life Sciences Collaborative. We also thank Trevor Hastie, Rob Tibshirani, Jerry Friedman, and Stephen Boyd for their helpful discussions.

\noindent\textbf{Code Availability.}
A packaged implementation of our procedure, along with code to reproduce the experiments in this paper, is available at \url{https://github.com/brianliu12437/TreeExtract/}.

\newpage
%check the page count
% \newcount\linenum
% \linenum=1
% \loop
% Line \the\linenum\ of sample text.\\
% \advance\linenum by 1
% \ifnum\linenum <28
% \repeat
% \newpage

\appendix

\section{Preliminaries and Related Work Discussion} 

\subsection{Decision Trees to Decision Rules} \label{trees_to_rules.appx}

\begin{figure}[h]
    \centering
  \begin{adjustbox}{width=.7\linewidth,center}  
    % --- First Subfigure: Full Decision Tree ---
    \begin{subfigure}{0.45\textwidth}
        \centering
        \begin{tikzpicture}[
            level distance=1cm, 
            sibling distance=1.5cm, 
            every node/.style={circle,draw,minimum size=12mm, inner sep=1.7pt},
            font=\scriptsize, 
            level 1/.style = {level distance=10mm, sibling distance=28mm},
            level 2/.style = {level distance=15mm, sibling distance=14mm}
        ]
            \node[draw=purple, text=purple, thick]  {$x_1 \leq s_1$}
                child { node {$x_1 \leq s_2$}
                    child { node[draw=darkpastelgreen, text=darkpastelgreen, thick] {leaf} }
                    child { node[draw=darkpastelgreen, text=darkpastelgreen, thick] {leaf} }
                }
                child { node {$x_2 \leq s_3$}
                    child { node[draw=darkpastelgreen, text=darkpastelgreen, thick] {leaf} }
                    child { node[draw=darkpastelgreen, text=darkpastelgreen, thick] {leaf} }
                };
        \end{tikzpicture}
    \end{subfigure}
    \hfill
    % --- Second Subfigure: Sum of Stumps ---
    \begin{subfigure}{0.45\textwidth}
        \centering
        \begin{tikzpicture}[
            level distance=14mm, 
            sibling distance=1.5cm, 
            every node/.style={circle,draw,minimum size=12mm, inner sep=1.7pt},
            font=\scriptsize
        ]

            % First stump (depth 2)
            \node[draw=purple, text=purple, thick]  (T1) {$x_1 \leq s_1$}
                child { node {$x_1 \leq s_2$}
                     node[below=16mm of T1, draw=darkpastelgreen, text=darkpastelgreen, thick] {$r_1(\mathbf{x})$} 
                };

            % Second stump (depth 2)
            \node[right=.5cm of T1,draw=purple, text=purple, thick] (T2) {$x_1 \leq s_2$}
                child { node {$x_1 > s_2$}
                    node[below=16mm of T2, draw=darkpastelgreen, text=darkpastelgreen, thick] {$r_2(\mathbf{x})$} };

            % Third stump (depth 2)
            \node[right=.5cm of T2,draw=purple, text=purple, thick] (T3) {$x_1 > s_1$}
                child { node {$x_2 \leq s_3$}
                 node[below=16mm of T3,draw=darkpastelgreen, text=darkpastelgreen, thick] {$r_3(\mathbf{x})$} };

            % Fourth stump (depth 2)
            \node[right=.5cm of T3,draw=purple, text=purple, thick] (T4) {$x_1 > s_1$}
                child { node {$x_2 > s_3$}
                     node[below=16mm of T4,draw=darkpastelgreen, text=darkpastelgreen, thick, label = ] {$r_4(\mathbf{x})$} }; 
        \end{tikzpicture}
    \end{subfigure}
\end{adjustbox}
    \caption{This tree has 4 leaf nodes and decomposes into 4 decision rules.}
    \label{tree_viz.fig}
\end{figure}

We show an example of how a decision tree can be decomposed into decision rules in  Figure \ref{tree_viz.fig}; in the left panel we show a tree of depth 2, $\Gamma(\mathbf{x})$, which is a function of the feature vector $\mathbf{x} = (x_1,x_2)^\top$. This tree has 3 internal nodes and 4 leaf nodes, and values $s_1$, $s_2$, and $s_3$ represent split thresholds. In the right panel, we decompose the decision tree into 4 decision rules. Each rule is obtained by traversing the tree from root node (purple) to leaf node (green). For example, the prediction function for rule 2 is given by $r_2(\mathbf{x}) = \mathbbm{1}(x_1 \leq s_1) \cdot \mathbbm{1}(x_1 > s_2)\cdot \mu_2$, where $\mu_2$ is the mean response of all of the observations in rule 2. We also have that $\Gamma(\mathbf{x}) = \sum_{j=1}^4 r_j(\mathbf{x})$ an important property that we leverage in our proposed estimator.

\subsection{Related Work Discussion}\label{related_works.appx}
We discuss additional details on existing ensemble pruning algorithms and highlight the advantages of our proposed estimator.

As mentioned in \S\ref{preliminaries.section}, the paper Importance Sampled Learning Ensembles \citep{friedman2003importance}, ISLE, introduces one of the earliest computational approaches to prune tree ensembles. Consider an ensemble of $T$ decision trees, $\{\Gamma_1, \ldots \Gamma_T\}$ and let~$\mathbf{\Gamma}_t$ denote the vector of predictions for the tree~$\Gamma_t$. 
ISLE uses this optimization problem to prune trees: $\min_{\bm{\beta}} ||\mathbf{y} - \sum_{t=1}^T\beta_t \  \mathbf{\Gamma}_t||_2^2 + \lambda||\bm{\beta}||_1$, where~$\bY \in \mathbb{R}^n$ is the response vector and~$\lambda$ is the weight for the LASSO penalty that encourages sparsity in trees. The method is only able to extract \emph{entire} trees from the ensemble, and cannot prune interaction depth or individual rules.

RuleFit \citep{friedman2008predictive} deconstructs a tree ensemble into a large collection of $m'$ decision rules and uses the LASSO to extract a sparse subset of rules: $\min_{\bm{\beta}} ||\mathbf{y} - \sum_{j=1}^{m'}\beta_j \  \mathbf{r}_j||_2^2 + \lambda||\bm{\beta}||_1$, where~$\mathbf{r}_j\in\mathbb{R}^n$ is the prediction vector for the rule~$r_j$. A similar procedure, Node Harvest \citep{meinshausen2010node} uses the non-negative garrote to sparsify rules. Both of these algorithms struggle to scale to larger problem sizes, e.g., ensembles of depth greater than 4, as shown in the empirical results of \cite{liu2023fire}. Moreover, neither method prunes interaction depth; the extracted rules retain the same depth as those in the original tree ensemble. 

FIRE \citep{liu2023fire} uses this optimization framework to extract rules $\min_{\bm{\beta}} |\mathbf{y} - \sum_{j=1}^{m'} \beta_j \mathbf{r}_j|_2^2 + h(\bm{\beta}, \lambda)$, where $h(\bm{\beta}, \lambda)$ is a non-convex, sparsity-inducing penalty, such as the minimax concave penalty, and greedy optimization techniques are used to improve computation. FIRE scales to pruning larger and deeper ensembles than RuleFit and Node Harvest, and achieves better predictive performance, as demonstrated in \cite{liu2023fire}. However, FIRE does not prune rule depth, and its optimization formulation cannot be readily extended to do so.

Compared to these existing rule-pruning methods, our proposed estimator can support interaction depth pruning, which greatly improves predictive performance as our experiments in \S\ref{interpretable_rules_experiment.section} show. In addition, we provide theoretical guarantees for our estimator, which are not offered by existing methods.

ForestPrune \citep{liu2023forestprune} is an algorithm designed to prune depth layers from tree ensembles. However, it is limited to pruning entire trees and cannot extract individual decision rules. Furthermore, its optimization framework—based on block coordinate descent over full trees—is not easily adaptable for rule extraction. While ForestPrune returns a collection of trees of varying depths, our goal is to extract a compact set of rules. The models produced by ForestPrune are typically larger and less flexible than those produced by our estimator, which supports both depth pruning and rule extraction. Consequently, our method significantly outperforms ForestPrune in extracting compact models in terms of predictive performance, as shown by our experiments in \S\ref{interpretable_rules_experiment.section}. Finally, ForestPrune lacks theoretical guarantees, in contrast to our estimator.

\color{black}

\subsubsection{Connections with Sparse Regression}
We note that ensemble pruning has close connections with sparse regression. As discussed above, RuleFit and ISLE both use the LASSO \citep{tibshirani1996regression} to extract rules and trees from an ensemble, while Node Harvest employs the non-negative garrote \citep{breiman1995better}. These algorithms treat ensemble pruning as a sparse regression problem, where each rule or tree serves as a predictor and regularization is used to select a compact subset. FIRE extends this connection by using the minimax concave penalty \citep{zhang2010nearly} to approximate $\ell_0$-regularization for more aggressive rule selection. These methods can also be extended to incorporate alternative regularization schemes, such as the $\ell_1$–$\ell_2$ elastic net penalty \citep{zou2005regularization} or the $\ell_0$–$\ell_2$ penalty proposed by \citet{mazumder2023subset}, which may further enhance performance in low signal-to-noise ratio settings. For the most part\footnote{FIRE can incorporate an optional fused lasso penalty to group adjacent rules within a decision tree.}, existing rule pruning algorithms do not account for tree structure. In contrast, our proposed estimator explicitly accounts for tree structure when pruning both interaction depth and rules. As a result, the optimization formulation underlying our estimator differs substantially from those used in sparse regression methods.

\subsubsection{Connections with Interaction Detection Algorithms} We also note that it may be possible to use our proposed estimator for detecting feature interactions. By simultaneously pruning rules and interaction depth, the extracted rules can be viewed as important interactions among predictors, selected from the large pool of potential interactions captured by the original tree ensemble. It would be an interesting direction to compare our approach against other tree-based methods for interaction detection, such as iterative random forests \citep{Basu2018iterative} and Bayesian tree ensembles \citep{du19a}, as well as optimization-based approaches for interaction detection \citep{radchenko2010variable}.

\color{black}

% \subsubsection{Connections with Interaction Detection} We also note that the

% consider an ensemble of $T$ decision trees, $\{\Gamma_1, \ldots \Gamma_T\}$ and let~$\mathbf{\Gamma}_t$ denote the vector of predictions for the tree~$\Gamma_t$. One of the earliest optimization-based frameworks to prune  ensembles, Importance Sampled Learning Ensembles (ISLE), uses this optimization formulation to prune trees: $\min_{\bm{\beta}} ||\mathbf{y} - \sum_{t=1}^T\beta_t \  \mathbf{\Gamma}_t||_2^2 + \lambda||\bm{\beta}||_1$, where~$\bY \in \mathbb{R}^n$ is the response vector and~$\lambda$ is the weight for the LASSO penalty that encourages sparsity in trees \citep{friedman2003importance}. Similarly, RuleFit \citep{friedman2008predictive} first deconstructs a tree ensemble into a large collection of $m'$ decision rules and uses the LASSO to extract a sparse subset of rules: $\min_{\bm{\beta}} ||\mathbf{y} - \sum_{j=1}^{m'}\beta_j \  \mathbf{r}_j||_2^2 + \lambda||\bm{\beta}||_1$, where~$\mathbf{r}_j\in\mathbb{R}^n$ is the prediction vector for the rule~$r_j$. A closely related algorithm, Node Harvest \citep{meinshausen2010node}, selects a sparse subset of rules using the non-negative 
\section{Optimal Algorithm}

\subsection{Proof of Propositions}\label{optimal_alg_prop.appx}

In this section, we present the proofs for the propositions used in our optimal algorithm.

\subsubsection{Proof of Proposition \ref{prop1_opt.prop}} \label{prop1_proof.appx}

We prove here that Problem \eqref{CIP_reformulation.prob} is an equivalent reformulation of Problem \eqref{tree_based_problem1.prob}. We display both of these problems below.

\vspace{2mm}
\noindent \textbf{Problem \eqref{tree_based_problem1.prob}}:
\begin{align*} 
\min_{\{z_i^t,\,w_i^t\}}\quad \frac{1}{2}\Bigl\| 
 \mathbf{y} - \sum_{t \in \mathcal{T}} \sum_{i \in \mathcal{N}_t} {\mathbf{M}_i^t}\,w_i^t\Bigr\|_2^2 + \frac{1}{2\gamma}\,\sum_{t \in \mathcal{T}} \sum_{i\in \mathcal{N}_t} (w_i^t)^2 \\
\text{s.t.} \quad \sum_{j \in \mathcal{C}_i^t} z_j^t \ \leq \ |\mathcal{C}_i^t| (1-z_i^t),
\quad \forall \ i \in \mathcal{N}_t, \ t \in \mathcal{T},  \\
\sum_{t \in \mathcal{T}} \sum_{i \in \mathcal{N}_t} a_i^t\,z_i^t 
\leq K,  \\    (1 - z_i^t)\;w_i^t = 0, \quad
\forall \ i \in \mathcal{N}_t,\ t \in \mathcal{T}, \\ 
z_i^t \in \{0,1\}, \quad \forall 
\ i \in \mathcal{N}_t, \ t \in \mathcal{T}. 
\end{align*}

\noindent \textbf{Problem \eqref{CIP_reformulation.prob}}:
\begin{align*}
\min_{\mathbf{z}} \quad \frac{1}{2} \mathbf{y}^\top  \biggl(\mathbb{I}_{n} + \gamma \sum_{t \in \mathcal{T}} \sum_{i\in \mathcal{N}_t} z_{i}^t {\mathbf{M}_i^t}\bigl({\mathbf{M}_i^t}\bigr )^\top\biggr)^{-1} \mathbf{y}  \\
\text{s.t.} \quad \sum_{j \in \mathcal{C}_i^t} z_j^t \ \leq \ |\mathcal{C}_i^t| (1-z_i^t),
\quad \forall \ i \in \mathcal{N}_t, \ t \in \mathcal{T},  \\
\sum_{t \in \mathcal{T}} \sum_{i \in \mathcal{N}_t} a_i^t\,z_i^t 
\leq K, \\  
z_i^t \in \{0,1\}, \quad \forall 
\ i \in \mathcal{N}_t, \ t \in \mathcal{T}. 
\end{align*}
Note that we use $\mathbf{z}$ and $w$ to refer to the decision vectors formed by stacking all $z_i^t$ and $w_i^t$ decision variables for all $i \in \mathcal{N}_t$ and $t \in \mathcal{T}$.

\noindent Consider some solution $\mathbf{z}^*$ to Problem \eqref{CIP_reformulation.prob}, with objective value:
\begin{align*}
    \frac{1}{2} \mathbf{y}^\top \biggl(\mathbb{I}_{n} + \gamma \sum_{t \in \mathcal{T}} \sum_{i\in \mathcal{N}_t} (z_{i}^t)^* {\mathbf{M}_i^t}\bigl({\mathbf{M}_i^t}\bigr )^\top\biggr)^{-1} \mathbf{y} 
\end{align*}

\noindent Note that solution $\mathbf{z}^*$ satisfies the first, second, and fourth constraint in Problem \eqref{tree_based_problem1.prob}. We fix decision variable $\mathbf{z}$ to be $\mathbf{z}^*$ in Problem \eqref{tree_based_problem1.prob} and minimize with respect to $w$. This is given by:\begin{align} \label{appx_prob_1}
\min_{\{w_i^t\}}\quad \frac{1}{2}\Bigl\| 
 \mathbf{y} - \sum_{t \in \mathcal{T}} \sum_{i \in \mathcal{N}_t} {\mathbf{M}_i^t}\,w_i^t\Bigr\|_2^2 + \frac{1}{2\gamma}\,\sum_{t \in \mathcal{T}} \sum_{i\in \mathcal{N}_t} (w_i^t)^2 \\
\text{s.t.} \quad     (1 - (z_i^t)^*)\;w_i^t = 0, \quad
\forall \ i \in \mathcal{N}_t,\ t \in \mathcal{T}, \notag \\ 
(z_i^t)^* \in \{0,1\}, \quad \forall 
\ i \in \mathcal{N}_t, \ t \in \mathcal{T}. \notag
\end{align}

\noindent Let $\mathcal{M}$ denote the $n \times \bigl( \sum_{t \in \mathcal{T}} |\mathcal{N}_t| \bigr)$ matrix formed by taking the vectors ${\mathbf{M}_i^t}$ as its columns. Let $S = \{ i : z^*_i \neq 0 \}$, and let $\mathcal{M}_S$ be the sub-matrix of $\mathcal{M}$ formed by selecting the columns corresponding to indices in $S$. 

\noindent The non-zero coefficients, $w_S^*$, to the optimal solution of Problem \eqref{appx_prob_1}, $\mathbf{w}^*$, are given by:
\begin{align*}
    w_S^* = \biggl(\frac{\mathbbm{I}_{|S|}}{\gamma} + \mathcal{M}_S^\top \mathcal{M}_S \biggr)^{-1}\mathcal{M} ^\top \mathbf{y}.
\end{align*}

\noindent In addition:
\begin{equation} \label{appx_matrix_eq1}
      \mathcal{M}_S (\mathcal{M}_S)^\top = \sum_{t \in \mathcal{T}} \sum_{i\in \mathcal{N}_t} (z_{i}^t)^* {\mathbf{M}_i^t}\bigl({\mathbf{M}_i^t}\bigr )^\top.
 \end{equation}

\noindent We plug in $\mathbf{w}^*$ into the objective function for Problem \eqref{appx_prob_1} and apply the Woodbury matrix identity \citep{petersen2008matrix}. This yields:
\begin{align*}
    \quad \frac{1}{2}\Bigl\| 
 \mathbf{y} - \sum_{t \in \mathcal{T}} \sum_{i \in \mathcal{N}_t} {\mathbf{M}_i^t}\,(w_i^t)^*\Bigr\|_2^2 + \frac{1}{2\gamma}\,\sum_{t \in \mathcal{T}} \sum_{i\in \mathcal{N}_t} \bigl((w_i^t)^*\bigr)^2 =  \frac{1}{2} \mathbf{y}^\top \biggl(\mathbb{I}_{n} + \gamma \sum_{t \in \mathcal{T}} \sum_{i\in \mathcal{N}_t} (z_{i}^t)^* {\mathbf{M}_i^t}\bigl({\mathbf{M}_i^t}\bigr )^\top\biggr)^{-1} \mathbf{y} 
\end{align*}

\noindent Any solution $\mathbf{z}^*$ to Problem \eqref{CIP_reformulation.prob} corresponds to solution $(\mathbf{z}^*, \mathbf{w}^*)$ to Problem \eqref{tree_based_problem1.prob}, and the corresponding objective values are equal. 

\vspace{2mm}
\noindent The reverse statement also holds. Consider some solution $\mathbf{z}^*,\mathbf{w}^*$ to Problem \eqref{tree_based_problem1.prob}; $\mathbf{z}^*$ is a solution to Problem \eqref{CIP_reformulation.prob} and it is apparent that the objective values of these two solutions are identical. Therefore, Problem \eqref{CIP_reformulation.prob} is an equivalent reformulation of Problem \eqref{tree_based_problem1.prob}.

\subsubsection{Proof of Proposition \ref{prop2_opt.prop}}

We prove that function $\mathbf{z} \mapsto q(\mathbf{z})$ where $q(\mathbf{z})$ is convex on $\mathbf{z} \in [0,1]^m$. We have that:
\begin{align*}
    q(\mathbf{z}) = \frac{1}{2} \mathbf{y}^\top  \biggl(\mathbb{I}_{n} + \gamma \sum_{t \in \mathcal{T}} \sum_{i\in \mathcal{N}_t} z_{i}^t {\mathbf{M}_i^t}\bigl({\mathbf{M}_i^t}\bigr )^\top\biggr)^{-1} \mathbf{y} 
\end{align*}
We use this result;  Example 3.4, \cite{boyd2004convex} states that function, 
$$ f(\bm{\alpha}, A) = \bm{\alpha}^\top A^{-1} \bm{\alpha}$$
is convex in $(\bm{\alpha}, A)$ where $\bm{\alpha} \in \mathbb{R}^n$ and $A$ is a symmetric positive definite matrix of size $n \times n$. For our function $q(\mathbf{z})$, vector $y$ is fixed. Matrix $ \biggl(\mathbb{I}_{n} + \gamma \sum_{t \in \mathcal{T}} \sum_{i\in \mathcal{N}_t} z_{i}^t {\mathbf{M}_i^t}\bigl({\mathbf{M}_i^t}\bigr )^\top\biggr)$ is positive definite and an affine function of $\mathbf{z}$. Function $q(\mathbf{z})$ is a convex composition of an affine function and is therefore convex over $\mathbf{z}$.

\subsection{Objective and Subgradient Evaluation} \label{obj_subgrad_optimal.appx}

We present here an efficient procedure to evaluate objective $q(\mathbf{z})$ and subgradient $\nabla q(\mathbf{z})$ for any solution $\mathbf{z}$. Recall that
\begin{align*}
    q(\mathbf{z}) = \frac{1}{2} \mathbf{y}^\top  \biggl(\mathbb{I}_{n} + \gamma \sum_{t \in \mathcal{T}} \sum_{i\in \mathcal{N}_t} z_{i}^t {\mathbf{M}_i^t}\bigl({\mathbf{M}_i^t}\bigr )^\top\biggr)^{-1} \mathbf{y} ,
\end{align*}
where, as previously defined, $\mathcal{M}$ is the $n \times \bigl( \sum_{t \in \mathcal{T}} |\mathcal{N}_t| \bigr)$ matrix formed by taking the vectors ${\mathbf{M}_i^t}$ as its columns. Again, $S = \{ i : z_i \neq 0 \}$, $\mathcal{M}_S$ is the sub-matrix of $\mathcal{M}$ consisting of the columns indexed by $S$. 
\noindent Applying \eqref{appx_matrix_eq1} we have that:
\begin{align*}
    q(\mathbf{z}) = \frac{1}{2} \mathbf{y}^\top  \biggl(\mathbb{I}_{n} + \gamma \mathcal{M}_S (\mathcal{M}_S)^\top\biggr)^{-1} \mathbf{y} 
\end{align*}

\noindent We apply the Woodbury matrix identity \citep{petersen2008matrix}  to obtain:
\begin{align}
q(\mathbf{z}) = \frac{1}{2} \left( \mathbf{y}^\top  \mathbf{y} - \mathbf{y}^\top  \mathcal{M}_S \left( \frac{\mathbb{I}_{|S|}}{\gamma} + \mathcal{M}_S^\top \mathcal{M}_S \right)^{-1} \mathcal{M}_S^\top \mathbf{y} \right).
\end{align}

\noindent We can evaluate this expression efficiently by taking the LU decomposition of matrix $(\frac{\mathbb{I}_{|S|}}{\gamma} + \mathcal{M}_S^\top \mathcal{M}_S)$.

\vspace{2mm} \noindent
To compute sub-gradient $\nabla q(\mathbf{z})$, we have that:
\begin{align*}
\frac{\partial q(\mathbf{z})}{\partial z_{i'}^{t'}} = \frac{1}{2} \mathbf{y}^\top  \frac{\partial}{\partial z_{i'}^{t'}} \left( \left( \mathbb{I}_n + \gamma \sum_{t \in \mathcal{T}} \sum_{i \in \mathcal{N}_t} z_i^t {\mathbf{M}_i^t}({\mathbf{M}_i^t})^\top \right)^{-1} \right) y.
\end{align*}

\noindent We apply the matrix inverse derivative  identity \citep{petersen2008matrix} and \eqref{appx_matrix_eq1} to obtain:
\begin{align*}
\frac{\partial q(\mathbf{z})}{\partial z_{i'}^{t'}} = -\frac{\gamma}{2} \cdot \left( (M_{i'}^{t'})^\top \bigl( \mathbb{I}_n + \gamma \mathcal{M}_S (\mathcal{M}_S)^\top \bigr)^{-1} y \right)^2.
\end{align*}

\noindent Applying the Woodbury matrix identity \citep{petersen2008matrix} again yields:
\begin{align}
\frac{\partial q(\mathbf{z})}{\partial z_{i'}^{t'}} = -\frac{\gamma}{2} \cdot \left( (M_{i'}^{t'})^\top \cdot \left(\mathbf{y} - \mathcal{M}_S \left( \frac{\mathbb{I}_{|S|}}{\gamma} + \mathcal{M}_S^\top \mathcal{M}_S \right)^{-1} \mathcal{M}_S^\top \mathbf{y} \right) \right)^2, \quad  \forall \ i' \in \mathcal{N}_t, \ t' \in \mathcal{T},
\end{align}
 and we can again evaluate this expression efficiently by taking the LU decomposition of matrix $(\frac{\mathbb{I}_{|S|}}{\gamma} + \mathcal{M}_S^\top \mathcal{M}_S)$.

\color{black}
\subsection{Convergence of Optimal Algorithm}

\begin{proposition}
\label{prop:OA_convergence}
Algorithm~\ref{OA_main.alg} converges to the optimal solution of 
Problem~\eqref{CIP_reformulation.prob} in a finite number of iterations.
\end{proposition}

% \noindent\textbf{Proof:}
% In iteration $h$, Algorithm~\eqref{OA_main.alg} adds the constraint
% \[
% \nu^{(h)} \;\ge\; q(\mathbf z^{(h)}) \;+\; \nabla q(\mathbf z^{(h)})^\top\!\big(\mathbf z-\mathbf z^{(h)}\big).
% \] For any subsequent iteration $h' > h$ evaluating the constraint at  $z = z^{(h)}$ implies that
% $\nu^{(h')} \geq q(\mathbf z^{(h)})$. Suppose $z^{(h)})$ is not optimal. Assume for the sake of contradiction that $\mathbf z^{(h)}$ is revisited in subsequent iteration $h'$. Since $\mathbf z^{(h)}$ is not optimal, we have that $\nu^{(h')}< q(\mathbf z^{(h)})$ which contradicts $\nu^{(h')} \geq q(\mathbf z^{(h)})$. Therefore, each in each iteration $h$, solution $\mathbf z^{(h)}$ is excluded if it is not optimal. Since $\mathbb C_3$ is finite, only finitely many non-optimal points can be selected, and the algorithm terminates after finitely many iterations. At convergence, we have that $\nu^{(h)} \geq q(\mathbf z^{(h)})$ and since $\nu^{(h)} \;\le\; \min_{\mathbf z\in\mathbb C_3} q(\mathbf z) \;\le\; q(\mathbf z^{(h)})$ we have that  $\nu^{(h)}=q(\mathbf z^{(h)})=\min_{\mathbf z\in\mathbb C_3} q(\mathbf z)$, and the returned $\mathbf z^{(h)}$ is globally optimal. \qed

\noindent\textbf{Proof:} This follows from \citet{duran1986outer}.
In iteration $h$, Algorithm~\eqref{OA_main.alg} adds the constraint
\[
\nu^{(h)} \;\ge\; q(\mathbf z^{(h)}) \;+\; \nabla q(\mathbf z^{(h)})^\top\!\big(\mathbf z-\mathbf z^{(h)}\big).
\]
For any subsequent iteration $h' > h$, evaluating this constraint at $\mathbf z=\mathbf z^{(h)}$ implies
$\nu^{(h')} \ge q(\mathbf z^{(h)})$.
Suppose $\mathbf z^{(h)}$ is not optimal. Assume for contradiction that $\mathbf z^{(h)}$ is revisited in some iteration $h'>h$.
We have
\[
\nu^{(h')} \;\le\; \min_{\mathbf z\in\mathbb C_3} q(\mathbf z) \;<\; q(\mathbf z^{(h)}),
\]
where the strict inequality uses that $\mathbf z^{(h)}$ is not optimal. This contradicts $\nu^{(h')} \ge q(\mathbf z^{(h)})$.
Therefore, at each iteration $h$, the point $\mathbf z^{(h)}$ is excluded from future selection unless it is optimal.
Since $\mathbb C_3$ is finite, only finitely many non-optimal points can be selected, and the algorithm terminates after finitely many iterations.
At convergence, we have $\nu^{(h)} \ge q(\mathbf z^{(h)})$, and since $\nu^{(h)} \le \min_{\mathbf z\in\mathbb C_3} q(\mathbf z) \le q(\mathbf z^{(h)})$, it follows that
$\nu^{(h)}=q(\mathbf z^{(h)})=\min_{\mathbf z\in\mathbb C_3} q(\mathbf z)$, so the returned $\mathbf z^{(h)}$ is globally optimal. \qed

\color{black}

\section{Approximate Algorithm}

We present here technical details regarding our approximate algorithm.

\subsection{CBCD Algorithm} \label{approx_alg_prop.appx} We show proofs and additional details regarding our efficient block update procedure in our CBCD algorithm.

\subsubsection{Proof of Proposition \ref{approx_alg_CBCD.prop}}

We first prove Problem \eqref{blockupdate_CIP.prob} is an equivalent reformulation of  Problem \eqref{blockupdate.prob}; we display the problems below.

\vspace{2mm}
\noindent \textbf{Problem \eqref{blockupdate.prob}}:

\begin{align*}
\min_{\mathbf{z}^t,\,w^t}\quad \frac{1}{2}\Bigl\| 
  r -  \sum_{i \in \mathcal{N}_t} {\mathbf{M}_i^t}\,w_i^t\Bigr\|_2^2 + \frac{1}{2\gamma}\, \sum_{i\in \mathcal{N}_t} (w_i^t)^2 + \lambda  \sum_{i \in \mathcal{N}_t} a_i^t\,z_i^t  \\
\text{s.t.} \quad \sum_{j \in \mathcal{C}_i^t} z_j^t \ \leq \ |\mathcal{C}_i^t| (1-z_i^t),
\quad
   (1 - z_i^t)\;w_i^t = 0, \quad 
z_i^t \in \{0,1\}, \quad \forall 
\ i \in \mathcal{N}_t, \notag
\end{align*}

\vspace{2mm}
\noindent \textbf{Problem \eqref{blockupdate_CIP.prob}}:
\begin{align*} 
  \min_{\mathbf{z}^t} \ \frac{1}{2} \mathbf{r}^\top \biggl(\mathbb{I}_{n} + \gamma \sum_{i\in \mathcal{N}_t} z_{i}^t {\mathbf{M}_i^t}\bigl({\mathbf{M}_i^t}\bigr )^\top\biggr)^{-1} \mathbf{r}  + \lambda  \sum_{i \in \mathcal{N}_t} a_i^t\,z_i^t\\
\text{s.t.} \quad \sum_{j \in \mathcal{C}_i^t} z_j^t \ \leq \ |\mathcal{C}_i^t| (1-z_i^t),
\quad z_i^t \in \{0,1\}, \quad \forall 
\ i \in \mathcal{N}_t. \notag
\end{align*}

\noindent Our proof follows closely the proof
presented in \S\ref{optimal_alg_prop.appx}. Consider some solution $(\mathbf{z}^t)^*$ to Problem \eqref{blockupdate_CIP.prob} with objective value:
\begin{align*}
\frac{1}{2} \mathbf{r}^\top \biggl(\mathbb{I}_{n} + \gamma \sum_{i\in \mathcal{N}_t} (z_{i}^t)^* {\mathbf{M}_i^t}\bigl({\mathbf{M}_i^t}\bigr )^\top\biggr)^{-1} \mathbf{r}  + \lambda  \sum_{i \in \mathcal{N}_t} a_i^t\, (z_i^t)^*,
\end{align*}
and note that $(\mathbf{z}^t)^*$ satisfies the first and third constraints in Problem \eqref{blockupdate.prob}.

\noindent We fix decision vector $\mathbf{z}^t$ to be 
$(\mathbf{z}^t)^*$ in Problem \eqref{blockupdate.prob} and minimize with respect to $w^t$. This is given by:
\begin{align} \label{cbcd_prob1_appx}
\min_{\,w^t}\quad \frac{1}{2}\Bigl\| 
  \mathbf{r} -  \sum_{i \in \mathcal{N}_t} {\mathbf{M}_i^t}\,w_i^t\Bigr\|_2^2 + \frac{1}{2\gamma}\, \sum_{i\in \mathcal{N}_t} (w_i^t)^2 + \lambda  \sum_{i \in \mathcal{N}_t} a_i^t\,(z_i^t)^*  \\
\text{s.t.} \quad \sum_{j \in \mathcal{C}_i^t} (z_i^t)^* \ \leq \ |\mathcal{C}_i^t| (z_i^t)^*,
\quad
   (1 - (z_i^t)^*)\;w_i^t = 0, \quad 
(z_i^t)^* \in \{0,1\}, \quad \forall 
\ i \in \mathcal{N}_t, \notag
\end{align}

\noindent Recall that $\mathcal{M}^t$ is the $n \times |\mathcal{N}_t|$ matrix with columns ${\mathbf{M}_i^t}, \ \forall \ i \in \mathcal{N}_t$, $S = \{ i : (z^t_i)^* \neq 0 \}$, and  $\mathcal{M}^t_S$ is the sub-matrix of $\mathcal{M}^t$ with columns indexed by $S$. We have that the non-zero coefficient $(w_S^t)^*$ of the optimal solution $(\mathbf{w}^t)^*$ to Problem \eqref{cbcd_prob1_appx} is given by:
\begin{align*}
    (w^t_S)^* = \biggl(\frac{\mathbbm{I}_{|S|}}{\gamma} + (\mathcal{M}^t_S)^\top \mathcal{M}^t_S\biggr)^{-1}(\mathcal{M}^t_S)^\top \mathbf{y},
\end{align*}
and that:
\begin{equation} \label{appx_matrix_eq2}
      \mathcal{M}^t_S(\mathcal{M}^t_S)^\top = \sum_{i\in \mathcal{N}_t} (z_{i}^t)^* {\mathbf{M}_i^t}\bigl({\mathbf{M}_i^t}\bigr )^\top.
 \end{equation}

 \noindent Plugging $(\mathbf{w}^t)^*$ in to the objective function of \eqref{cbcd_prob1_appx} and applying the Woodbury matrix identity \citep{petersen2008matrix}  yields:

 \begin{align*}
     \frac{1}{2}\Bigl\| 
  \mathbf{r} -  \sum_{i \in \mathcal{N}_t} {\mathbf{M}_i^t}\,(w_i^t)^*\Bigr\|_2^2 + \frac{1}{2\gamma}\, \sum_{i\in \mathcal{N}_t} ((w_i^t)^*)^2 + \lambda  \sum_{i \in \mathcal{N}_t} a_i^t\,(z_i^t)^*  \\
  = \frac{1}{2} \mathbf{r}^\top \biggl(\mathbb{I}_{n} + \gamma \sum_{i\in \mathcal{N}_t} (z_{i}^t)^* {\mathbf{M}_i^t}\bigl({\mathbf{M}_i^t}\bigr )^\top\biggr)^{-1} \mathbf{r}  + \lambda  \sum_{i \in \mathcal{N}_t} a_i^t\, (z_i^t)^*
 \end{align*}
\noindent As such, any solution $(\mathbf{z}^t)^*$ to Problem \eqref{blockupdate_CIP.prob} corresponds to solution $((\mathbf{w}^t)^*,(\mathbf{z}^t)^*)$ to Problem \eqref{blockupdate.prob}, with an identical objective value. The reverse statement is apparent so Problem \eqref{blockupdate_CIP.prob} is an equivalent reformulation of Problem \eqref{blockupdate.prob}.

\vspace{2mm} \noindent Next, we show that objective function $\mathbf{z}^t \mapsto q_{\mathbf{r}}(\mathbf{z}^t)$ is convex on $\mathbf{z}^t \in [0,1]^{|\mathcal{N}_t|}$. We have that:
\begin{align*}
    q_{\mathbf{r}}(\mathbf{z}^t) = \frac{1}{2} \mathbf{r}^\top \biggl(\mathbb{I}_{n} + \gamma \sum_{i\in \mathcal{N}_t} z_{i}^t {\mathbf{M}_i^t}\bigl({\mathbf{M}_i^t}\bigr )^\top\biggr)^{-1} \mathbf{r}  + \lambda  \sum_{i \in \mathcal{N}_t} a_i^t\,z_i^t.
\end{align*}
Note that $r$ is fixed and that matrix $\biggl(\mathbb{I}_{n} + \gamma \sum_{i\in \mathcal{N}_t} z_{i}^t {\mathbf{M}_i^t}\bigl({\mathbf{M}_i^t}\bigr )^\top\biggr)$ is positive definite. Using the result from Example 3.4 in \cite{boyd2004convex} we have that the first term is convex over $\mathbf{z}$. The second term is linear over $\mathbf{z}$, so $q_{\mathbf{r}}(\mathbf{z}^t)$ is convex over $\mathbf{z}$.

\subsubsection{Objective and Subgradient Evaluation}

\noindent We present here a procedure to efficiently evaluate objective $q_{\mathbf{r}}(\mathbf{z}^t)$ and subgradient $\nabla q_{\mathbf{r}}(\mathbf{z}^t)$, where:
\begin{align*}
    q_{\mathbf{r}}(\mathbf{z}^t) = \frac{1}{2} \mathbf{r}^\top \biggl(\mathbb{I}_{n} + \gamma \sum_{i\in \mathcal{N}_t} z_{i}^t {\mathbf{M}_i^t}\bigl({\mathbf{M}_i^t}\bigr )^\top\biggr)^{-1} \mathbf{r}  + \lambda  \sum_{i \in \mathcal{N}_t} a_i^t\,z_i^t \\ =\frac{1}{2} \mathbf{r}^\top \biggl(\mathbb{I}_{n} + \gamma \mathcal{M}^t_S(\mathcal{M}^t_S)^\top \biggr)^{-1} \mathbf{r}  + \lambda  \sum_{i \in \mathcal{N}_t} a_i^t\,z_i^t 
\end{align*}

\noindent Applying the Woodbury matrix identity \citep{petersen2008matrix} yields:
\begin{align} \label{block_update_obj}
q_{\mathbf{r}}(\mathbf{z}^t) = \frac{1}{2} \left( \mathbf{r}^\top \mathbf{r} - \mathbf{r}^\top \mathcal{M}^t_S \left( \frac{\mathbb{I}_{|S|}}{\gamma} + (\mathcal{M}^t_S)^\top \mathcal{M}^t_S \right)^{-1} (\mathcal{M}^t_S)^\top \mathbf{r} \right) + \lambda  \sum_{i \in \mathcal{N}_t} a_i^t\,z_i^t ,
\end{align}
and we can compute this efficiently by taking the LU decomposition of $\left( \frac{\mathbb{I}_{|S|}}{\gamma} + (\mathcal{M}^t_S)^\top \mathcal{M}^t_S \right)$.

\vspace{2mm}
\noindent For gradient $\nabla q_{\mathbf{r}}(\mathbf{z}^t)$ we have that:
\begin{align*} \label{block_update_subgradient}
\frac{\partial q_{\mathbf{r}}(\mathbf{z}^t)}{\partial z_{i'}^{t}} = \frac{1}{2} \mathbf{r}^\top \frac{\partial}{\partial z_{i'}^{t}} \left( \left( \mathbb{I}_n + \gamma  \sum_{i \in \mathcal{N}_t} z_i^t {\mathbf{M}_i^t}({\mathbf{M}_i^t})^\top \right)^{-1} \right) r + \lambda a_{i'}^t.
\end{align*} We apply the matrix inverse derivative identity and the Woodbury matrix identity \citep{petersen2008matrix} to obtain:
\begin{align}
\frac{\partial q_{\mathbf{r}}(\mathbf{z}^t)}{\partial z_{i'}^{t}} = -\frac{\gamma}{2} \cdot \left( (M_{i'}^{t})^\top \cdot \left( \mathbf{r} - \mathcal{M}^t_S \left( \frac{\mathbb{I}_{|S|}}{\gamma} + (\mathcal{M}^t_S)^\top \mathcal{M}^t_S \right)^{-1} (\mathcal{M}^t_S)^\top \mathbf{r} \right) \right)^2 + \lambda a_{i'}^t   \quad  \forall \ i' \in \mathcal{N}_t,
\end{align}
which we can compute efficiently by taking the LU decomposition of $\left( \frac{\mathbb{I}_{|S|}}{\gamma} + (\mathcal{M}^t_S)^\top \mathcal{M}^t_S \right)$.

\subsubsection{Cutting Plane Recycling} \label{cutting_plane_recycle.appx}

We further discuss here our procedure to recycle cutting planes. We consider two subsequent updates of tree $t \in \mathcal{T}$ in our CBCD procedure. During the first update, say we obtain feasible solution  $\mathbf{z}^t$ to Problem \eqref{blockupdate_CIP.prob} and add cutting plane:
\begin{align*}
\nu \geq q_{\mathbf{r}}(\mathbf{z}^t)+ \nabla q_{\mathbf{r}}(\mathbf{z}^t)^{\top}(\mathbf{z}-\mathbf{z}^t),
\end{align*}
where $q_{\mathbf{r}}(\mathbf{z}^t)$ is given by \eqref{block_update_obj} and $\nabla q_{\mathbf{r}}(\mathbf{z}^t)$ is given by \eqref{block_update_subgradient}. We cache the LU decomposition of: $
    \left( \frac{\mathbb{I}_{|S|}}{\gamma} + (\mathcal{M}^t_S)^\top \mathcal{M}^t_S \right)$ during these evaluations.

\noindent Now, consider the second subsequent update of tree $t \in \mathcal{T}$ in our CBCD procedure. The only change to Problem \eqref{blockupdate_CIP.prob} is that residual vector changes to $\mathbf{r}'$; solution $\mathbf{z}^t$ is still feasible. We want to add cutting plane:
\begin{align*}
\nu \geq q_{\mathbf{r}'}(\mathbf{z}^t)+ \nabla q_{\mathbf{r}'}(\mathbf{z}^t)^{\top}(\mathbf{z}-\mathbf{z}^t),
\end{align*} which requires that we evaluate 
$q_{\mathbf{r}'}(\mathbf{z}^t)$ and $\nabla q_{\mathbf{r}'}(\mathbf{z}^t)$ using \eqref{block_update_obj} and \eqref{block_update_subgradient}. These evaluations require solving the system:
\begin{align*}
    \left( \frac{\mathbb{I}_{|S|}}{\gamma} + (\mathcal{M}^t_S)^\top \mathcal{M}^t_S \right) v = (\mathcal{M}^t_S)^\top \mathbf{r}
',
\end{align*} for $v$, which can be done extremely efficiently since we cache the LU decomposition of $\left( \frac{\mathbb{I}_{|S|}}{\gamma} + (\mathcal{M}^t_S)^\top \mathcal{M}^t_S \right).$ 

\noindent We repeat this cutting plane recycling procedure across all CBCD updates for all solutions $\mathbf{z}^t$ encountered.

\subsection{Comparisons Between Optimal and Approximate Algorithm} \label{opt_v_approx.appx}
\begin{figure}[h]
    \centering
    \includegraphics[width=0.8\linewidth]{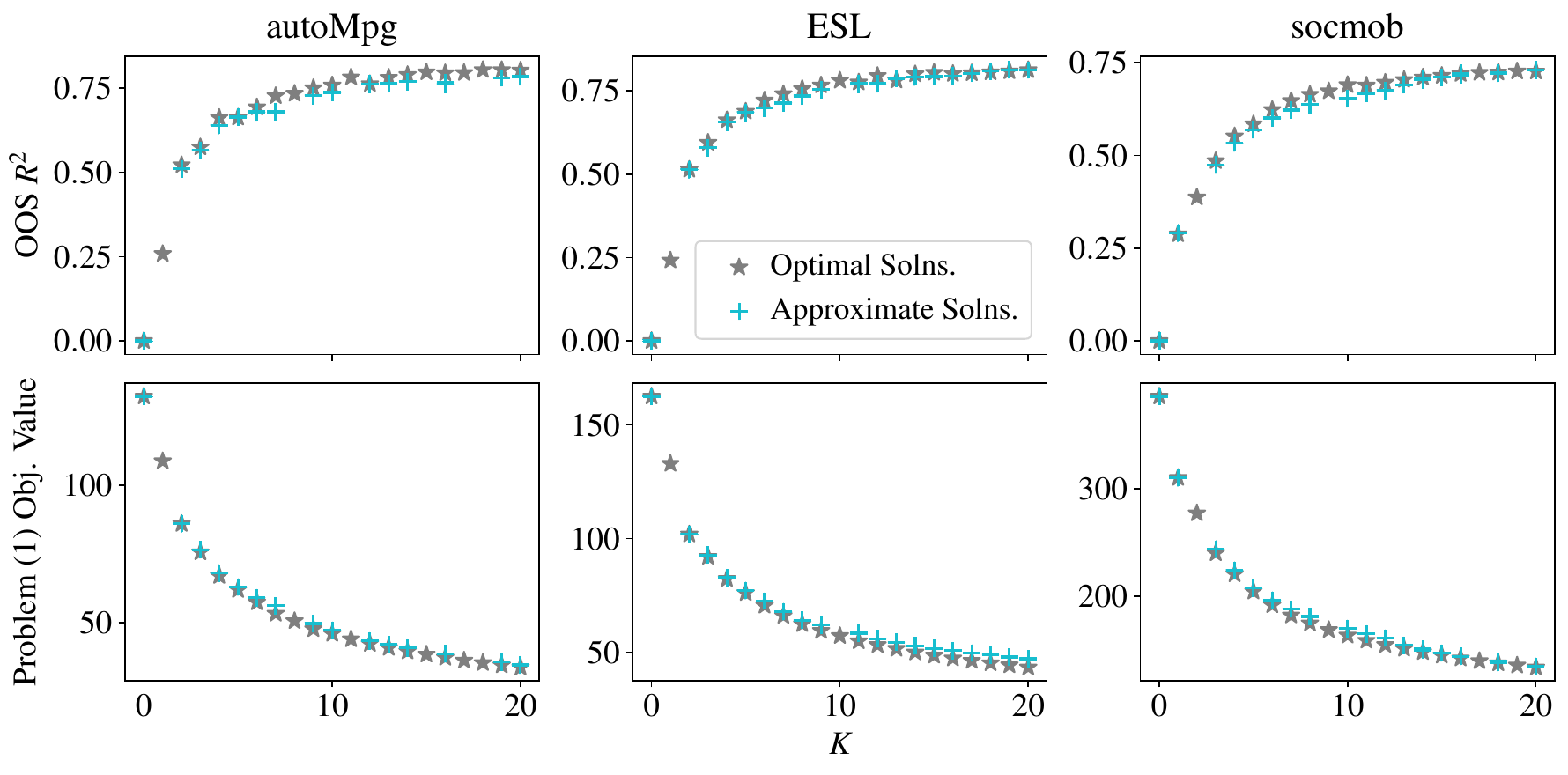}
    \caption{Comparison of solutions obtained by our optimal and approximate algorithms. The top plots show out-of-sample $R^2$ vs. $K$ and the bottom plots show the objective value of Problem \eqref{tree_based_problem1.prob} for each solution.}
    \label{opt_v_approx.fig}
\end{figure}

 In Figure~\ref{opt_v_approx.fig}, we apply our exact Algorithm~\ref{OA_main.alg} and approximate Algorithm~\ref{cbcd.alg} to prune tree ensembles consisting of 100 depth 3 decision trees, fit on the \textsc{autoMpg}, \textsc{ESL}, and \textsc{Socmob} datasets obtained from the OpenML repository \citep{bischl2017openml}. For both algorithms, we compute solutions along the regularization path for $K \in \{1,2,…,20\}$. The top row of plots in Figure~\ref{opt_v_approx.fig} shows the out-of-sample $R^2$ scores of the solutions as a function of $K$; optimal solutions are marked by grey stars, and approximate solutions are indicated by teal crosses. The bottom row of plots shows the corresponding objective value of Problem~\eqref{tree_based_problem1.prob} for each solution. We observe that the solutions produced by our approximate algorithm are slightly worse in terms of both out-of-sample accuracy and the objective value of Problem~\eqref{tree_based_problem1.prob}. Nevertheless, the approximate solutions are sufficiently close to optimal solutions to remain useful in practice. Finally, because CBCD Algorithm~\ref{cbcd.alg} can become stuck in local minima, our approximate algorithm may occasionally fail to yield a solution for every model size $K$ along the regularization path, a limitation that our exact algorithm avoids.

\section{Exploring the Convex Relaxation} \label{exploring_convex.appx}

As an aside, we discuss a interesting new convex estimator that is of independent interest. Consider this optimization problem: \begin{align} \label{CIP_penalized.prob}
\min_{\mathbf{z}}\quad \frac{1}{2} \mathbf{y}^\top  \biggl(\mathbb{I}_{n} + \gamma \sum_{t \in \mathcal{T}} \sum_{i\in \mathcal{N}_t} z_{i}^t {\mathbf{M}_i^t}\bigl({\mathbf{M}_i^t}\bigr )^\top\biggr)^{-1}\mathbf{y} + \lambda\sum_{t \in \mathcal{T}} \sum_{i \in \mathcal{N}_t} a_i^t\,z_i^t \\
\text{s.t.} \quad \sum_{j \in \mathcal{C}_i^t} z_j^t \ \leq \ |\mathcal{C}_i^t| (1-z_i^t),
\quad \forall \ i \in \mathcal{N}_t, \ t \in \mathcal{T}, \tag{Constraint \ref{CIP_penalized.prob}a} \\
z_i^t \in \{0,1\}, \quad \forall 
\ i \in \mathcal{N}_t, \ t \in \mathcal{T}.\tag{Constraint \ref{CIP_penalized.prob}b}
\end{align} We show in \S\ref{ws_prob_reform.appx} of the  appendix that Problem \eqref{CIP_penalized.prob} is a convex integer program and is an equivalent reformulation of Problem \eqref{tree_based_penalized.prob}. Again, note that constraints (\ref{CIP_penalized.prob}a) and (\ref{CIP_penalized.prob}b) are separable across trees $t \in \mathcal{T}$. 

\vspace{2mm}
\noindent{\textbf{Convex Relaxation :}} Consider a convex relaxation of Problem \eqref{CIP_penalized.prob}, where we relax and replace integrality constraint (\ref{CIP_penalized.prob}b) with $0\leq z_i^t \leq 1, \ \forall 
\ i \in \mathcal{N}_t, \ t \in \mathcal{T}$. We can solve the resulting convex, semidefinite program using an efficient tailored algorithm that we present in \S\ref{lr_alg.appx} of the appendix. 
% Our tailored algorithm only requires a linear programming solver for which many high-performing open-source options are readily available \citep{matthew2018coin,huangfu2018parallelizing}.
Let $\zeta_i^t, \ \forall \ i \in \mathcal{N}_t, \ t \in \mathcal{T}$ be the resulting \emph{fractional} optimal solution. We find solutions $z_i^{t}, \ \forall \ i \in \mathcal{N}_t, \ t \in \mathcal{T}$ that are feasible for Problem \eqref{CIP_penalized.prob} and close to $\zeta_i^t$ using the rounding procedure below.

\begin{algorithm}[h] 
    \caption{ Relax and Round Algorithm for Problem \eqref{CIP_penalized.prob}} \label{LRround.alg} 
solve linear relaxation of Problem \eqref{CIP_penalized.prob} for fractional  $\zeta_i^t, \ \forall \ i \in \mathcal{N}_t, \ t \in \mathcal{T}$.

    \For{$t \in \mathcal{T}$}{
    solve ILP \eqref{roundingILP.prob} for integral solutions $z_i^t, \ \forall \ i \in \mathcal{N}_t$
    }
    \textbf{return} $\mathbf{z}$
  \end{algorithm}
  
\vspace{2mm}
\noindent{\textbf{Rounding Procedure fo Feasible Solutions}} Let solutions $z_i^{t}$ represent which elements of $\zeta_i^t$ to round to 1, for all $i \in \mathcal{N}_t, \ t \in \mathcal{T}$. Note that constraint (\ref{CIP_penalized.prob}a) separates across trees $t \in \mathcal{T}$. For each tree $t$, we want to round the largest elements of $\zeta_i^t, \ \forall \ i \in \mathcal{N}_t$ to 1, such that constraint (\ref{CIP_penalized.prob}a) is satisfied. We do so by solving the following integer linear program:
\begin{align} \label{roundingILP.prob} \sum_{i \in \mathcal{N}_t}\max_{z_i^t} \quad \zeta_i^t z_i^t, \quad 
\text{s.t.} \quad \sum_{j \in \mathcal{C}_i^t} z_j^t \ \leq \ |\mathcal{C}_i^t| (1-z_i^t), \quad
z_i^t \in \{0,1\}, \quad \forall 
\ i \in \mathcal{N}_t.
\end{align}
We repeat this rounding procedure for all trees $t \in \mathcal{T}$ to obtain our final feasible solution to Problem \eqref{CIP_penalized.prob}. We present our full relax and round procedure in Algorithm \ref{LRround.alg}, which returns integral solutions  $z_i^t, \ \forall \ i \in \mathcal{N}_t, \ t \in \mathcal{T}$ feasible for Problem \eqref{CIP_penalized.prob}.

\noindent{\textbf{Solving on ILP} \eqref{roundingILP.prob}:} We exploit problem structure in order to solve ILP \eqref{roundingILP.prob} efficiently. In the proposition below, we establish two important properties of the problem.
\begin{proposition} \label{KPCG.prop}
    ILP \eqref{roundingILP.prob} is a knapsack problem with conflict graph $G$ = ($V$,$E$) where $V = \mathcal{N}_t$ and $E = \bigl\{(i,j) \ \forall \ i \in \mathcal{N}_t, \ j \in \mathcal{C}_i^t \bigr\}$. Conflict graph $G$ is chordal.
\end{proposition}

We prove this proposition in \S\ref{knapsack.appx} of the appendix. As a result of Proposition \ref{KPCG.prop}, we can apply \textsc{AlgCH} from \cite{pferschy2009knapsack} to solve Problem \eqref{roundingILP.prob} to optimality in pseudo-polynomial time. We can also use the associated  polynomial-time approximation scheme  to find good solutions efficiently.

% Consequently, Algorithm \ref{LRround.alg} can function efficiently in pseudo-polynomial or polynomial time without relying on commercial solvers.
% \subsection{Iterative Algorithm: Block Coordinate Descent}
% Consider this penalized reformulation of MIP Problem \eqref{tree_based_problem1.prob}:
% \begin{align} \label{tree_based_penalized.prob}
% \min_{\{z_i^t,\,w_i^t\}}\quad \frac{1}{2}\Bigl\| 
%   y - \sum_{t \in \mathcal{T}} \sum_{i \in \mathcal{N}_t} {\mathbf{M}_i^t}\,w_i^t\Bigr\|_2^2 + \frac{1}{2\gamma}\,\sum_{t \in \mathcal{T}} \sum_{i\in \mathcal{N}_t} (w_i^t)^2 + \lambda \sum_{t \in \mathcal{T}} \sum_{i \in \mathcal{N}_t} a_i^t\,z_i^t  \\
% \text{s.t.} \quad \sum_{j \in \mathcal{C}_i^t} z_j^t \ \leq \ |\mathcal{C}_i^t| (1-z_i^t), \quad (1 - z_i^t)\;w_i^t = 0, \quad  z_i^t \in \{0,1\}, \quad \forall 
% \ i \in \mathcal{N}_t, \ t \in \mathcal{T}, \notag
% \end{align}
% and note that this is an equivalent reformulation of CIP Problem \eqref{CIP_penalized.prob}; we show this in \S\ref{} of the appendix. The first term in the objective is smooth with respect to $w$ and the second and third terms are separable with respect to trees $t \in \mathcal{T}$. The constraints are also all separable with respect to trees $t \in \mathcal{T}$. Motivated by the success of block coordinate descent algorithms for high-dimensional regression \citep{hazimeh2020fast,hazimeh2023grouped}, we apply the following block coordinate descent algorithm to find high-quality solutions to Problem \eqref{tree_based_penalized.prob}.
\begin{figure}[h]
    \centering
    \includegraphics[width=\linewidth]{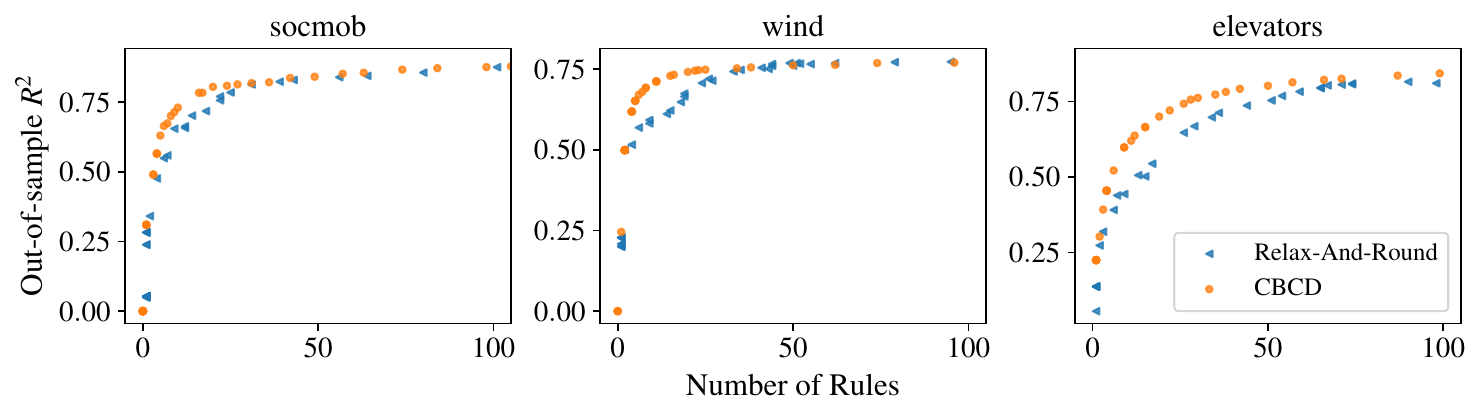}
    \caption{Comparison of regularization paths obtained by our relax-and-round algorithm and CBCD. CBCD typically obtains higher quality solutions compared to relax-and-round.}
    \label{cbcd_vs_lr.fig}
\end{figure}

\noindent{\textbf{Discussion:}} Our relax-and-round procedure has either a pseudo-polynomial or polynomial time complexity, depending on the rounding method used. However, we note that the solutions obtained through the relax-and-round procedure are typically of lower quality than those produced by our CBCD approximate algorithm. In Figure~\ref{cbcd_vs_lr.fig}, we compare the performance of relax-and-round and CBCD by computing regularization paths on three examples: ensembles of 500 depth-7 trees trained on the \textsc{Socmob}, \textsc{Wind}, and \textsc{Elevators} datasets from the OpenML benchmark repository~\cite{bischl2017openml}. From these plots, we observe that our CBCD approximate algorithm noticeably outperforms relax-and-round in terms of predictive accuracy.

\subsection{Problem Reformulation} \label{ws_prob_reform.appx}

We show here that Problem \eqref{CIP_penalized.prob} is an equivalent CIP reformulation of Problem \eqref{tree_based_penalized.prob}. We display both of these problems below.

\noindent \textbf{Problem} \eqref{tree_based_penalized.prob}:

\begin{align*} 
\min_{\{z_i^t,\,w_i^t\}}\quad \frac{1}{2}\Bigl\| 
 \mathbf{y} - \sum_{t \in \mathcal{T}} \sum_{i \in \mathcal{N}_t} {\mathbf{M}_i^t}\,w_i^t\Bigr\|_2^2 + \frac{1}{2\gamma}\,\sum_{t \in \mathcal{T}} \sum_{i\in \mathcal{N}_t} (w_i^t)^2 + \lambda \sum_{t \in \mathcal{T}} \sum_{i \in \mathcal{N}_t} a_i^t\,z_i^t  \\
\text{s.t.} \quad \sum_{j \in \mathcal{C}_i^t} z_j^t \ \leq \ |\mathcal{C}_i^t| (1-z_i^t), \quad \forall 
\ i \in \mathcal{N}_t, \ t \in \mathcal{T}   \\ \quad (1 - z_i^t)\;w_i^t = 0, \quad \forall 
\ i \in \mathcal{N}_t, \ t \in \mathcal{T} \\ \quad  z_i^t \in \{0,1\}, \quad \forall 
\ i \in \mathcal{N}_t, \ t \in \mathcal{T}, 
\end{align*}

\noindent \textbf{Problem} \eqref{CIP_penalized.prob}: \begin{align*} 
\min_{\mathbf{z}}\quad \frac{1}{2} \mathbf{y}^\top  \biggl(\mathbb{I}_{n} + \gamma \sum_{t \in \mathcal{T}} \sum_{i\in \mathcal{N}_t} z_{i}^t {\mathbf{M}_i^t}\bigl({\mathbf{M}_i^t}\bigr )^\top\biggr)^{-1} \mathbf{y}  + \lambda\sum_{t \in \mathcal{T}} \sum_{i \in \mathcal{N}_t} a_i^t\,z_i^t \\
\text{s.t.} \quad \sum_{j \in \mathcal{C}_i^t} z_j^t \ \leq \ |\mathcal{C}_i^t| (1-z_i^t),
\quad \forall \ i \in \mathcal{N}_t, \ t \in \mathcal{T},  \\
z_i^t \in \{0,1\}, \quad \forall 
\ i \in \mathcal{N}_t, \ t \in \mathcal{T}.
\end{align*}

\vspace{2mm}
\noindent The proof that \eqref{CIP_penalized.prob} is an equivalent reformulation of \eqref{tree_based_penalized.prob} follows directly from the proof of Proposition \ref{prop1_opt.prop}, presented in \S\ref{prop1_proof.appx}.

\vspace{2mm}
\noindent The objective of IP \eqref{CIP_penalized.prob}: \begin{align*}
    \frac{1}{2} \mathbf{y}^\top  \biggl(\mathbb{I}_{n} + \gamma \sum_{t \in \mathcal{T}} \sum_{i\in \mathcal{N}_t} z_{i}^t {\mathbf{M}_i^t}\bigl({\mathbf{M}_i^t}\bigr )^\top\biggr)^{-1} \mathbf{y}  + \lambda\sum_{t \in \mathcal{T}} \sum_{i \in \mathcal{N}_t} a_i^t\,z_i^t,
\end{align*}
is again convex over $\mathbf{z}$ since $\biggl(\mathbb{I}_{n} + \gamma \sum_{t \in \mathcal{T}} \sum_{i\in \mathcal{N}_t} z_{i}^t {\mathbf{M}_i^t}\bigl({\mathbf{M}_i^t}\bigr )^\top\biggr)$ is positive definite. Using the result from Example 3.4 \citep{boyd2004convex} yields that convex compositions of affine functions are convex.

\subsection{Linear Relaxation Algorithm} \label{lr_alg.appx}
We consider this linear relaxation to Problem \eqref{CIP_penalized.prob}:\begin{align} \label{linear_relax.prob}
\min_{\mathbf{z}}\quad q_{LR}(\mathbf{z}) = \frac{1}{2} \mathbf{y}^\top  \biggl(\mathbb{I}_{n} + \gamma \sum_{t \in \mathcal{T}} \sum_{i\in \mathcal{N}_t} z_{i}^t {\mathbf{M}_i^t}\bigl({\mathbf{M}_i^t}\bigr )^\top\biggr)^{-1} \mathbf{y}  + \lambda\sum_{t \in \mathcal{T}} \sum_{i \in \mathcal{N}_t} a_i^t\,z_i^t  \\
\text{s.t.} \quad \sum_{j \in \mathcal{C}_i^t} z_j^t \ \leq \ |\mathcal{C}_i^t| (1-z_i^t),
\quad \forall \ i \in \mathcal{N}_t, \ t \in \mathcal{T}, \notag \\
z_i^t \in [0,1], \quad \forall 
\ i \in \mathcal{N}_t, \ t \in \mathcal{T}. \notag
\end{align}

\noindent We present a specialized outer approximation algorithm to solve Problem \eqref{linear_relax.prob} efficiently. Objective $q_{LR}(\mathbf{z})$ is convex and sub-differentiable; for all $\mathbf{z}' \in [0,1]^m$ we have that: $q_{LR}(\mathbf{z}' ) \geq q_{LR}(\mathbf{z}) + \nabla q_{LR}(\mathbf{z})^{\top}(\mathbf{z}' - \mathbf{z})$. Given a sequence of feasible solutions $\mathbf{z}^{(0)}, \mathbf{z}^{(1)}, \ldots \mathbf{z}^{(h)}$ for Problem \eqref{linear_relax.prob}, we have that:
\begin{align*}
 \max_{k=0,..,h}~q_{LR}(\mathbf{z}^{(k)}) + \nabla q_{LR}(\mathbf{z}^{(k)})^{\top}(\mathbf{z}-\mathbf{z}^{(k)})~\leq q_{LR}(\mathbf{z}).
\end{align*}.

\noindent In iteration $h$ of our outer approximation algorithm, we solve LP:
\begin{align} \label{LP.prob}
\min_{z,\nu} \quad & \nu  \\
\text{s.t.} \quad & \sum_{j \in \mathcal{C}_i^t} z_j^t \ \leq \ |\mathcal{C}_i^t| (1-z_i^t), \quad  z_i^t \in [0,1],
\quad \forall \ i \in \mathcal{N}_t, \ t \in \mathcal{T}, \notag \\
& \nu \geq q_{LR}(\mathbf{z}^{(k)}) + \nabla q_{LR}(\mathbf{z}^{(k)})^\top (\mathbf{z}-\mathbf{z}^{(k)}) 
\quad \forall \ k \in \{0, 1, \ldots, h\}, \notag
\end{align}
where $\nu$ is a lower bound for the optimal objective value $q_{LR}(\mathbf{z}^*)$, to obtain $\mathbf{z}^{(h+1)}$ and $\nu^{(h+1)}$. We terminate our algorithm when $q_{LR}(\mathbf{z}^{(h)})$ is equal to $\nu^{(h)}$; our algorithm terminates in a finite number of iterations since the feasible for $\mathbf{z}$ is compact and returns optimal solution $\mathbf{z}^*$ to Problem \eqref{linear_relax.prob}.

\subsection{Proof of Proposition \eqref{KPCG.prop}} \label{knapsack.appx}

We first prove that ILP \eqref{roundingILP.prob} is a knapsack problem with conflict graph $G$ = ($V$,$E$) where $V = \mathcal{N}_t$ and $E = \bigl\{(i,j) \ \forall \ i \in \mathcal{N}_t, \ j \in \mathcal{C}_i^t \bigr\}$. Consider the these two optimization problems.

\vspace{2mm}
\noindent \textbf{Problem} \eqref{roundingILP.prob}\begin{align*} \sum_{i \in \mathcal{N}_t}\max_{z_i^t} \quad \zeta_i^t z_i^t, \\\quad 
\text{s.t.} \quad \sum_{j \in \mathcal{C}_i^t} z_j^t \ \leq \ |\mathcal{C}_i^t| (1-z_i^t), \quad \forall 
\ i \in \mathcal{N}_t, \\ \quad
z_i^t \in \{0,1\}, \quad \forall 
\ i \in \mathcal{N}_t.
\end{align*}

\vspace{2mm}
\noindent \textbf{Knapsack Conflict Graph} (KCG) 
\begin{align*}  \sum_{i \in \mathcal{N}_t}\max_{z_i^t} \quad \zeta_i^t z_i^t, \\\quad 
\text{s.t.} \quad z_i^t  + z_j^t \leq 1, \quad \forall \ (i,j) \in \bigl\{(i,j) \ \forall \ i \in \mathcal{N}_t, \ j \in \mathcal{C}_i^t \bigr\} \\ \quad
z_i^t \in \{0,1\}, \quad \forall 
\ i \in \mathcal{N}_t. \notag
\end{align*}

\noindent Problem KCG is the ILP form of a knapsack conflict graph problem shown in \cite{pferschy2009knapsack}. Any feasible solution to Problem \eqref{roundingILP.prob} is feasible for (KCG), and vice versa, as the first constraint in both problems is equivalent. The objective functions of both problems are equivalent, so (KCG) is an equivalent reformulation of Problem \eqref{roundingILP.prob}.

\vspace{2mm} \noindent Next, consider conflict graph $G$ = ($V$,$E$) where $V = \mathcal{N}_t$ and $E = \bigl\{(i,j) \ \forall \ i \in \mathcal{N}_t, \ j \in \mathcal{C}_i^t \bigr\}$. The edges in set $E$ connect every node $i \in \mathcal{N}_t$ with all of its descendant nodes $j \in \mathcal{C}_i^t$, so graph $G$ is a  transitive closure of a tree which is chordal.

% Let node $i_0 \in \mathcal{N}_t$ represent the root node of our tree, $i_0$ is connected to all of the other nodes. Consider this collection of edges $E_0 = \{(i_0,j) \ \forall j \in \mathcal{N}_t \setminus i_0  \}$. Any arbitrary cycle 

\newpage
\section{Proofs of the Theoretical Results in Section~\ref{theory.section}} 
\label{proofs.theory.section}

Given a function $f:\mathbb{R}^p\mapsto\mathbb{R}$, we define $\bff=(f(\bx_1),...,f(\bx_n))^\top$; we define vectors such as~$\bff^*$ and~$\widehat{\bff}$ by analogy. We also let $\bY=(y_1,...,y_n)^\top$ and $\bepsilon=(\epsilon_1,...,\epsilon_n)^\top$. We use $\|\cdot\|$ to denote the Euclidean norm. As before, we use $\bx$ to denote vectors in~$\mathbb{R}^p$ and write $x_j$ for the $j$-th coordinate of~$\bx$.

Consider an arbitrary positive integer~$s$. In our analysis, we replace random functional classes $\mathcal{C}_k(s)$ with larger deterministic classes $\mathcal{F}_k(s)$ for each $k\in\{1,2,3\}$. More specifically, we  let~$\mathcal{F}_k(s)$ consist of all possible functions that might appear in~$\mathcal{C}_k(s)$. For example, we define $\mathcal{F}_1(s)$ as the class of all functions $f:\mathbb{R}^p\mapsto \mathbb{R}$ of the form $f(\bx)=\sum_{t=1}^s \beta_t \Pi_{l=1}^{d_t} I_{lt}(\bx)$, where $d_t\in\{1,...,D\}$, $\beta_t$ are real-valued coefficients, and $I_{lt}$ are indicator functions of the form $I_{lt}(\bx)=\mathbf{1}_{\{x_j\le s_{lt}\}}$ or $I_{lt}(\bx)=\mathbf{1}_{\{x_j> s_{lt}\}}$ for some $j\in\{1,...,p\}$ and real-valued split-points $s_{lt}$.

We define~$\mathcal{F}_2(s)$ as a subset of~$\mathcal{F}_1(s)$ by imposing an additional attribute constraint: $\sum_{t=1}^s d_t\mathbf{1}_{\{\beta_t\ne 0\}} \le s$. Similarly, we define~$\mathcal{F}_3(s)$ as a subset of~$\mathcal{F}_1(s)$ by requiring $\sum_{t=1}^s v_t\mathbf{1}_{\{\beta_t\ne 0\}} \le s$, where~$v_t$ is the number of distinct features~$j$ that appear in the term $\Pi_{l=1}^{d_t} I_{lt}(\bx)$. Consequently, $\mathcal{C}_k(K)\subseteq\mathcal{F}_k(K)$ for each training sample and each $k\in\{1,2,3\}$.

We will use the following result, which is derived in Appendix~\ref{lem.prf.section}.
\begin{lemma}
\label{lem.max.ineq}
Each of the following inequalities,
\begin{enumerate}
\item[(i)] $\sup_{f\in\mathcal{F}_1(2K)}\bepsilon^\top\bff \lesssim \sigma \sqrt{KD\log(enp)+\log(1/\delta_0)}\,\|f\|$
\item[(ii)] $\sup_{f\in\mathcal{F}_2(2K)}\bepsilon^\top\bff \lesssim \sigma \sqrt{K\log(enp)+\log(1/\delta_0)}\,\|f\|$
\item[(iii)] $\sup_{f\in\mathcal{F}_3(2K)}\bepsilon^\top\bff \lesssim \sigma \sqrt{K\log(enp)+\log(1/\delta_0)}\,\|f\|$,
\end{enumerate}
holds with probability at least $1-\delta_0$.
\end{lemma}

\subsection{Proof of Theorem~\ref{thm.rates}}

We first establish the result for~$\widehat{f}_1$. Consider an arbitrary~$f\in \mathcal{C}_1(K)$ and note that
\begin{equation*}
\|\bY-\widehat{\bff}_1\|^2+\tfrac1{\gamma}\|w(f)\|\le \|\bY-\bff\|^2+\tfrac1{\gamma}\|w(f)\|^2,
\end{equation*}
which implies
\begin{equation}
\label{basic.ineq}
\|\bff^*-\widehat{\bff}_1\|^2\le \|\bff^*-\bff\|^2+2\bepsilon^\top(\widehat{\bff}_1-\bff)+\tfrac1{\gamma}\|w(f)\|^2.
\end{equation}
Note that, for each training sample, $\widehat{f}_1-f\in\mathcal{F}_1(2K)$. Consequently, by Lemma~\ref{lem.max.ineq}, we have
\begin{equation}
\label{cross.prod.ineq}
  \bepsilon^\top(\widehat{\bff}_1-\bff)\lesssim \sigma\sqrt{nr_1+\log(1/\delta_0)}\|\widehat{\bff}_1-\bff\|
\end{equation}
with probability at least $1-\delta_0$. Moreover, this stochastic bound holds uniformly over all $f\in\mathcal{C}_1(K)$. We conduct the rest of the argument on the event where inequality~(\ref{cross.prod.ineq}) holds. Combining inequalities~(\ref{cross.prod.ineq}) with~(\ref{basic.ineq}), we derive, for some universal constant~$c$,
\begin{eqnarray}
\|\bff^*-\widehat{\bff}_1\|^2&\le& \|\bff^*-\bff\|^2+c\sigma\sqrt{nr_1+\log(1/\delta_0)}\|\widehat{\bff}_1-\bff\|+\tfrac1{\gamma}\|w(f)\|^2\nonumber\\
&\le& \|\bff^*-\bff\|^2+c\sigma\sqrt{nr_1+\log(1/\delta_0)}\big(\|\bff^*-\widehat{\bff}_1\|+\|\bff^*-\bff\|\big)
+\tfrac1{\gamma}\|w(f)\|^2.\label{basic.ineq2}
\end{eqnarray}
We bound the second and third terms on the right-hand side of the last inequality by applying inequality
\begin{equation}
\label{ab.ineq}
2ab\le q a^2+q^{-1}b^2,
\end{equation}
which holds for every~$q>0$ and $a,b\in\mathbb{R}$.  Setting~$q=2$, we derive bounds
\begin{equation*}
c\sigma\sqrt{nr_1+\log(1/\delta_0)}\|\bff^*-\widehat{\bff}_1\|\le 2c\sigma^2\big[nr_1+\log(1/\delta_0)\big]+\|\bff^*-\widehat{\bff}_1\|^2/2
\end{equation*}
and
\begin{equation*}
c\sigma\sqrt{nr_1+\log(1/\delta_0)}\|\bff^*-\bff\|\lesssim \sigma^2\big[nr_1+\log(1/\delta_0)\big]+\|\bff^*-\bff\|^2.
\end{equation*}
Combining these bounds with inequality~\ref{basic.ineq2}, we derive
\begin{equation*}
\|\bff^*-\widehat{\bff}_1\|^2\lesssim \|\bff^*-\bff\|^2+\sigma^2\big[nr_1+\log(1/\delta_0)\big]+\tfrac1{\gamma}\|w(f)\|^2,
\end{equation*}
which implies
\begin{equation*}
\|f^*-\widehat{f}_1\|_n^2\lesssim \|f-f^*\|_n^2+\sigma^2\big[r_1-\log(\delta_0)/n\big]+\tfrac1{n\gamma}\|w(f)\|^2.
\end{equation*}

Thus, we have established the bound in the first claim of Theorem~\ref{thm.rates}. The corresponding bounds for~$\widehat{f}_2$ and~$\widehat{f}_3$ follow by (nearly) identical arguments, which only require replacing subscript ``1'' with ``2'' or ``3'', respectively.  \qed

\begin{comment}
\subsection{Proof of Corollary~\ref{cor.rates}}
\label{cor1.prf.section}

We fix an arbitrary $k\in\{1,2,3\}$ and define $\tilde{\mathcal{C}}_k(K)=\{f\in\mathcal{C}_k(K),\;\text{s.t.}\;\|w(f)\|_\infty\le c\}$. It follows from Theorem~\ref{thm.rates} that inequality
\begin{equation}
\label{start.rt.cor1}
\|\widehat{f}_k-f^*\|_n^2\lesssim \inf_{f\in \tilde{\mathcal{C}}_k(K)} \Big\{\|f_k-f^*\|_n^2 +\tfrac1{n\gamma}\|w(f)\|^2\Big\}+ \sigma^2\big[r_k-\log(\delta_0)/n\big]
\end{equation}
holds with probability at least $1-\delta_0$. Note that $\|w(f)\|_\infty\le c$ implies $\|w(f)\|^2\le c^2K$, and hence $\tfrac1{n\gamma}\|w(f)\|^2\lesssim \sigma^2r_k$ because of the assumption $\gamma\gtrsim K(\sigma^2nr_k)^{-1}$. Consequently, taking into account the assumption on $\inf_{f\in \tilde{\mathcal{C}}_k(K)} \|f_k-f^*\|_n^2$, we can re-write inequality~(\ref{start.rt.cor1}) as follows:
\begin{equation*}
\|\widehat{f}_k-f^*\|_n^2\lesssim \sigma^2\big[r_k-\log(\delta_0)/n\big].
\end{equation*}
We note that $\log(1/\delta_k)=nr_k$. Thus, taking $\delta_0=\delta_k$, we conclude that $\|\widehat{f}_k-f^*\|_n^2\lesssim \sigma^2r_k$ with probability at least $1-\delta_k$.\qed
\end{comment}

\subsection{Proof of Corollary~\ref{cor.approx}}
\label{cor2.prf.section}

Fix any $k\in\{1,2,3\}$ and take an arbitrary function $f_k\in\mathcal{C}_k(K)$. 
%By the assumptions of Corollary~\ref{cor.approx}, there exists a (data-dependent) function $f_k\in\mathcal{C}_k(K)$, such that
%\begin{equation}
%\label{fk.rel}
%\|\bff_k-\bff^*\|^2\lesssim \sigma^2 nr_k  \qquad \text{and} \qquad \|w(f_k)\|^2\lesssim K
%\end{equation}
%with probability at least $1-\epsilon$. 
To simplify the presentation, we define $G(f)=\| \bY - \bff \|^2+\tfrac1{\gamma}\|w(f)\|$. Because $UB=G(\widetilde{f}_k)$, $LB\le G(f_k)$, and $UB=LB/(1-\tau)$, we derive
\begin{equation*}
G(\widetilde{f}_k)\le G(f_k)/(1-\tau),
\end{equation*}
and hence
\begin{equation*}
\|\bY-\widetilde{\bff}_k\|^2\le \|\bY-\bff_k\|^2/(1-\tau) +\tfrac1{\gamma}\|w(f_k)\|^2/(1-\tau).
\end{equation*}
Consequently,
\begin{equation*}
\|\bff^*-\widetilde{\bff}_k\|^2\le \|\bff^*-\bff_k\|^2 +2\bepsilon^\top(\widetilde{\bff}_k-\bff_k) +2\tau\|\bff^*-\bff_k\|^2+2\tau\|\bepsilon\|^2 +\tfrac1{\gamma}\|w(f_k)\|^2/(1-\tau),
\end{equation*}
which is a slightly modified version of inequality~(\ref{basic.ineq}) in the proof of Theorem~\ref{thm.rates}. Note that $\widetilde{f}_k\in\mathcal{C}_k(K)$ and $1/(1-\tau)\lesssim 1$. Repeating the steps in the proof of Theorem~\ref{thm.rates}, while accounting for the new terms, we conclude that inequality
\begin{equation*}
\|\widetilde{\bff}_k-\bff^*\|^2\lesssim \|\bff_k-\bff^*\|^2  + \sigma^2 \big[nr_k+\log(1/\delta_0)\big]+\tfrac1{\gamma}\|w(f_k)\|^2+\tau\|\bepsilon\|^2
\end{equation*}
holds uniformly over all $f_k\in\mathcal{C}_k(K)$ with probability at least $1-\delta_0$. Setting $\delta_0=\delta_k/2$, we derive inequality
\begin{equation}
\label{prefin.ineq}
\|\widetilde{\bff}_k-\bff^*\|^2\lesssim \|\bff_k-\bff^*\|^2+\tfrac1{\gamma}\|w(f_k)\|^2+\sigma^2 nr_k+\|\bepsilon\|^2\tau,
\end{equation}
which holds with probability at least $1-\delta_k/2$.

Standard chi-square tail bounds \citep[for example, those in Section 8.3.2 of][]{buhlmann2011statistics} imply that, with an appropriate multiplicative constant, inequality $\|\bepsilon\|^2\lesssim \sigma^2n(1+r_k)$ holds with probability at least $1-\delta_k/2$. Thus, after dividing by~$n$, we can re-write inequality~(\ref{prefin.ineq}) as
\begin{equation*}
\|\widetilde{f}_k-f^*\|^2_n\lesssim \|f_k-f^*\|_n^2 +\tfrac1{n\gamma}\|w(f_k)\|^2 + \sigma^2 r_k+\sigma^2\tau,
\end{equation*}
which holds with probability at least $1-\delta_k$. \qed

\subsection{Proof of Lemma~\ref{lem.max.ineq}}
\label{lem.prf.section}

Recall that functions in the class~$\mathcal{F}_1(2K)$ have the form $f(\bx)=\sum_{t=1}^{2K} \beta_t \Pi_{l=1}^{d_t} I_{lt}(\bx)$, where each indicator $I_{lt}$ determines a split. We define $h_{t}=\Pi_{l=1}^{d_t} I_{lt}$ and, as before, we let $\bh_t=(h_{t}(\bx_1),...,h_{t}(\bx_n))^\top$. Note that $\bff=\sum_{t=1}^{2K} \beta_t\bh_t$.

We start by fixing the collection $\{\bh_1,...,\bh_{2K}\}$, while allowing coefficients~$\beta_t$ to vary, and uniformly bounding the corresponding inner products $\bepsilon^\top\bff$.
We choose an orthonormal basis $\bPh=[{\boldsymbol \psi}_1,...,{\boldsymbol \psi}_{2K}]$, such that the corresponding linear space contains the space spanned by the vectors $\bh_1,...,\bh_{2K}$. First, observe that
\begin{equation*}
\bepsilon^\top\bff \le \|\bPh^\top\bepsilon\| \|\bff\|
\end{equation*}
for every choice of coefficients $\beta_1,...,\beta_{2K}$. Next, note that $\|\bPh^\top\bepsilon\|^2/\sigma^2$ has chi-square distribution with at most~$2K$ degrees of freedom. Using a chi-square tail bound \citep[for example, the one in Section 8.3.2 of][]{buhlmann2011statistics}, we derive that $\|\bPh^\top\bepsilon\|^2 \lesssim  \sigma^2 K(1+a)$ with probability at least $1-\exp(-4K a)$. Putting everything together, we conclude that, with probability at least $1-\exp(-4K a)$, inequality
\begin{equation}
\label{fixed.S.bnd}
\bepsilon^T\bff \lesssim   \Big[ \sigma^2 K(1+a) \Big]^{1/2} \, \|\bff\|
\end{equation}
holds uniformly for all $\beta_1,...,\beta_{2K}$.

We now extend this bound so that it holds for \textit{all} collections $\{\bh_t\}_{t=1}^{2K}$ allowed by the corresponding functional class, i.e., $\mathcal{F}_1(2K)$, $\mathcal{F}_2(2K)$, or $\mathcal{F}_3(2K)$.

\noindent\textbf{Class~$\mathcal{F}_1(2K)$ and claim (i) of Lemma~\ref{lem.max.ineq}}.  Consider all the possible values that vectors~$\bh_t$ can take. For each~$\bh_t$, the underlying indicator function~$h_t$ corresponds to at most~$D$ splits, each involving one of the~$p$ features. Because~$\bh_t$ evaluates~$h_t$ at the training data that has~$n$ observations, we only need to consider $n$ possible locations for each split point; there are also two options for the direction of each split inequality. Consequently, there are at most $(2np)^D$ values that vectors~$\bh_t$ can take, and hence the number of distinct sets $\{\bh_t\}_{t=1}^{2K}$ is bounded by $(2np)^{2KD}$. Applying the union bound over all such sets, we deduce that~(\ref{fixed.S.bnd}) holds uniformly over all~$f\in\mathcal{F}_1(2K)$ with probability at least $1-\exp(-4K a+2KD\log(2np))$. Taking $a=D\log(2np)/2+\log(1/\delta_0)/(4K)$, we conclude that inequality
\begin{equation*}
\sup_{f\in\mathcal{F}_1(2K)}\bepsilon^T\bff \lesssim   \Big[ \sigma^2 KD\log(enp)+\sigma^2\log(1/\delta_0) \Big]^{1/2} \|\bff\|
\end{equation*}
holds with probability at least $1-\delta_0$.

\noindent\textbf{Class~$\mathcal{F}_2(2K)$ and claim (ii) of Lemma~\ref{lem.max.ineq}}.  Functions in the class~$\mathcal{F}_2(2K)$ have the form $f(\bx)=\sum_{t=1}^{s} \beta_t \Pi_{l=1}^{d_t} I_{lt}(\bx)$, where each indicator $I_{lt}$ determines a split, $s\in\{1,...,2K\}$ is the number of rules, and the sum of the rule-depths is at most~$2K$. We will suppose that the sum is exactly~$2K$, as we can achieve this by repeating some of the splits.

Let $N_{2K}$ denote the number of ways we can distribute the total depth~$2K$ among the rules associated with function~$f$. We note that during this distribution process we also specify the \textit{number} of rules for~$f$, which is allowed to be between one and~$2K$. Our distribution problem is equivalent to counting the number of ways we can distribute~$2K$ balls into $L$ buckets, where $1\le L\le 2K$, and the balls are indistinguishable. Given an~$L$, we can upper-bound the count by the number of ways we can place $L-1$ separators in $2K-1$ locations. Hence,
\begin{equation*}
N_{2K}\le \sum_{k=0}^{2K-1} {2K-1 \choose k} = 2^{2K-1}.
\end{equation*}
Note that the total number of splits in the rules equals the total depth, which is~$2K$. Once we have decided on the number of rules and distributed the depths among the rules, we just need to consider all the possible ways of making~$2K$ splits. Arguing as before, we bound the number of ways by $(2np)^{2K}$.

Putting everything together, we can bound the number of distinct sets $\{\bh_t\}_{t=1}^{2K}$ by $2^{2K-1}(2np)^{2K}$. Applying the union bound over all such sets, we deduce that~(\ref{fixed.S.bnd}) holds uniformly over all~$f\in\mathcal{F}_2(2K)$ with probability at least $1-\exp(-4K a+2K\log(4enp))$. Taking $a=\log(4enp)/2+\log(1/\delta_0)/(4K)$, we conclude that inequality
\begin{equation*}
\sup_{f\in\mathcal{F}_2(2K)}\bepsilon^T\bff \lesssim   \Big[ \sigma^2 K\log(enp)+\sigma^2\log(1/\delta_0) \Big]^{1/2} \|\bff\|
\end{equation*}
holds with probability at least $1-\delta_0$.

\noindent\textbf{Class~$\mathcal{F}_3(2K)$ and claim (iii) of Lemma~\ref{lem.max.ineq}}. Relative to the case with the class~$\mathcal{F}_1(2K)$ we have the additional restriction that the sum over the rules of the numbers of distinct features per rule is at most~$2K$. The number of ways to chose the features used in the rules is bounded by $p^{2K}$. Arguing as we did for claim (ii), we note that the number of ways to distribute these chosen~$2K$ features across the rules is bounded by $2^{2K-1}$. Note that a given feature in a given rule can be used in at most two splits, otherwise we can discard some of the splits. Hence, given the allocation of the $2K$ features across the rules, we need to consider at most~$4K$ splits, each with at most $n$ possible split-point locations and two directions for the split inequality. This leads to at most $(2n)^{4K}$ possibilities.

Putting everything together, we can bound the number of distinct sets $\{\bh_t\}_{t=1}^{2K}$ by $p^{2K}2^{2K-1}(2n)^{4K}$. Applying the union bound over all such sets, we deduce that~(\ref{fixed.S.bnd}) holds uniformly over all~$f\in\mathcal{F}_2(2K)$ with probability at least $1-\exp(-4K a+4K\log(2enp))$. Taking $a=\log(2enp)+\log(1/\delta_0)/(4K)$, we conclude that inequality
\begin{equation*}
\sup_{f\in\mathcal{F}_3(2K)}\bepsilon^T\bff \lesssim   \Big[ \sigma^2 K\log(enp)+\sigma^2\log(1/\delta_0) \Big]^{1/2} \|\bff\|
\end{equation*}
holds with probability at least $1-\delta_0$. \qed

\newpage
\section{Experiments} We present additional details and results regarding the experimental evaluation of our framework.

\color{black}

\subsection{Timing Comparisons} \label{OA_vs_approx.appx} 

We compare the computation time of our proposed exact algorithm with that of the approximate algorithm. Following the experimental setup in \S\ref{optimal_timing.section}, we apply the approximate algorithm to compute regularization paths up to $K = 20$ rules, the same number extracted by the exact algorithm. Table~\ref{OA_vs_approx.table} reports the computation time of each method averaged over five runs, with standard errors shown in parentheses. As shown, the approximate algorithm is consistently  faster than the exact algorithm across all problem sizes considered.

\begin{table}[h]\centering \scalebox{0.7}{
\begin{tabular}{|c|c|c|c|}
\hline
\textbf{\begin{tabular}[c]{@{}c@{}}Ensemble Size\\ \# Trees / Depth\end{tabular}} & \textbf{\begin{tabular}[c]{@{}c@{}}Problem Size\\ Obs. / Binary Vars.\end{tabular}} & \textbf{\begin{tabular}[c]{@{}c@{}}Proposed Exact \\ Algorithm\end{tabular}} & \textbf{\begin{tabular}[c]{@{}c@{}}Proposed Approximate Algorithm\\  (regularization path up to K = 20)\end{tabular}} \\ \hline
\textbf{100 / 3}                                                                  & \textbf{1000 / 1500}                                                                & 1.69s (0.04)                                                                 & 1.38s (0.02)                                                                                                          \\ \hline
\textbf{100 / 5}                                                                  & \textbf{1000 / 5000}                                                                & 7.64s (0.14)                                                                 & 2.20s (0.03)                                                                                                          \\ \hline
\textbf{500 / 4}                                                                  & \textbf{1000 / 10000}                                                               & 14.2s (0.18)                                                                 & 7.84s (0.40)                                                                                                          \\ \hline
\textbf{100 / 3}                                                                  & \textbf{3000 / 1500}                                                                & 8.8s (0.15)                                                                  & 1.89s (0.04)                                                                                                          \\ \hline
\textbf{100 / 5}                                                                  & \textbf{3000 / 5000}                                                                & 18.3s (0.18)                                                                 & 2.65s (0.06)                                                                                                          \\ \hline
\textbf{500 / 4}                                                                  & \textbf{3000 / 10000}                                                               & 38.1s (0.42)                                                                 & 8.31s (0.15)                                                                                                          \\ \hline
\textbf{100 / 3}                                                                  & \textbf{5000 / 1500}                                                                & 164.8s (1.23)                                                                & 2.22s (0.12)                                                                                                          \\ \hline
\textbf{100 / 5}                                                                  & \textbf{5000 / 5000}                                                                & 247.49s (2.45)                                                               & 3.42s (0.24)                                                                                                          \\ \hline
\textbf{500 / 4}                                                                  & \textbf{5000 / 10000}                                                               & 10m 21s (10.41)                                                              & 11.3s (0.324)                                                                                                         \\ \hline
\textbf{100 / 3}                                                                  & \textbf{10000 / 1500}                                                               & 6m 39s (12.50)                                                               & 2.30s (0.31)                                                                                                          \\ \hline
\textbf{100 / 5}                                                                  & \textbf{10000 / 5000}                                                               & 41m 21s (6.19)                                                               & 3.63s (0.94)                                                                                                          \\ \hline
\textbf{500 / 4}                                                                  & \textbf{10000 / 10000}                                                              & 58m 30s (10.65)                                                              & 8.28s (0.51)                                                                                                          \\ \hline
\end{tabular}}
\caption{Timing comparisons between our optimal and approximate algorithm.}
\label{OA_vs_approx.table}
\end{table}

\color{black}
\newpage
\subsection{Datasets} \label{datasets.appx}

\begin{table}[h]
\centering
\begin{tabular}{lrr}
\hline
\textbf{Name} & \textbf{Observations} & \textbf{Features} \\
\hline
autoMpg & 398 & 25 \\
ESL & 488 & 4 \\
no2 & 500 & 7 \\
stock & 950 & 9 \\
socmob & 1156 & 39 \\
Moneyball & 1232 & 72 \\
us\_crime & 1994 & 1954 \\
space\_ga & 3107 & 6 \\
pollen & 3848 & 4 \\
abalone & 4177 & 10 \\
Mercedes\_Benz\_Greener\_Manufacturing & 4209 & 563 \\
mtp & 4450 & 202 \\
satellite\_image & 6435 & 36 \\
wine\_quality & 6497 & 11 \\
wind & 6574 & 14 \\
cpu\_small & 8192 & 12 \\
bank32nh & 8192 & 32 \\
kin8nm & 8192 & 8 \\
puma32H & 8192 & 32 \\
Ailerons & 13750 & 40 \\
pol & 15000 & 48 \\
elevators & 16599 & 18 \\
houses & 20640 & 8 \\
house\_16H & 22784 & 16 \\
mv & 40768 & 14 \\
\hline
\end{tabular}\caption{Datasets considered in our experimental evaluation} 
\end{table}

\color{black}
\subsection{Performance Comparisons Against Original Ensemble} \label{comparison_original.appx}

\begin{figure}[h]
    \centering
    \includegraphics[width=0.65\linewidth]{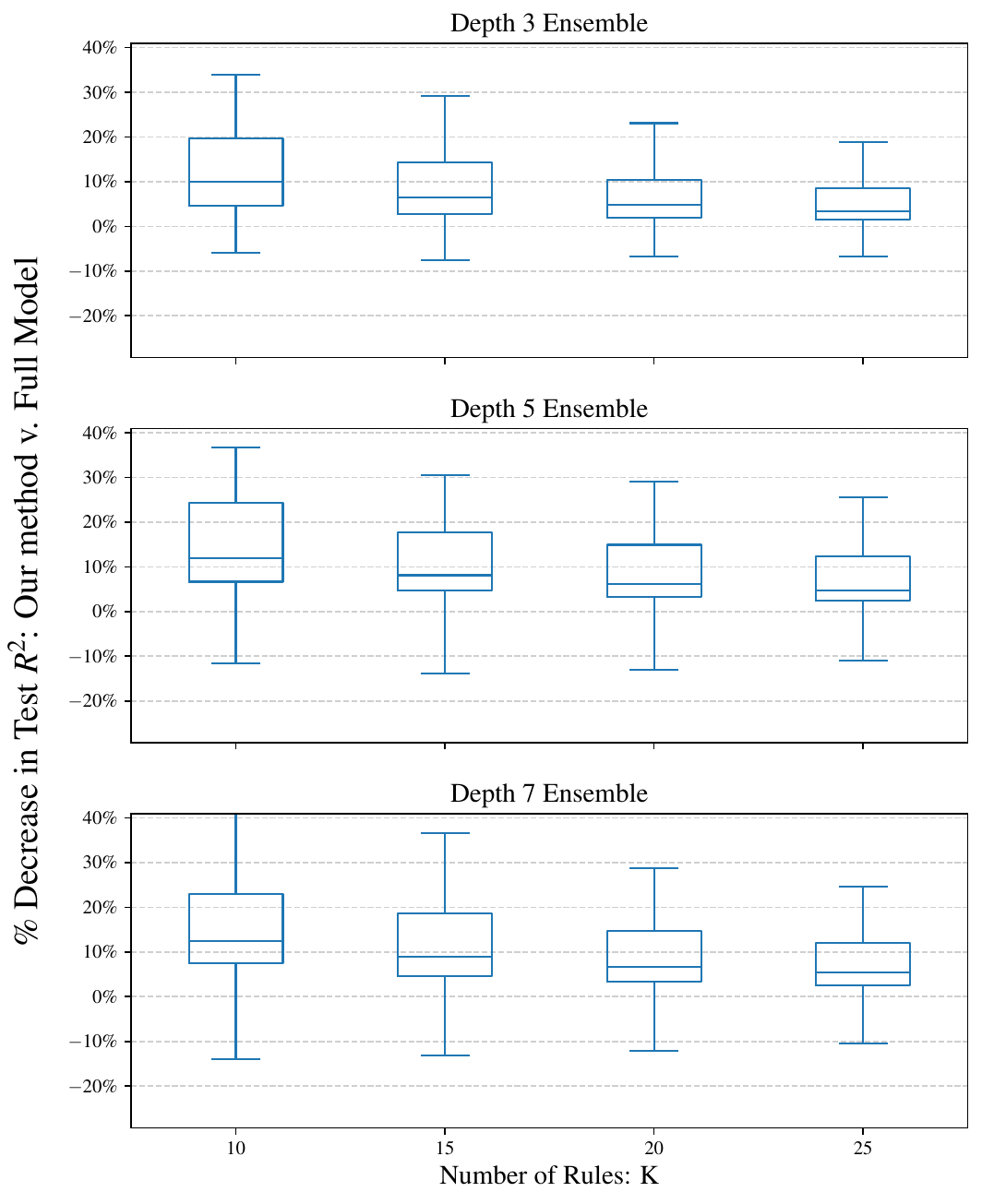}
    \caption{Predictive performance of extracted rule sets against original ensemble.}
    \label{extracted_vs_full.fig}
\end{figure}

In this section, we compare the predictive performance of the interpretable rule sets extracted by our estimator in \S\ref{interpretable_rules_experiment.section} against the original ensemble. We show these comparisons in Figure~\ref{extracted_vs_full.fig}. The vertical axes of each plot shows the percentage decrease in test $R^2$ between our extracted model and the original ensemble, given by $(\text{Original Model} \ R^2 - \text{Extracted Model} \ R^2)/(\text{Original Model} \ R^2)$. Smaller percentage decreases in $R^2$ indicate that the extracted model retains more predictive accuracy. The horizontal axes show the sizes of the extracted models and each panel in the figure corresponds to a different depth of the original ensemble.

From this figure, we observe that the median percentage decrease in test $R^2$ between the extracted model and the original hovers around $10\%$. As expected, this gap shrinks as the size of the extracted model increases. Notably, extracted models of 25 decision rules remain interpretable while performing within approximately $5\%$ of the original ensemble in terms of accuracy. These results underscore the ability of our estimator to extract compact, interpretable models that retain a high degree of predictive accuracy.

\color{black}

\subsection{Model Compression} \label{model_compress.appx}

We also evaluate how effectively our algorithm compresses large tree ensembles, an important consideration for reducing inference time and memory footprint. The models extracted in this setting contain hundreds of rules and are therefore not intended to be interpretable. Instead, the goal of this experiment is to demonstrate how our proposed method can improve memory and compute efficiency.

\vspace{2mm}
\noindent{\textbf{Experimental Procedure:}} On each of the 25 OpenML datasets from above, we conduct a 5-fold cross validation and further split each training fold into a training set and a validation set. On the training sets, we fit large boosting ensembles consisting of 500 depth 7 decision trees; such ensembles  typically contain around $10^5$ nodes. We then apply our framework to prune these ensembles by computing regularization paths and selecting the smallest pruned model that achieves a validation $R^2$ within a prespecified margin  $ \phi \in \{1\%,2.5\%,5\%,…,20 \% \}$ of the full model.  Here, $\phi$ serves as a tuning parameter in the compression procedure. We evaluate the test $R^2$ and size of the selected pruned model.

We report the following metrics: the percentage decrease in test $R^2$ between our pruned model and the full model, given by: $(\text{Full Model} \ R^2 - \text{Pruned Model} \ R^2)/(\text{Full Model} \ R^2)$, and the compression factor, given by: $( \# \text{ Nodes Full Model})/(\#\text{ Nodes Pruned Model})$. Finally, we repeat this procedure using  FIRE, RuleFit, ForestPrune, and ISLE.

\begin{figure}[h]
    \centering
    \includegraphics[width=0.7\linewidth]{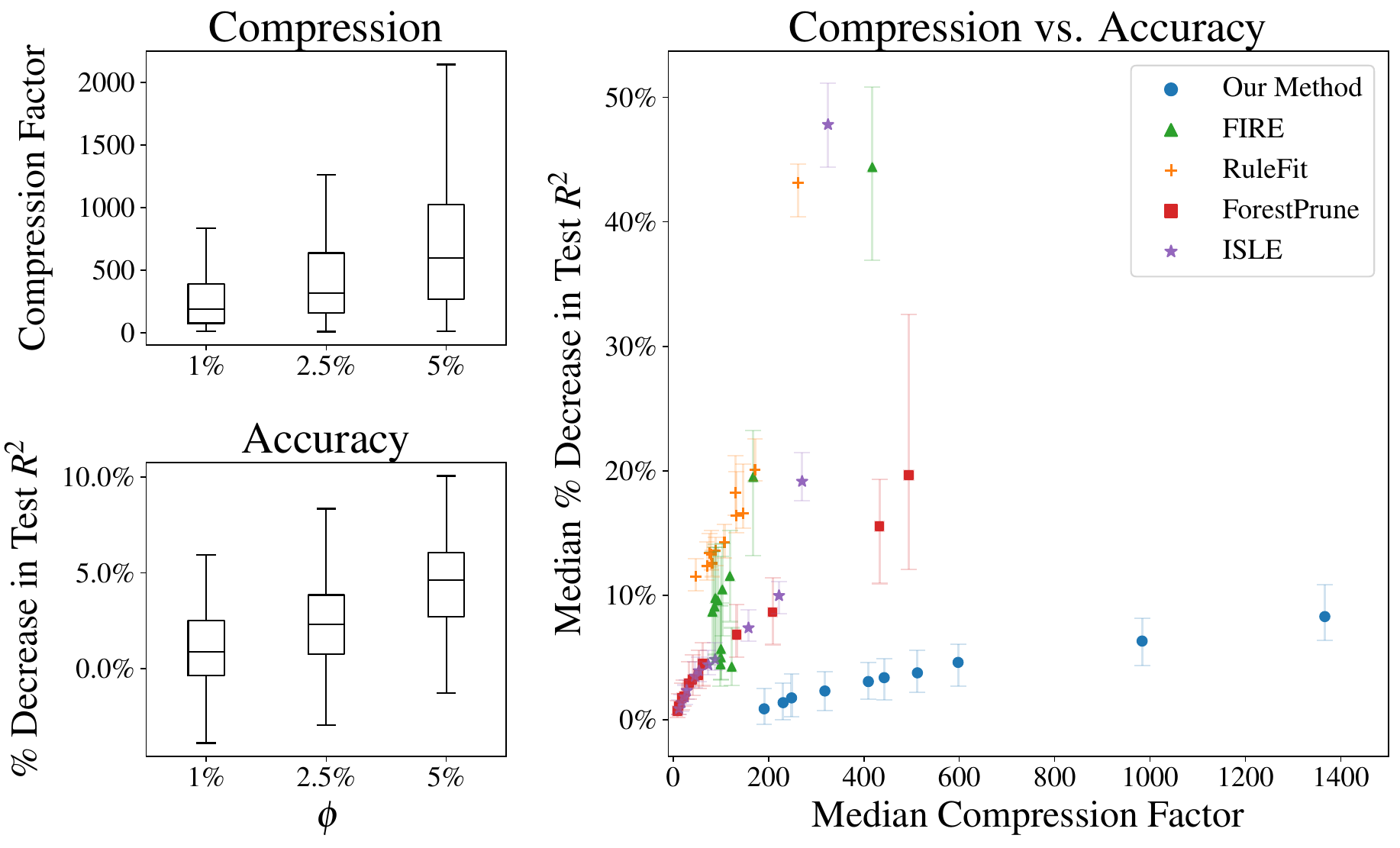}
    \caption{Results from our model compression experiment. The panels on the left show the distributions of the compression factor and the percentage decrease in test $R^2$ between our pruned model and the full model, obtained across all datasets and folds. The right panel compares the performance of our pruning method against competing algorithms.}
    \label{model_compression_result.fig}
\end{figure}

\vspace{2mm}
\noindent{\textbf{Experimental Results:}} We show the results of our experiment in Figure~\ref{model_compression_result.fig}. In the left and center panels, we display boxplots of the distributions of the compression factor (top left) and the percentage decrease in test \( R^2 \) (bottom left) between our pruned model and the full model, across all datasets and folds, for $\phi \in \{1\%, 2.5\%, 5\%\}$. For the $\phi = 1\%$ setting, our method achieves a median compression factor of $\mathbf{190}\boldsymbol{\times}$ relative to the full model, while incurring only a median decrease in test $R^2$ of $\mathbf{0.8\%}$. For the $\phi = 2.5\%$ and $\phi = 5\%$ settings, our method can prune models on median $\mathbf{300}\boldsymbol{\times}$ and $\mathbf{600}\boldsymbol{\times}$ times smaller than the full model, while incurring a median decrease in test $R^2$ of $\mathbf{2.3\%}$ and $\mathbf{4.6\%}$, respectively. These results show that our method can compress large  ensembles by factors exceeding two orders of magnitude, while incurring only nominal losses in predictive accuracy.

The right panel in Figure~\ref{model_compression_result.fig} compares our method  against the competing algorithms, across all values of $\phi$ in our experiment. The horizontal axis shows the median compression factor obtained by each method (higher is better), and the vertical axis shows the median percentage decrease in test $R^2$ between the pruned models and the full model (lower is better).
We observe that our method achieves significantly higher compression factors with lower decreases in test $R^2$ compared to all competing algorithms. Notably, we can prune models to nearly $\mathbf{1000}\boldsymbol{\times}$ smaller than the original boosting ensemble while retaining out-of-sample $R^2$ scores within around \textbf{5\%} of the full model. These  results demonstrate that our method achieves significantly better compression of large models than competing algorithms.

\newpage
\color{black}
\section{Case Study}

\subsection{Additional Details} 
\label{additional_details_case.appx}
We present additional details regarding our real-world case study.

\begin{figure}[h]
    \centering
\includegraphics[width=0.5\linewidth]{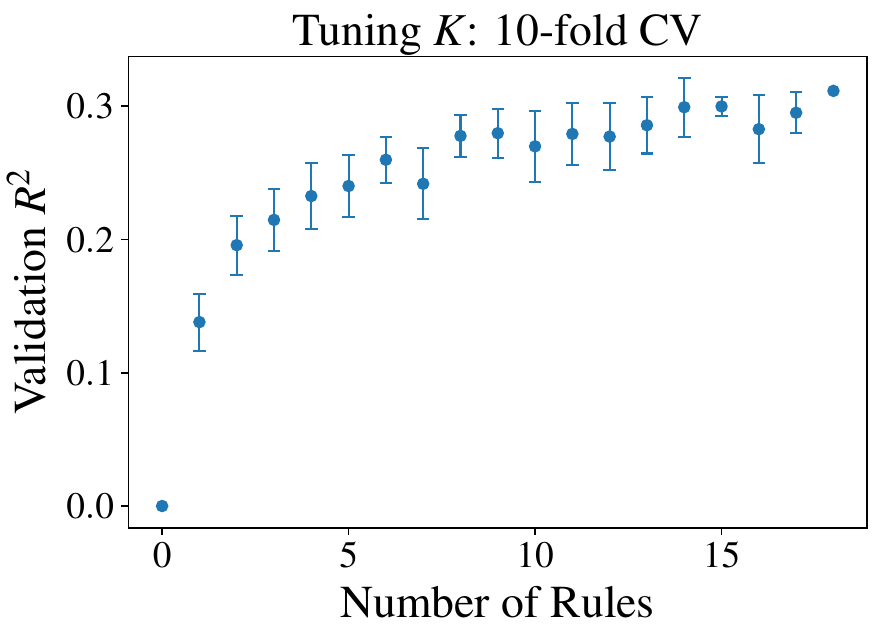}
    \caption{Mean validation $R^2$ as a function of the number of rules. The error bars show standard error. Model performance plateaus after approximately ten rules, which we select for the final model.}
    \label{case_study_val.fig}
\end{figure}

\vspace{2mm}
\noindent{\textbf{Selecting the Number of Rules:}} To determine the number of rules to extract in our case study, we perform a 10-fold cross-validation on the training data and use our estimator to compute regularization paths. The results are shown in Figure~\ref{case_study_val.fig}. From this figure, we observe that the validation accuracy plateaus after approximately ten rules; therefore, we select ten rules for our final model.

\vspace{2mm}
\noindent{\textbf{Rule Importance:}} Recall from \S\ref{scorecard.section} that we discussed several metrics for evaluating rule importance. Below, we illustrate one such approach by assessing rule importance based on the magnitude of each rule’s contribution. Table~\ref{scorecard_form1.table} presents the scorecard representation of our extracted model, where the rules are sorted in decreasing order of contribution magnitude. This table contains the same rules as Table \ref{scorecard_form.table}, but in a different order since rule importance is computed differently.

\begin{table}[ht]
\centering \scalebox{0.76}{
\renewcommand{\arraystretch}{1.25}  % row spacing
\setlength{\tabcolsep}{8pt}        % column spacing
\begin{tabular}{c >{\raggedright\arraybackslash}p{10.8cm} r}
\toprule
 & \textbf{Rule Conditions} & \textbf{Contribution} \\
\midrule
$\square$ &  \texttt{mechvent\_24h} = True AND \texttt{dx\_unique\_icd9\_roots} $>$ 16.50 AND \texttt{vent\_mode\_ps\_24h} = False & \textcolor{green!40!black}{+3.05 days} \\
\addlinespace[0.7em]
$\square$ &  \texttt{mechvent\_24h} = True AND \texttt{dx\_unique\_icd9\_roots} $>$ 8.50 AND \texttt{vt\_ml\_count\_24h} $>$ 9.50 & \textcolor{green!40!black}{+2.93 days} \\
\addlinespace[0.7em]
$\square$ &  \texttt{mechvent\_24h} = True AND \texttt{first\_careunit\_csru} = False AND \texttt{niv\_24h} = False & \textcolor{green!40!black}{+2.35 days} \\
\addlinespace[0.7em]
$\square$ &  \texttt{plateau\_max\_24h} $>$ 29.05 AND \texttt{hco3\_min\_24h} $>$ 11.50 & \textcolor{green!40!black}{+2.11 days} \\
\addlinespace[0.7em]
$\square$ &  \texttt{dx\_unique\_icd9\_roots} $>$ 21.50 & \textcolor{green!40!black}{+1.86 days} \\
\addlinespace[0.7em]
$\square$ &  \texttt{plateau\_max\_24h} $\leq$ 23.35 & \textcolor{red!70!black}{-1.82 days} \\
\addlinespace[0.7em]
$\square$ &  \texttt{cci\_mets} = False AND \texttt{rr\_mean\_24h} $>$ 20.85 AND \texttt{mechvent\_24h} = True & \textcolor{green!40!black}{+1.65 days} \\
\addlinespace[0.7em]
$\square$ &  \texttt{cci\_cvd} = True & \textcolor{green!40!black}{+1.65 days} \\
\addlinespace[0.7em]
$\square$ &  \texttt{vt\_ml\_std\_24h} $>$ 170.29 & \textcolor{red!70!black}{-0.97 days} \\
\addlinespace[0.7em]
$\square$ &  \texttt{dx\_unique\_icd9\_roots} $\leq$ 7.50 & \textcolor{red!70!black}{-0.85 days} \\
\bottomrule
\end{tabular}}
\caption{Scorecard form of extracted rule set, ordered by magnitude of contribution.}
\label{scorecard_form1.table}
\end{table}

\vspace{2mm}
\noindent{\textbf{Definition of Features:}} We list the features used and their definitions below.

\begin{itemize}
  \item \texttt{mechvent\_24h}: Indicator for whether the patient received invasive mechanical ventilation within the first 24 hours of ICU admission.
  \item \texttt{vent\_mode\_ps\_24h}: Indicator for whether pressure support ventilation mode was used within the first 24 hours.
  \item \texttt{vt\_ml\_count\_24h}: Number of distinct tidal volume measurements recorded during the first 24 hours.
  \item \texttt{vt\_ml\_std\_24h}: Standard deviation of tidal volume (mL) measurements within the first 24 hours, reflecting variability in ventilator settings.
  \item \texttt{plateau\_max\_24h}: Maximum plateau airway pressure (cmH\textsubscript{2}O) observed in the first 24 hours, a measure of lung compliance and mechanical stress.
  \item \texttt{hco3\_min\_24h}: Minimum serum bicarbonate (HCO\textsubscript{3}, mmol/L) level measured in the first 24 hours, an indicator of metabolic acidosis.
  \item \texttt{rr\_mean\_24h}: Mean respiratory rate (breaths per minute) recorded during the first 24 hours.
  \item \texttt{first\_careunit\_csru}: Indicator for whether the first ICU unit of admission was the Cardiac Surgery Recovery Unit (CSRU).
  \item \texttt{niv\_24h}: Indicator for whether the patient received non-invasive ventilation within the first 24 hours.
  \item \texttt{dx\_unique\_icd9\_roots}: Number of unique ICD-9 diagnosis code roots assigned to the patient, representing comorbidity burden and diagnostic complexity.
  \item \texttt{cci\_mets}: Charlson Comorbidity Index (CCI) indicator for metastatic solid tumor.
  \item \texttt{cci\_cvd}: Charlson Comorbidity Index (CCI) indicator for cerebrovascular disease.
\end{itemize}

\subsection{Connections with  Real-World Scoring Models} \label{existing_interpretable.appx}

When presented in scorecard form, the rule sets extracted by our estimator closely resemble interpretable scoring models already deployed in high-stakes domains such as criminal justice and healthcare. For example, the Public Safety Assessment (PSA), developed by Arnold Ventures, has been adopted across many jurisdictions in the United States to support judicial decision-making on pretrial outcomes. The model consists of a set of clearly defined risk factors whose contributions are summed to produce scores for failure to appear and new criminal activity,  which are then used by judges to inform decisions on pretrial release, supervision, and detention.

\begin{figure}[h]
    \centering
    \includegraphics[width=\linewidth]{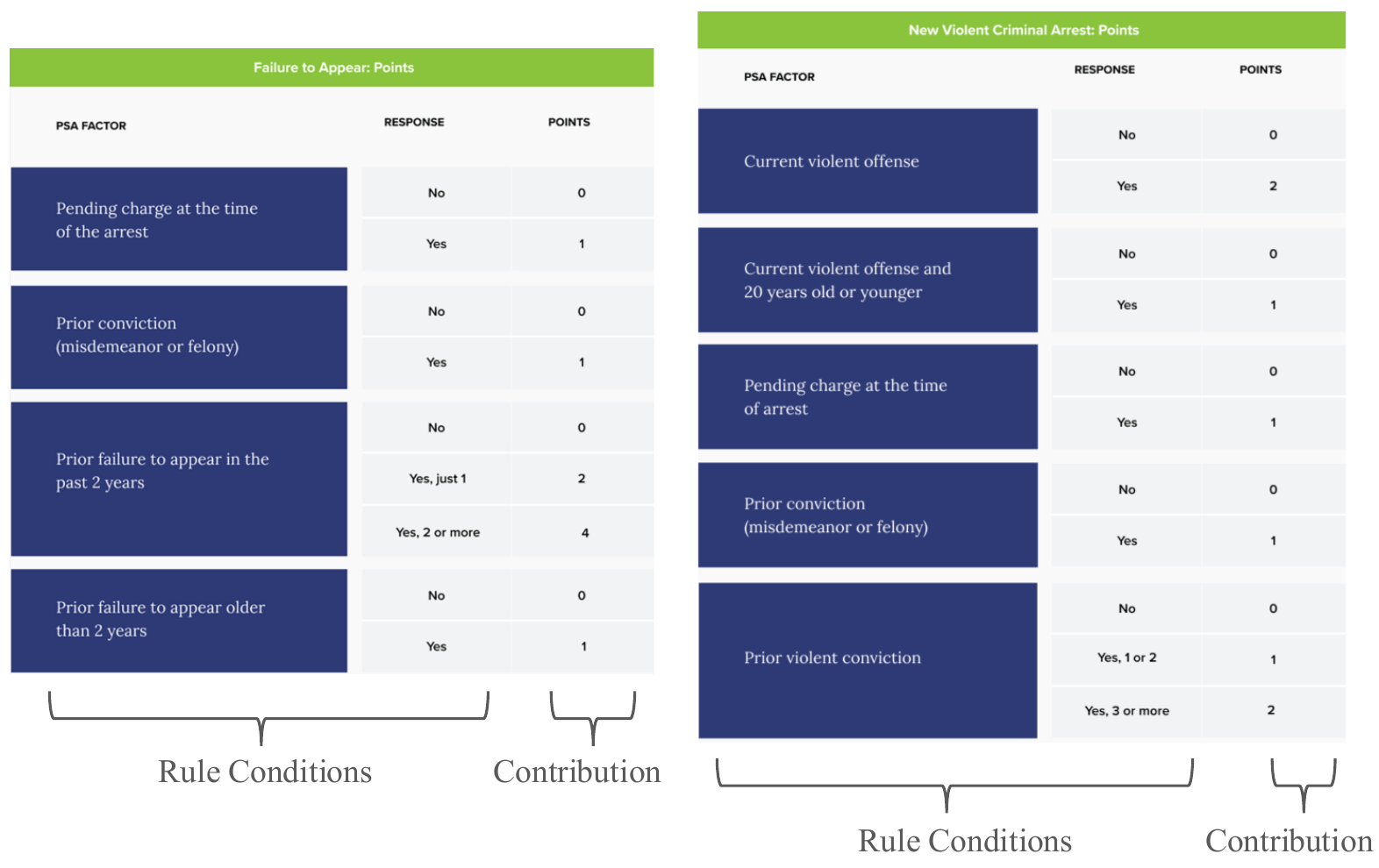}
    \caption{In the Public Safety Assessment model, the final Failure to Appear (FTA) and New Violent Criminal Activity (NVCA) scores are obtained by summing the contributions from each factor and normalizing. The table for the model is provided by \cite{APPR2020HowItWorks}.}
    \label{psa_scorecard.fig}
\end{figure}

In Figure \ref{psa_scorecard.fig}, we show the PSA model for Failure to Appear (FTA) and New Violent Criminal Activity (NVCA) scores, provided by \cite{APPR2020HowItWorks}. The model works as follows: for a given individual, each PSA risk factor is evaluated and, if applicable, contributes a specified number of points. These points are then summed across all applicable factors and normalized to produce the final risk score. This procedure closely parallels how our extracted rule sets are used. For example, in the scorecard in Table \ref{scorecard_form1.table}, each set of rule conditions is evaluated and, if satisfied, contributes a value to the prediction. The final prediction for an observation is obtained by summing the contributions across all satisfied rules.

\begin{figure}[p]
    \centering
    \includegraphics[width=\linewidth]{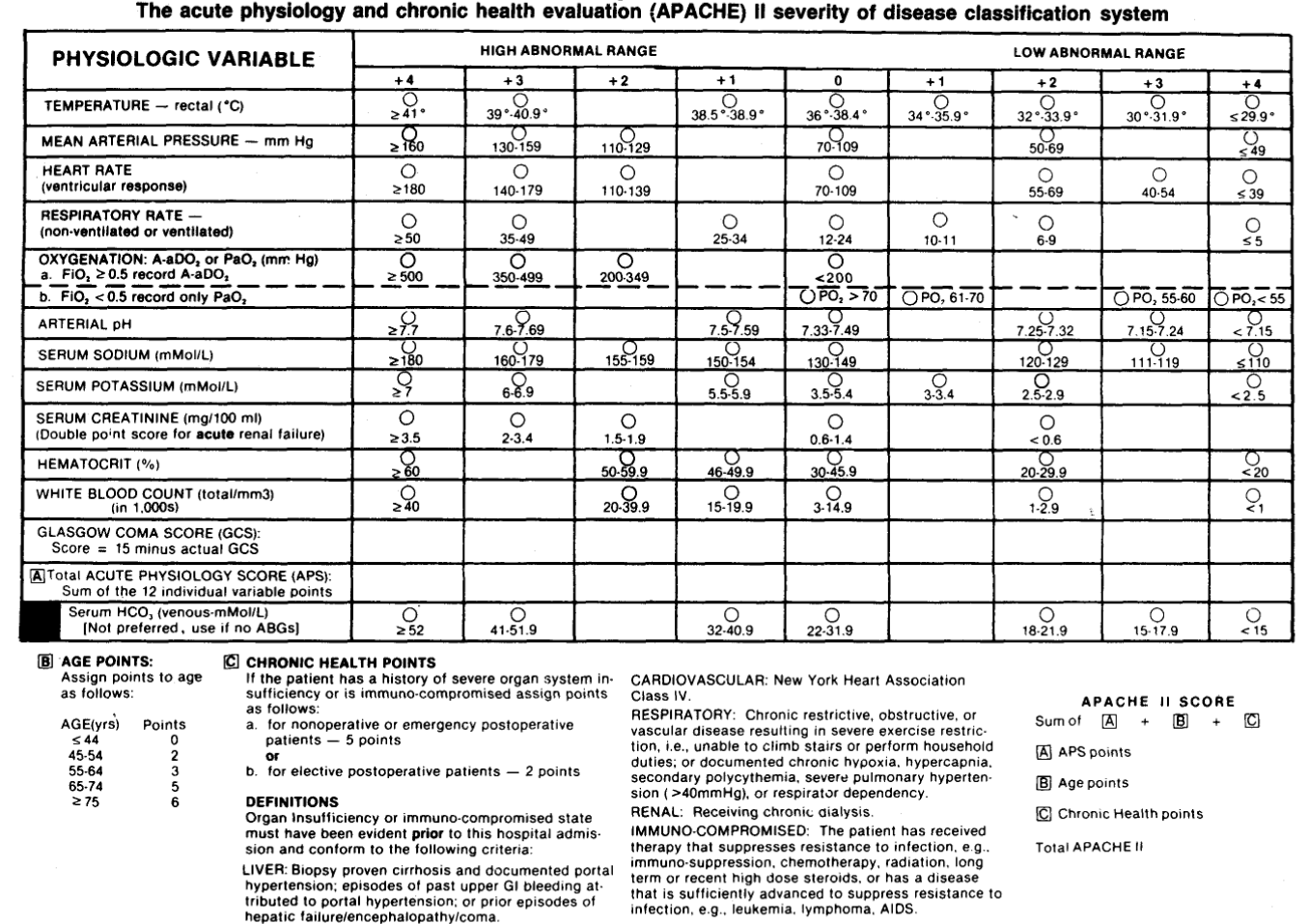}
    \caption{APACHE II scoring system, taken from Figure 1 in \cite{Knaus1985APACHEII}.}
    \label{apache_scoring.fig}
\end{figure}

Similarly, in Figure \ref{apache_scoring.fig} we show the APACHE II scoring system, taken from \cite{Knaus1985APACHEII}, which is used to assess disease severity in critically ill patients. The system works as follows: points are assigned according to how much physiologic variables deviate from the normal range, with additional points for age and chronic health problems. The final score is obtained by summing points across all categories. This model is similar to the rule sets extracted by our estimator (Table \ref{scorecard_form.table} and Table \ref{scorecard_form1.table}): each physiologic variable and range bracket corresponds to a rule condition, e.g., if Heart Rate between $140$--$179$, and the points assigned for that range represent the contribution.

As these examples show, the rule sets extracted by our estimator, when presented in scorecard form, mirror established scoring algorithms already used in practice. This similarity highlights that our method produces models interpretable in forms already familiar to practitioners across disciplines.

\subsection{Discussion on Extracted Rules} \label{rule_discuss.appx}
We discuss the remaining rules in our model and highlight some interesting real-world insights. The ordering of the rules below correspond to the scorecard shown in Table \ref{scorecard_form.table}.

\begin{itemize}
    \item \textbf{Rule 5:} If a patient does not have metastatic disease and are ventilated with a higher respiratory rate, then the predicted length of stay increases by 1.65 days.
\end{itemize}

\begin{itemize}
    \item \textbf{Rule 6:} If a patient has cardiac comorbidities, then the predicted length of stay increases by 1.65 days.
\end{itemize}

\begin{itemize}
    \item \textbf{Rule 7:} If a patient has a very high diagnostics complexity (measured using ICD-9 root codes at admission), then the predicted length of stay increases by 1.86 days.
\end{itemize}

\noindent Rules 6 and 7 highlight that diagnostic complexity and comorbidities play an important role in predicting overall length of stay, and both of these covariates are available at admission.

\begin{itemize}
    \item \textbf{Rule 8:} If a patient's maximum plateau pressure is above  29.05 cmH$_2$O and the patient does not have severe metabolic acidosis, then the predicted length of stay increases by 2.11 days. 
\end{itemize}
This rule indicates that patients with aggressive ventilator settings during the first 24 hours (as reflected by high plateau pressures) but who avoid very low bicarbonate levels are predicted to have longer ICU stays. In other words, they are sick enough to require prolonged ventilation but not so acidotic that they deteriorate rapidly.

\begin{itemize}
    \item \textbf{Rule 9:} If a patient has a lower diagnostics complexity (measured using ICD-9 root codes at admission), then the predicted length of stay decreases by 0.85 days.
\end{itemize}

\begin{itemize}
    \item \textbf{Rule 10:} If the standard deviation of a patient's tidal volume measurements is high, the the predicted length of stay decreases by by 0.97 days.
\end{itemize}
\noindent Greater variability in tidal volume measurements corresponds to less rigid ventilator control, which suggests lower severity. This rule complements Rule 1, which shows that more frequent ventilator adjustments are associated with longer predicted lengths of stay.

\section{Additional Discussions}
We discuss below some additional considerations of our proposed methodology.

\subsection{Nested Regularization Paths}

In \S\ref{regularization_paths.section}, we discuss computing regularization paths using our estimator by finding solutions to Problem \eqref{tree_based_problem1.prob} across a range of $K$ values. These solutions correspond to extracted rule sets of varying sizes. It may be desired to compute \emph{nested} regularization paths, where for any two model sizes $K'$ and $K''$ with $K' < K''$, all rules extracted in the smaller model are contained in the larger model. This nesting structure may improve interpretability, since practitioners can trace how the rule set grows with $K$ without rules being swapped in or out. This setup is closely related to forward stepwise selection in regression, where covariates are added sequentially to form a nested sequence of models \cite{hastie2009elements}.

We can readily extend our framework to compute nested regularization paths. Consider an increasing sequence of values of $K$ for Problem \eqref{tree_based_problem1.prob}, or, analogously, a decreasing sequence of values of $\lambda$ for Problem \eqref{tree_based_penalized.prob}. As we compute solutions along these sequences, we impose the constraint that each solution must contain all rules selected in the preceding model, by enforcing $z_i^t = 1$ for all previously selected rules.

\begin{figure}[h]
    \centering
    \includegraphics[width=\linewidth]{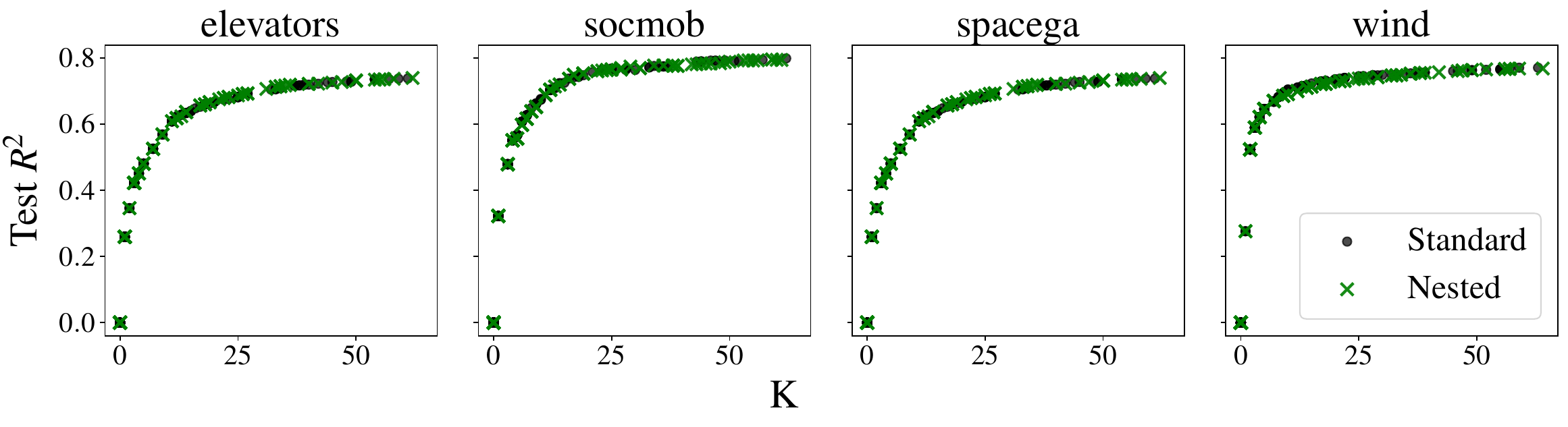}
    \caption{Nested versus standard regularization paths. Our approximate algorithm tends to produce paths with largely nested structures.}
    \label{nested_paths.fig}
\end{figure}
In Figure \ref{nested_paths.fig}, we compare nested regularization paths against standard regularization paths computed using our approximate algorithm (\S\ref{approximate_algorithm.section}) on four datasets from OpenML. The rule sets are extracted from an initial ensemble of 100 depth 5 decision trees. From this figure, we observe that the nested regularization paths perform similarly to the standard ones. Our approximate algorithm often produces regularization paths that are largely nested, owing to the warm-start procedures described in \S\ref{approximate_algorithm.section}. Nevertheless, the straightforward extension introduced here is required to guarantee exact nesting.

\subsection{Effects of Node Attribute Choices} \label{attribute_weights.appx}

\begin{figure}[p]
\centering
\begin{subfigure}{\textwidth}
    \includegraphics[width=\textwidth]{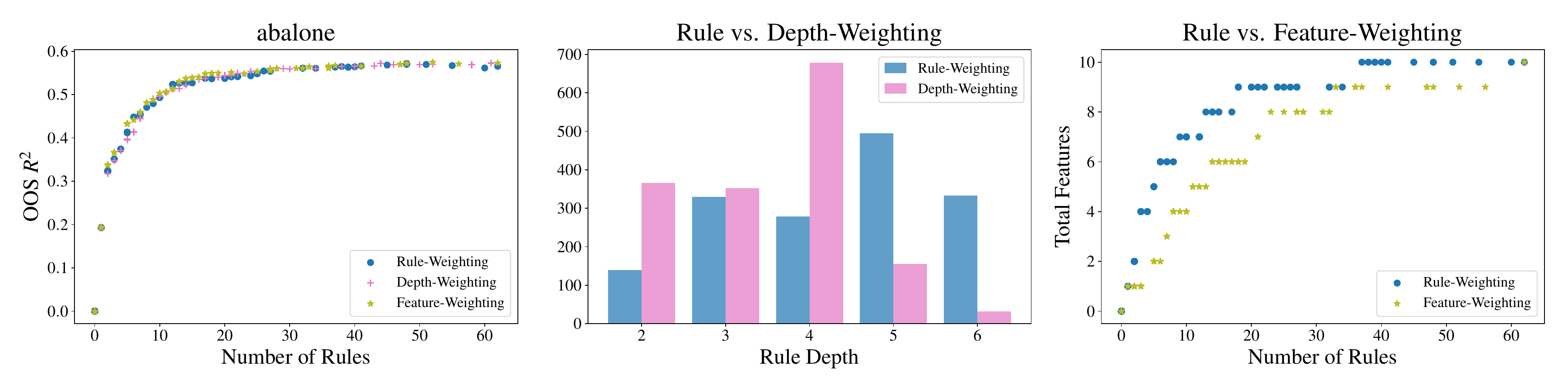}
\end{subfigure}
\begin{subfigure}{\textwidth}
    \includegraphics[width=\textwidth]{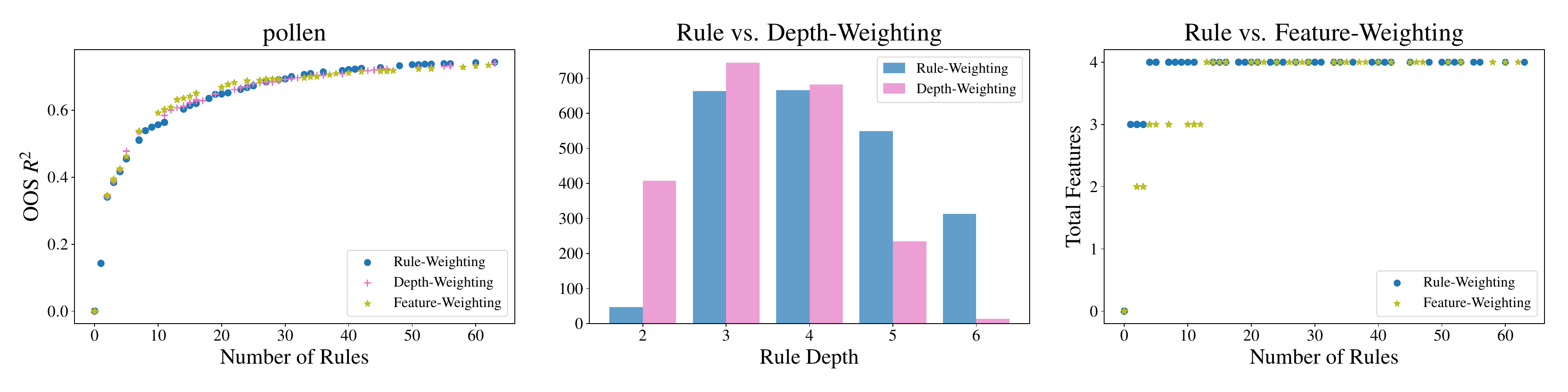}
\end{subfigure}
\begin{subfigure}{\textwidth}
    \includegraphics[width=\textwidth]{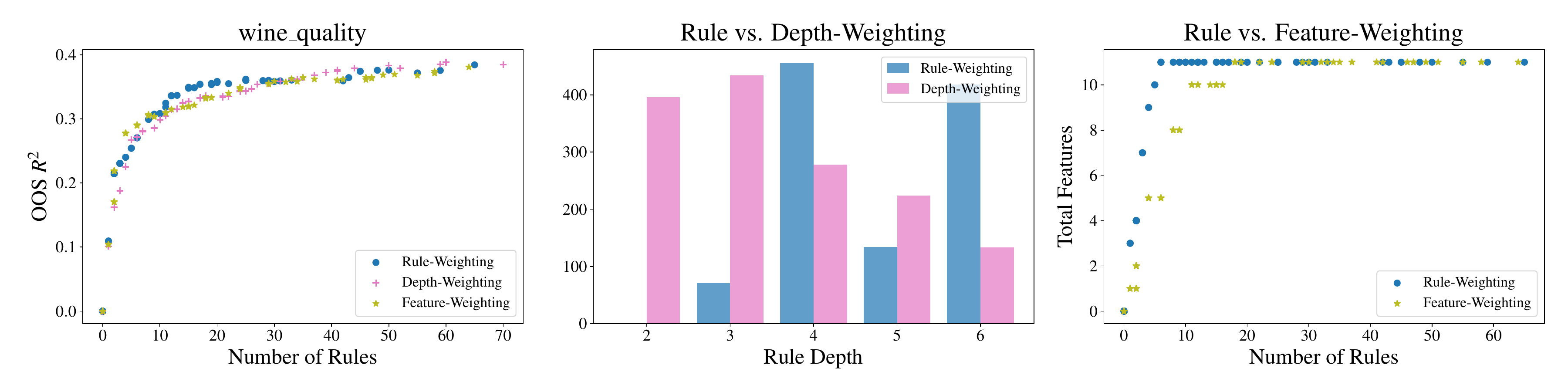}
\end{subfigure}
\begin{subfigure}{\textwidth}
    \includegraphics[width=\textwidth]{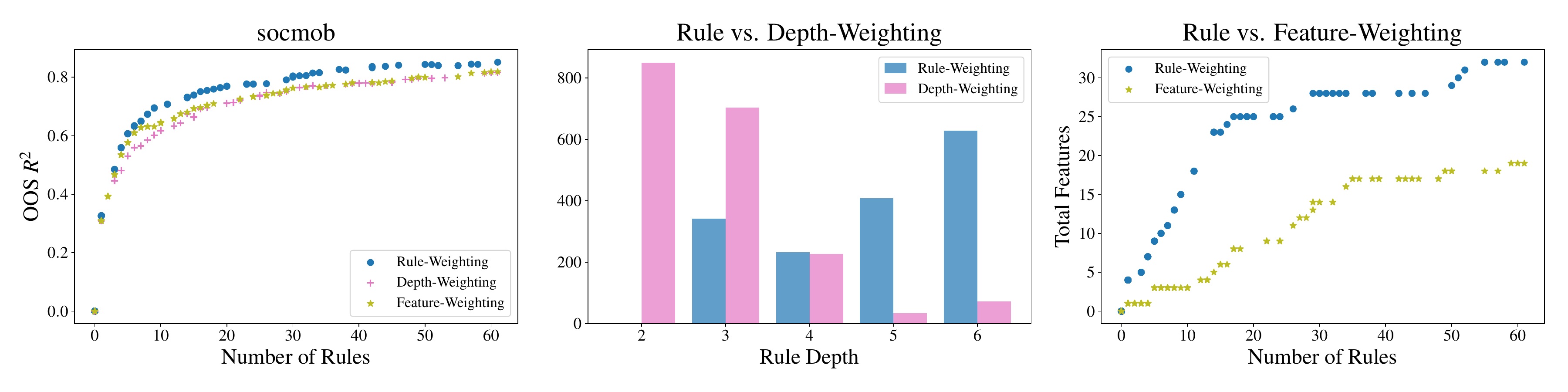}
\end{subfigure}
\begin{subfigure}{\textwidth}
    \includegraphics[width=\textwidth]{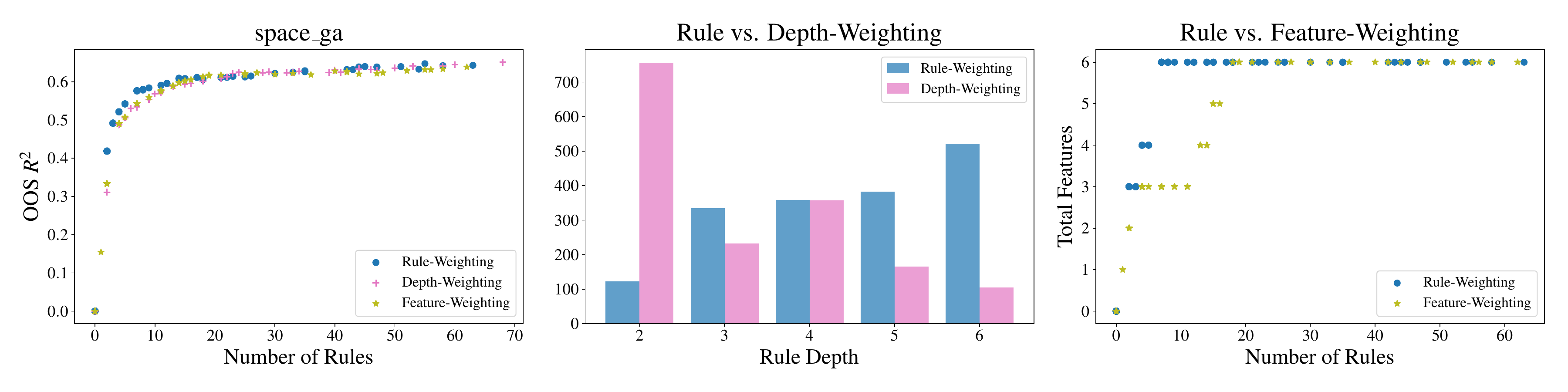}
\end{subfigure}
\caption{Comparison of node attribute choices across several datasets from OpenML.}
\label{attribute_comparisons.fig}
\end{figure}

In this section, we investigate further how different node attribute choices affect the performance and interpretability of our estimator. Following the setup described in \S\ref{node_attributes.section}, we compare rule-weighting, depth-weighting, and feature-weighting when computing regularization paths. Figure~\ref{attribute_comparisons.fig} presents results across several datasets from OpenML \citep{bischl2017openml}. For each dataset, we start with a boosting ensemble of 500 depth 7 decision trees and compute regularization paths under the different weighting schemes.

The leftmost column in Figure~\ref{attribute_comparisons.fig} shows the regularization paths computed under each weighting scheme: the horizontal axes show the number of rules and the vertical axes show out-of-sample $R^2$. The middle column of plots compares the distribution of interaction depths of the extracted rule sets, between rule-weighting and depth-weighting, across the regularization path. The rightmost column of plots compares the number of features used under rule-weighting versus feature-weighting across different model sizes.

From Figure~\ref{attribute_comparisons.fig}, we observe that on the \texttt{socmob} dataset, rule-weighting achieves the best out-of-sample performance across the regularization path. On the \texttt{pollen} dataset, feature-weighting and depth-weighting outperform rule-weighting, while on the \texttt{wine\_quality} dataset, feature-weighting performs slightly better for smaller model sizes. Overall, the differences in predictive performance across the weighting schemes are fairly small, and we examine in greater detail below how different weighting schemes can impact predictive performance. We note, however, that the extracted models differ substantially in structure. Depth-weighting encourages the extraction of shallower rules, as shown in the middle column of Figure~\ref{attribute_comparisons.fig}, while feature-weighting encourages the extraction of rule sets that use fewer overall features. These structural differences can enhance the interpretability of the extracted models.

\subsubsection{Node Attribute Choice and Predictive Performance}

Our \S\ref{theory.section} discussion of the prediction error rates established in Theorem~\ref{thm.rates} notes that depth-weighting and feature-weighting may outperform rule-weighting when the trees in the ensemble are deep and the true regression function can be well-approximated within the allocated regression budget. In this section, we empirically explore situations where this occurs.

Consider the case where we generate synthetic data of the form: $y = f(\mathbf{x}) + \epsilon$, where \(\epsilon \sim \mathcal{N}(0, \sigma^2)\),
\(\mathbf{x} = \big(x^{(1)}, x^{(2)}, \ldots, x^{(20)}\big)\),
and \(x^{(j)}\) denotes the \(j\text{-th}\) feature of observation \(\mathbf{x}\). Let the true regression function be defined by:
\[
\begin{aligned}
f(\mathbf{x}) &= 
\begin{cases}
-2.0, & \text{if } x^{(1)} < 0 \text{ and } x^{(2)} < 0.5, \\[4pt]
-0.8, & \text{if } x^{(1)} < 0 \text{ and } x^{(2)} \ge 0.5, \\[4pt]
\phantom{-}1.0, & \text{if } x^{(1)} \ge 0 \text{ and } x^{(3)} < -0.5, \\[4pt]
\phantom{-}2.3, & \text{if } x^{(1)} \ge 0 \text{ and } x^{(3)} \ge -0.5,
\end{cases} 
\end{aligned}
\]
a decision tree of depth 2. Using this model, we generate 2000 datapoints by sampling  \(\mathbf{x} \sim \mathcal{N}(0, \mathbbm{I}_{20})\) and setting $\sigma^2 = 4$ for the error term. We split the data  into a training set (80\%) and a test set (20\%), fit a deep boosting tree ensemble of 250 depth 10 decision trees, and apply our estimator to compute regularization paths under the three weighting schemes discussed above. 

\begin{figure}[h]
    \centering
    \includegraphics[width=\linewidth]{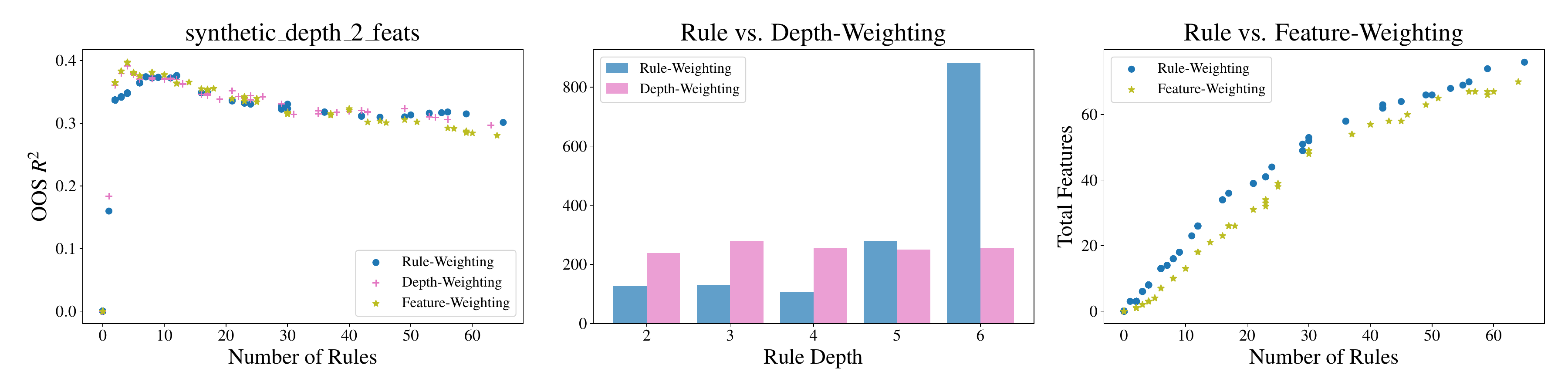}
    \caption{Synthetic data experiment: the underlying true regression function is a depth 2 tree.}
    \label{depth_2_synthetic.fig}
\end{figure}

We show the results of this study in Figure~\ref{depth_2_synthetic.fig}. From the leftmost plot, we observe that both feature-weighting and depth-weighting significantly outperform rule-weighting at extracting models with 4 rules; the underlying true regression function can be well-approximated with 4 rules. These results empirically illustrate our theoretical findings in Theorem~\ref{thm.rates}.

\begin{figure}[h]
    \centering
    \includegraphics[width=\linewidth]{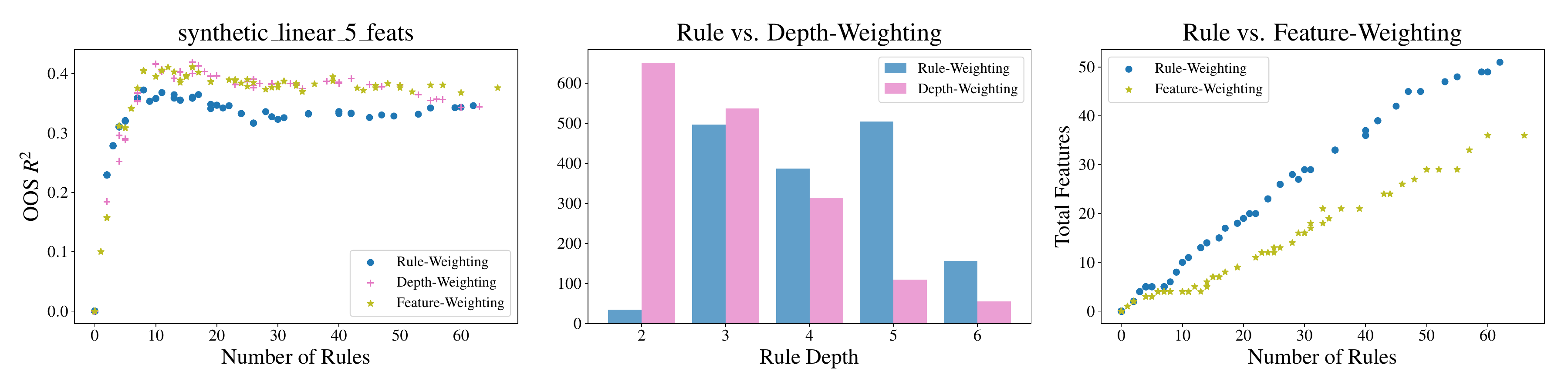}
    \caption{Synthetic data experiment: the underlying true regression function is linear.}
    \label{linear_synthetic.fig}
\end{figure}

Next, we consider the same setup but with the true regression function defined by:
\[f(\mathbf{x}) =  2.5 x^{(1)} - 2.0 x^{(2)} + 1.5 x^{(3)} - 1.0 x^{(4)} + 0.7 x^{(5)},\]
a linear function of 5 features. We compute regularization paths under the three weighting schemes and show the results of this study in Figure~\ref{linear_synthetic.fig}.

From the rightmost plot in this figure, we observe that both depth-weighting and feature-weighting outperform rule-weighting across the regularization path. This is because the underlying true regression function does not contain interaction terms and can be well-approximated using shallower decision rules.

These empirical studies, together with our theoretical results in \S\ref{theory.section}, show that applying depth-weighting or feature-weighting to our estimator can improve predictive performance over rule-weighting, depending on the structure of the underlying true regression function. In practice, we recommend using depth-weighting or feature-weighting when there is prior belief that the underlying relationships in the data can be well captured by shallower rules, or alternatively, evaluating the weighting schemes on a validation dataset and selecting the one that performs best.

\subsection{Tuning $\gamma$ Parameter} \label{tuning_gamma.appx}

In this section, we present a procedure for selecting appropriate values of the ridge penalty parameter $\gamma$. As discussed in \S\ref{formulation.section}, the ridge penalty influences both the computational efficiency of our optimization algorithms and the predictive performance of our estimator. Larger ridge penalties (corresponding to smaller values of $\gamma$) lead to faster computation and stronger regularization, which can be beneficial in low signal-to-noise ratio settings. We also note that the computation time of our approximate algorithm is less sensitive to $\gamma$ than that of our optimal algorithm. The value of $\gamma$ affects the number of outer-approximation iterations required for convergence, and our approximate algorithm solves subproblems that are smaller by a factor of one over the number of trees in the ensemble.

Our tuning procedure proceeds as follows: on validation data, vary $\gamma$ and use our approximate algorithm to compute regularization paths across multiple values of $K$ of interest. Examine these paths and select the smallest value of $\gamma$ that corresponds to acceptable validation performance. Then, use this value of $\gamma$ for the optimal algorithm if desired.

\begin{figure}
    \centering
\includegraphics[width=\linewidth]{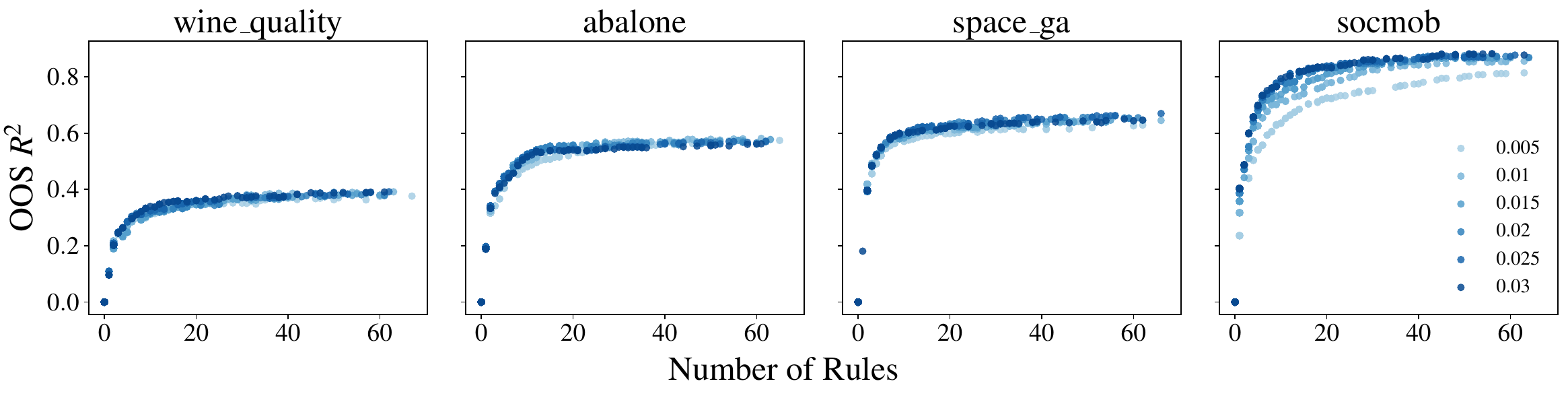}
    \caption{Regularization paths across varying values of $\gamma$.}
    \label{tuning_gamma.fig}
\end{figure}

In Figure~\ref{tuning_gamma.fig}, we visualize this procedure on four datasets from OpenML \citep{bischl2017openml}. The horizontal axes show the number of rules in the extracted model, the vertical axes show the out-of-sample $R^2$, and the color shading of the points indicates the value of $\gamma$ used to compute the regularization path. From these plots, we observe that on datasets with low to moderate signal-to-noise ratios, such as the \texttt{wine\_quality}, \texttt{abalone}, and \texttt{space\_ga} datasets, the predictive performance of the regularization path is largely insensitive to the choice of $\gamma$. In fact, for the \texttt{abalone} dataset, smaller values of $\gamma$ (corresponding to a larger regularization penalty) slightly improve out-of-sample performance, likely due to  regularization. On the \texttt{socmob} dataset, which has a higher signal-to-noise ratio, we observe that too much regularization significantly reduces the predictive performance of our estimator. In such cases, it is important to tune $\gamma$ using our proposed procedure to balance predictive accuracy and computation time. As a default, we recommend setting $\gamma = 0.02$ as an initial choice, as this value performs well empirically across many of the datasets considered.

\color{black}

\bibliographystyle{JASA}
\bibliography{ref}

\begin{thebibliography}{43}
\newcommand{\enquote}[1]{``#1''}
\expandafter\ifx\csname natexlab\endcsname\relax\def\natexlab#1{#1}\fi

\bibitem[{{Advancing Pretrial Policy and Research}(2020)}]{APPR2020HowItWorks}
{Advancing Pretrial Policy and Research} (2020), \enquote{Public Safety Assessment: How It Works,} \url{https://cdn.filestackcontent.com/5gCeQzRTuWKKCf5WL7mg}, developed by Arnold Ventures. Accessed September 23, 2025.

\bibitem[{Agarwal et~al.(2025)Agarwal, Kenney, Tan, Tang, and Yu}]{Agarwal2025Integrating}
Agarwal, A., Kenney, A.~M., Tan, Y.~S., Tang, T.~M.,  and Yu, B. (2025), \enquote{Integrating Random Forests and Generalized Linear Models for Improved Accuracy and Interpretability,} \textit{arXiv preprint arXiv:2501.12345}.

\bibitem[{Agarwal et~al.(2022)Agarwal, Tan, Ronen, Singh, and Yu}]{pmlr-v162-agarwal22b}
Agarwal, A., Tan, Y.~S., Ronen, O., Singh, C.,  and Yu, B. (2022), \enquote{Hierarchical Shrinkage: Improving the accuracy and interpretability of tree-based models.} in \textit{International Conference on Machine Learning}, PMLR, pp. 111--135.

\bibitem[{Basu et~al.(2018)Basu, Kumbier, Brown, and Yu}]{Basu2018iterative}
Basu, K., Kumbier, K., Brown, J.~B.,  and Yu, B. (2018), \enquote{Iterative random forests to discover predictive and stable high-order interactions,} \textit{Proceedings of the National Academy of Sciences}, 115, 1943--1948.

\bibitem[{Belgiu and {{Dr\u{a}gu{\c{t}}}}(2016)}]{belgiu2016random}
Belgiu, M.,  and {{Dr\u{a}gu{\c{t}}}}, L. (2016), \enquote{Random forest in remote sensing: A review of applications and future directions,} \textit{ISPRS journal of photogrammetry and remote sensing}.

\bibitem[{Bertsimas and Van~Parys(2020)}]{bertsimas2020sparse}
Bertsimas, D.,  and Van~Parys, B. (2020), \enquote{Sparse high-dimensional regression,} \textit{The Annals of Statistics}, 48, 300--323.

\bibitem[{Biblarz and Raftery(1993)}]{biblarz1993effects}
Biblarz, T.~J.,  and Raftery, A.~E. (1993), \enquote{The effects of family disruption on social mobility,} \textit{American Sociological Review}, 97--109.

\bibitem[{Bischl et~al.(2017)Bischl, Casalicchio, Feurer, Gijsbers, Hutter, Lang, Mantovani, van Rijn, and Vanschoren}]{bischl2017openml}
Bischl, B., Casalicchio, G., Feurer, M., Gijsbers, P., Hutter, F., Lang, M., Mantovani, R.~G., van Rijn, J.~N.,  and Vanschoren, J. (2017), \enquote{Openml benchmarking suites,} \textit{arXiv preprint arXiv:1708.03731}.

\bibitem[{Boyd and Vandenberghe(2004)}]{boyd2004convex}
Boyd, S.~P.,  and Vandenberghe, L. (2004), \textit{Convex optimization}, Cambridge University Press.

\bibitem[{Breiman(1995)}]{breiman1995better}
Breiman, L. (1995), \enquote{Better subset regression using the nonnegative garrote,} \textit{Technometrics}, 37, 373--384.

\bibitem[{Breiman(2001{\natexlab{a}})}]{breiman2001random}
Breiman, L. (2001{\natexlab{a}}), \enquote{Random forests,} \textit{Machine Learning}, 45, 5--32.

\bibitem[{Breiman(2001{\natexlab{b}})}]{breiman2001statistical}
Breiman, L. (2001{\natexlab{b}}), \enquote{Statistical modeling: The two cultures (with comments and a rejoinder by the author),} \textit{Statistical Science}, 16, 199--231.

\bibitem[{Buciluǎ et~al.(2006)Buciluǎ, Caruana, and Niculescu-Mizil}]{bucilua2006model}
Buciluǎ, C., Caruana, R.,  and Niculescu-Mizil, A. (2006), \enquote{Model compression,} in \textit{Proceedings of the 12th ACM SIGKDD International Conference on Knowledge Discovery and Data Mining}, ACM, pp. 535--541.

\bibitem[{B{\"u}hlmann and Van De~Geer(2011)}]{buhlmann2011statistics}
B{\"u}hlmann, P.,  and Van De~Geer, S. (2011), \textit{Statistics for high-dimensional data: methods, theory and applications}, Springer Science \& Business Media.

\bibitem[{Chen and Ishwaran(2012)}]{chen2012random}
Chen, X.,  and Ishwaran, H. (2012), \enquote{Random forests for genomic data analysis,} \textit{Genomics}.

\bibitem[{Chipman et~al.(2010)Chipman, George, and McCulloch}]{chipman2010bart}
Chipman, H.~A., George, E.~I.,  and McCulloch, R.~E. (2010), \enquote{{BART: Bayesian additive regression trees},} \textit{The Annals of Applied Statistics}, 4, 266 -- 298.

\bibitem[{Davern et~al.(2024)Davern, Bautista, Freese, Herd, and Morgan}]{GSS2024}
Davern, M., Bautista, R., Freese, J., Herd, P.,  and Morgan, S.~L. (2024), \enquote{General Social Survey 1972-2024,} NORC ed. Chicago.

\bibitem[{Du and Linero(2019)}]{du19a}
Du, J.,  and Linero, A.~R. (2019), in \textit{Proceedings of the 22nd International Conference on Artificial Intelligence and Statistics (AISTATS)}, eds. de~G.~Matthews, A.~G.,  and Singh, A., Naha, Okinawa, Japan: PMLR, vol.~89 of \textit{Proceedings of Machine Learning Research}, pp. 2796--2806.

\bibitem[{Duran and Grossmann(1986)}]{duran1986outer}
Duran, M.~A.,  and Grossmann, I.~E. (1986), \enquote{An outer-approximation algorithm for a class of mixed-integer nonlinear programs,} \textit{Mathematical Programming}, 36, 307--339.

\bibitem[{Efron and Hastie(2021)}]{efron2021computer}
Efron, B.,  and Hastie, T. (2021), \textit{Computer age statistical inference, student edition: algorithms, evidence, and data science}, vol.~6, Cambridge University Press.

\bibitem[{Fletcher and Leyffer(1994)}]{fletcher1994solving}
Fletcher, R.,  and Leyffer, S. (1994), \enquote{Solving mixed integer nonlinear programs by outer approximation,} \textit{Mathematical programming}, 66, 327--349.

\bibitem[{Friedman(2002)}]{friedman2002stochastic}
Friedman, J.~H. (2002), \enquote{Stochastic gradient boosting,} \textit{Computational Statistics \& Data Analysis}, 38, 367--378.

\bibitem[{Friedman and Popescu(2008)}]{friedman2008predictive}
Friedman, J.~H.,  and Popescu, B.~E. (2008), \enquote{Predictive learning via rule ensembles,} \textit{The Annals of Applied Statistics}, 2, 916--954.

\bibitem[{Friedman et~al.(2003)Friedman, Popescu, et~al.}]{friedman2003importance}
Friedman, J.~H., Popescu, B.~E. et~al. (2003), \enquote{Importance sampled learning ensembles,} \textit{Journal of Machine Learning Research}, 94305, 1--32.

\bibitem[{{Gurobi Optimization, LLC}(2024)}]{gurobi}
{Gurobi Optimization, LLC} (2024), \textit{Gurobi Optimizer Reference Manual}.

\bibitem[{Haslett and Raftery(1989)}]{haslett1989space}
Haslett, J.,  and Raftery, A.~E. (1989), \enquote{Space-time modelling with long-memory dependence: Assessing Ireland's wind power resource,} \textit{Journal of the Royal Statistical Society: Series C (Applied Statistics)}, 38, 1--21.

\bibitem[{Hastie et~al.(2009)Hastie, Tibshirani, and Friedman}]{hastie2009elements}
Hastie, T., Tibshirani, R.,  and Friedman, J. (2009), \textit{The Elements of Statistical Learning: Data Mining, Inference, and Prediction}, New York: Springer, 2nd ed.

\bibitem[{Hazimeh et~al.(2023)Hazimeh, Mazumder, and Radchenko}]{hazimeh2023grouped}
Hazimeh, H., Mazumder, R.,  and Radchenko, P. (2023), \enquote{Grouped variable selection with discrete optimization: Computational and statistical perspectives,} \textit{The Annals of Statistics}, 51, 1--32.

\bibitem[{Knaus et~al.(1985)Knaus, Draper, Wagner, and Zimmerman}]{Knaus1985APACHEII}
Knaus, W.~A., Draper, E.~A., Wagner, D.~P.,  and Zimmerman, J.~E. (1985), \enquote{APACHE II: A severity of disease classification system,} \textit{Critical Care Medicine}, 13, 818--829.

\bibitem[{{Laura and John Arnold Foundation}(2013)}]{LJAF2013PSA}
{Laura and John Arnold Foundation} (2013), \enquote{Public Safety Assessment: Risk Factors and Formula,} Tech. rep., LJAF.

\bibitem[{Liu and Mazumder(2023{\natexlab{a}})}]{liu2023fire}
Liu, B.,  and Mazumder, R. (2023{\natexlab{a}}), \enquote{Fire: An optimization approach for fast interpretable rule extraction,} in \textit{Proceedings of the 29th ACM SIGKDD Conference on Knowledge Discovery and Data Mining}, pp. 1396--1405.

\bibitem[{Liu and Mazumder(2023{\natexlab{b}})}]{liu2023forestprune}
Liu, B.,  and Mazumder, R. (2023{\natexlab{b}}), \enquote{Forestprune: Compact depth-pruned tree ensembles,} in \textit{International Conference on Artificial Intelligence and Statistics}, PMLR, pp. 9417--9428.

\bibitem[{Mazumder et~al.(2023)Mazumder, Radchenko, and Dedieu}]{mazumder2023subset}
Mazumder, R., Radchenko, P.,  and Dedieu, A. (2023), \enquote{Subset selection with shrinkage: Sparse linear modeling when the SNR is low,} \textit{Operations Research}, 71, 129--147.

\bibitem[{Meinshausen(2010)}]{meinshausen2010node}
Meinshausen, N. (2010), \enquote{Node harvest,} \textit{The Annals of Applied Statistics}, 2049--2072.

\bibitem[{{MOSEK ApS}(2024)}]{mosek}
{MOSEK ApS} (2024), \textit{MOSEK Optimizer API Manual}.

\bibitem[{Petersen et~al.(2008)Petersen, Pedersen, et~al.}]{petersen2008matrix}
Petersen, K.~B., Pedersen, M.~S. et~al. (2008), \enquote{The matrix cookbook,} \textit{Technical University of Denmark}, 7, 510.

\bibitem[{Pferschy and Schauer(2009)}]{pferschy2009knapsack}
Pferschy, U.,  and Schauer, J. (2009), \enquote{The knapsack problem with conflict graphs,} \textit{Journal of Graph Algorithms and Applications}, 13, 233--249.

\bibitem[{Radchenko and James(2010)}]{radchenko2010variable}
Radchenko, P.,  and James, G.~M. (2010), \enquote{Variable selection using adaptive nonlinear interaction structures in high dimensions,} \textit{Journal of the American Statistical Association}, 105, 1541--1553.

\bibitem[{Shwartz-Ziv and Armon(2022)}]{shwartz2022tabular}
Shwartz-Ziv, R.,  and Armon, A. (2022), \enquote{Tabular data: Deep learning is not all you need,} \textit{Information Fusion}, 81, 84--90.

\bibitem[{Tan et~al.(2025)Tan, Singh, Nasseri, Agarwal, Duncan, Ronen, Epland, Kornblith, and Yu}]{tan2025fast}
Tan, Y.~S., Singh, C., Nasseri, K., Agarwal, A., Duncan, J., Ronen, O., Epland, M., Kornblith, A.,  and Yu, B. (2025), \enquote{Fast Interpretable Greedy-Tree Sums,} \textit{Proceedings of the National Academy of Sciences}, 122, e2310151122.

\bibitem[{Tibshirani(1996)}]{tibshirani1996regression}
Tibshirani, R. (1996), \enquote{Regression shrinkage and selection via the lasso,} \textit{Journal of the Royal Statistical Society: Series B (Methodological)}, 58, 267--288.

\bibitem[{Zhang(2010)}]{zhang2010nearly}
Zhang, C.-H. (2010), \enquote{Nearly unbiased variable selection under minimax concave penalty,} \textit{The Annals of Statistics}, 38, 894--942.

\bibitem[{Zou and Hastie(2005)}]{zou2005regularization}
Zou, H.,  and Hastie, T. (2005), \enquote{Regularization and variable selection via the elastic net,} \textit{Journal of the Royal Statistical Society: Series B (Statistical Methodology)}, 67, 301--320.

\end{thebibliography}

\end{document}